\documentclass{WileyMSP-template}

\usepackage{graphicx}%
\usepackage{multirow}%
\usepackage{amsmath,amssymb,amsfonts}%
\usepackage{amsthm}%
\usepackage{mathrsfs}%
\usepackage{longtable}
\usepackage{adjustbox}
\usepackage[title]{appendix}%
\usepackage{tabularx}
\usepackage{xcolor}%
\usepackage{graphicx}
\usepackage{etoolbox}
\usepackage{textcomp}%
\usepackage{notoccite}
\usepackage{manyfoot}%
\usepackage{placeins} 
\usepackage{booktabs}%
\usepackage{algorithm}%
\usepackage{algorithmicx}%
\usepackage{algpseudocode}%
\usepackage{listings}%
\usepackage{hyperref}
\usepackage{CJKutf8}
\usepackage{array}
\usepackage{tabularray}
\usepackage{geometry} 
\usepackage{lineno}
\usepackage{graphicx} 
\usepackage{adjustbox} 
\usepackage{float}
\usepackage{aliascnt}   

\begin{document}

\pagestyle{fancy}
\rhead{\includegraphics[width=2.5cm]{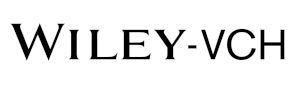}}

\title{GloPath: An Entity-Centric Foundation Model for Glomerular Lesion Assessment and Clinicopathological Insights}

\maketitle


\author{Qiming He+}
\author{Jing Li+}
\author{Tian Guan+}
\author{Yifei Ma}
\author{Zimo Zhao}
\author{Yanxia Wang}
\author{Hongjing Chen}
\author{Yingming Xu}
\author{Shuang Ge}
\author{Yexing Zhang}
\author{Yizhi Wang}
\author{Xinrui Chen}
\author{Lianghui Zhu}
\author{Yiqing Liu}
\author{Qingxia Hou}
\author{Shuyan Zhao}
\author{Xiaoqin Wang}
\author{Lili Ma}
\author{Peizhen Hu}
\author{Qiang Huang}
\author{Zihan Wang}
\author{Zhiyuan Shen}
\author{Junru Cheng}
\author{Siqi Zeng}
\author{Jiurun Chen}
\author{Zhen Song}
\author{Chao He*}
\author{Zhe Wang*}
\author{Yonghong He*}


\dedication{}

\begin{affiliations}
Prof. Qiming He \textsuperscript{1,2,6}, Prof Jing Li \textsuperscript{3}, Prof. Tian Guan \textsuperscript{1}, Dr. Yifei Ma \textsuperscript{4}, Dr. Zimo Zhao \textsuperscript{4}, Prof. Yanxia Wang \textsuperscript{3}, Dr. Hongjing Chen \textsuperscript{8}, Dr. Yingming Xu \textsuperscript{1}, Dr. Shuang Ge \textsuperscript{1,7}, Dr. Yexing Zhang \textsuperscript{9}, Dr. Yizhi Wang \textsuperscript{1}, Dr. Xinrui Chen \textsuperscript{1}, Prof. Lianghui Zhu \textsuperscript{1}, Prof. Yiqing Liu \textsuperscript{1}, Dr. Qingxia Hou \textsuperscript{5}, Dr. Zihan Wang \textsuperscript{1}, Dr. Zhiyuan Shen \textsuperscript{1}, Dr. Junru Cheng\textsuperscript{1}, Dr. Siqi Zeng\textsuperscript{1}, Dr. Jiurun Chen\textsuperscript{1,7}, Prof. Zhen Song \textsuperscript{7}, Prof. Chao He \textsuperscript{4}, Prof. Zhe Wang  \textsuperscript{3}, Prof. Yonghong He \textsuperscript{1}

\textsuperscript{1} Institute of Biopharmaceutical and Health Engineering, Tsinghua Shenzhen International Graduate School, Tsinghua University, Shenzhen, China

\textsuperscript{2} Interdisciplinary Institute for Medical Engineering, Fuzhou University, Fuzhou, China

\textsuperscript{3} Department of Pathology, School of Basic Medicine and Xijing Hospital, Fourth Military Medical University, Xi'an, China

\textsuperscript{4} Department of Engineering Science, University of Oxford, Oxford, UK

\textsuperscript{5} Department of Pathology, Xijing Hospital, Fourth Military Medical University, Xi'an, China

\textsuperscript{6} Medical Optical Technology R\&D Center, Research Institute of Tsinghua, Pearl River Delta, Guangzhou, China

\textsuperscript{7} Peng Cheng Laboratory, Shenzhen, China

\textsuperscript{8} Department of Pathology, Ankang Central Hospital, Ankang, China

\textsuperscript{9} Jinan Inspur Data Technology Co., Ltd., Jinan, China

\textsuperscript{+} Qiming He, Jing Li, and Tian guan contribute equalliy to this work.

\textsuperscript{*} Chao He, Zhe Wang, and Yonghong He are the corresponding authors.

\end{affiliations}



\begin{abstract}

Glomerular pathology is central to the diagnosis and prognosis of renal diseases, yet the heterogeneity of glomerular morphology and fine-grained lesion patterns remain challenging for current AI approaches. We present \textbf{GloPath}, an entity-centric foundation model trained on over one million glomeruli extracted from 14,049 renal biopsy specimens using multi-scale and multi-view self-supervised learning. GloPath addresses two major challenges in nephropathology: glomerular lesion assessment and clinicopathological insights discovery. For lesion assessment, GloPath was benchmarked across three independent cohorts on 52 tasks—including lesion recognition, grading, few-shot classification, and cross-modality diagnosis—outperforming state-of-the-art methods in 42 tasks (80.8\%). In the large-scale real-world study, it achieved an ROC-AUC of 91.51\% for lesion recognition, demonstrating strong robustness in routine clinical settings. For clinicopathological insights, GloPath systematically revealed statistically significant associations between glomerular morphological parameters and clinical indicators across 224 morphology–clinical variable pairs, demonstrating its capacity to connect tissue-level pathology with patient-level outcomes. Together, these results position GloPath as a scalable and interpretable platform for glomerular lesion assessment and clinicopathological discovery, representing a step toward clinically translatable AI in renal pathology.

\end{abstract}

\section{Introduction}
Kidney disease affects over 850 million people worldwide and is projected to become the fifth leading cause of death by 2040 \cite{francis2024chronic, khadzhynov2019incidence}. Renal pathology remains the gold standard for definitive diagnosis and staging, with glomeruli serving as the most critical functional and diagnostic units \cite{murray2023histology, fogo2015glomerulus, pollak2014glomerulus, fogo2024crosstalk, clair2024spatially}. Lesion assessment within glomeruli provides essential information for disease classification, prognosis estimation, and therapeutic decision-making. Beyond lesion recognition and grading, clinicopathological insights—derived from linking glomerular morphology with patient-level variables such as laboratory indices and clinical outcomes—offer unique opportunities to uncover disease mechanisms and inform precision medicine in nephrology.

However, glomeruli are highly heterogeneous entities. Their morphology reflects complex interactions among specialized cells, basement membranes, and mesangial matrix, and varies widely across disease processes such as sclerosis, inflammation, and capillary loop remodeling. Subtle yet diagnostically significant changes often require both fine-grained recognition of microstructural alterations and contextual interpretation of their spatial organization. This heterogeneity makes manual evaluation challenging and underscores the need for computational approaches that explicitly model glomeruli as the fundamental pathological entities of the kidney.

Artificial intelligence (AI) has shown promise in computational pathology, but current approaches fall short in glomerular pathology. Most models are slide-level or patch-based and lack explicit entity-centric representation learning. As a result, they struggle to capture disease-relevant microstructural features, encode long-range dependencies within glomeruli, and generalize across cohorts or modalities \cite{santo2020artificial, zheng2021deep, lei2024artificial, deng2025kpis, ochi2025pathology, campanella2025clinical, aben2024towards, lee2025benchmarking, bilal2025foundation}. These limitations hinder their utility in clinically important tasks such as few-shot learning, multi-cohort validation, and cross-modality adaptation.

\begin{figure}[H]
  \centering
  \includegraphics[width=\textwidth]{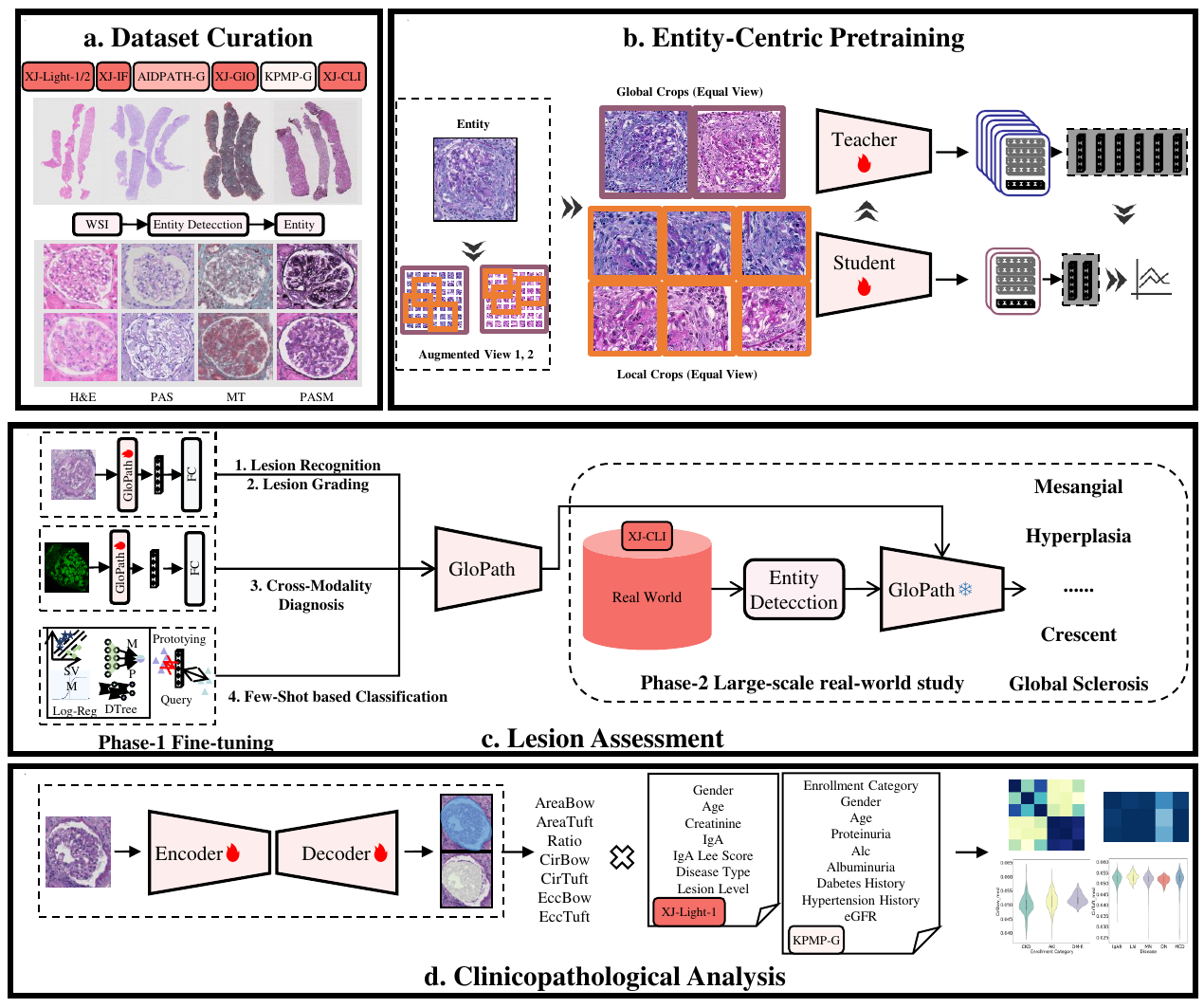}
  \caption{Overview of GloPath. \textbf{a, } Dataset curation. The pathological images are obtained from seven cohorts including XJ-Light-1, XJ-Light-2, XJ-IF, AIDPATH-G, XJ-GIO and KPMP-G, where WSI is processed by entity detection. \textbf{b, }Entity-centric self-supervised pretraining based on multi-scale multi-view. \textbf{c, }Lesion assessment. \textbf{d, }Clinicopathological correlation analysis.} \label{overview}
\end{figure}

To address these challenges, we developed GloPath, an entity-centric foundation model specifically designed for glomerular pathology (Figure \ref{overview}). In contrast to conventional patch-based pretraining that treats local image regions as independent samples, GloPath adopts glomeruli as the fundamental learning units, with each training instance corresponding to a biologically defined pathological entity automatically localized from whole-slide images. Pre-trained on over one million glomeruli extracted from 14,049 renal biopsy specimens using self-supervised multi-scale and multi-view learning, GloPath learns robust semantic representations of glomerular morphology with improved interpretability. We demonstrate its broad applicability across two key domains: glomerular lesion assessment, where GloPath achieves superior performance across diverse recognition, grading, and generalization tasks; and clinicopathological insights, where it reveals novel associations between glomerular morphology and patient-level clinical indicators. Together, these contributions highlight GloPath as a scalable and clinically translatable platform for computational nephropathology.

\section{Results}

\subsection{An overvew of GloPath}\label{sec2}

Figure \ref{overview} provides an overview of GloPath. To enable accurate and scalable glomerular lesion assessment and clinicopathological analysis, we first assembled a large-scale, multicenter, multi-staining renal pathology dataset. Pathological images were collected from seven independent cohorts, and glomerular entities were automatically extracted from whole-slide images (WSIs) using a pretrained detection model. In total, 1,017,879 glomeruli were obtained across four staining modalities—H\&E, PAS, Masson’s trichrome (MT), and PASM (Figure \ref{overview}a).

We then conducted self-supervised pretraining with multi-scale and multi-view constraints to obtain GloPath (Figure \ref{overview}b; Sec. \ref{entity_centric_pretraining}). This pretraining strategy enables the model to capture fine-grained cellular morphology alongside larger-scale structural organization, while preserving the integrity of glomerular entities and their contextual tissue features.

We systematically evaluated GloPath across two major domains. For glomerular lesion assessment (Figure \ref{overview}c), GloPath was fine-tuned on multiple datasets and demonstrated strong performance in lesion recognition, lesion grading, cross-modality diagnosis, and few-shot classification. The model was further deployed in a real-world clinical setting for open-set lesion recognition, where it exhibited robust generalization beyond closed-set training distributions. For clinicopathological insights (Figure \ref{overview}d), GloPath was used to extract quantitative glomerular morphological parameters, which were subsequently associated with clinical variables from two independent cohorts, revealing meaningful links between glomerular morphology and patient-level clinical indicators.

\subsection{An Overview of Lesion Assessment}\label{overview_lesion_assessment}

\begin{figure}
  \centering
  \includegraphics[width=\textwidth]{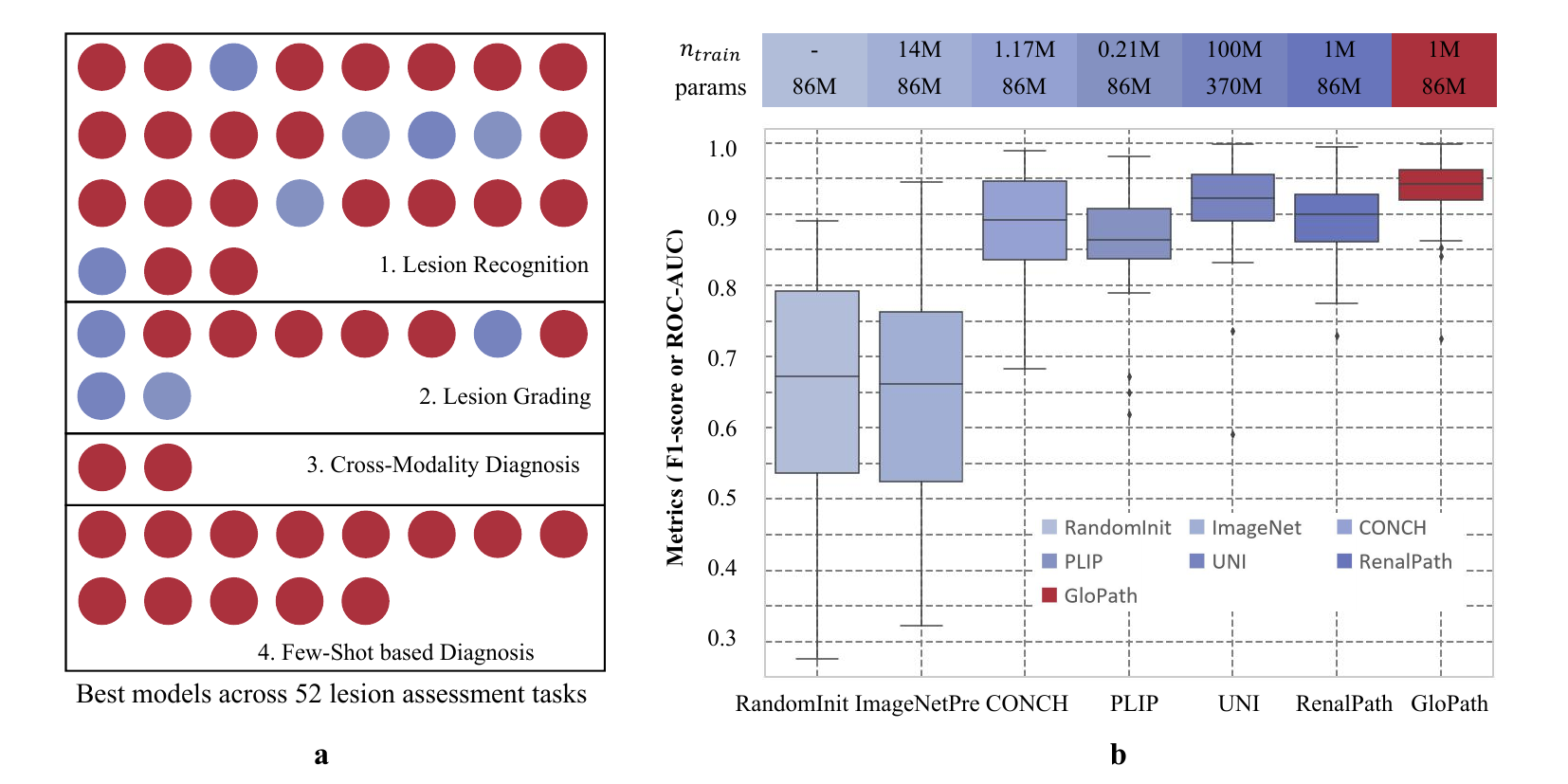}
  \caption{Overview of the performance of lesion assessment. \textbf{a, }The best model on each lesion assessment task across lesion recognition, lesion grading, cross-modality diagnosis, and few-shot-based diagnosis. Refer to the legend in (b). \textbf{b, }Comparison of the distribution of the results of lesion recognition.} \label{overview_la}
\end{figure}

Glomerular lesions constitute the core of glomerular pathology and are critical determinants of kidney disease classification, progression, and clinical outcomes. Leveraging an entity-centric pretraining strategy, GloPath was applied to four major lesion assessment tasks: lesion recognition, grading, cross-modality diagnosis, and few-shot classification. Across all 52 tasks, GloPath consistently achieved state-of-the-art performance, Across all 52 tasks, GloPath consistently achieved leading performance, demonstrating clear advantages over general-purpose pathology foundation models with substantially larger parameter counts, larger training datasets, and more diverse modalities (Figure \ref{overview_la}; Supplementary Material Sec. \ref{models_being_compared}).

Figure \ref{overview_la}a summarizes the best-performing models for each of the 52 lesion assessment tasks. Key observations include:
\begin{itemize}
\item Overall superiority: GloPath attained the highest performance on 42 out of 52 tasks (80.8\%), substantially exceeding the performance of UNI, demonstrating efficient parameter specificity despite a smaller model size and training volume.
\item Performance of lesion recognition and grading: For tasks on internal cohorts, GloPath led in 21 lesion recognition tasks and 6 grading tasks, highlighting its capacity to capture fine-grained entity features, consistent with the use of multi-scale and multi-view feature constraints.
\item Cross-modality generalization: GloPath achieved the overall performance lead in cross-modality diagnosis tasks, indicating that its learned glomerular representations generalize across different staining modalities.
\item Few-shot learning: On 13 external or cross-modality few-shot tasks, GloPath outperformed all baselines, supporting the robustness and specificity of its pretrained features.
\end{itemize}

Figure \ref{overview_la}b provides further quantitative evidence of GloPath’s advantages. First, it exhibits superior generalization, with overall performance significantly higher than all competing models (p \textless 0.001) and a median score approaching 0.95. Second, GloPath demonstrates remarkable data efficiency: compared with the second-best model (UNI), it achieves superior performance using only 1\% of the training data and 23.2\% of the model parameters, in contrast to methods such as CONCH and PLIP that rely on multimodal data. Third, GloPath shows strong stability, as indicated by the narrowest interquartile range among all models, concentrated in the high-performance region. Notably, the RenalPath model developed in this study ranked second only to UNI among general-purpose models, highlighting the effectiveness of large-scale renal pathology pretraining. The additional performance gains achieved by GloPath over RenalPath can be primarily attributed to entity-level pretraining on explicitly detected glomeruli, rather than differences in network architecture or training strategy, underscoring the importance of aligning pretraining units with disease-relevant pathological entities.

In summary, GloPath establishes a strong new benchmark in glomerular lesion assessment, combining high accuracy, data efficiency, and stability, providing a robust foundation for clinical deployment.

\subsubsection{Performance of Lesion Recognition}

\begin{figure}
  \centering
  \includegraphics[width=\textwidth]{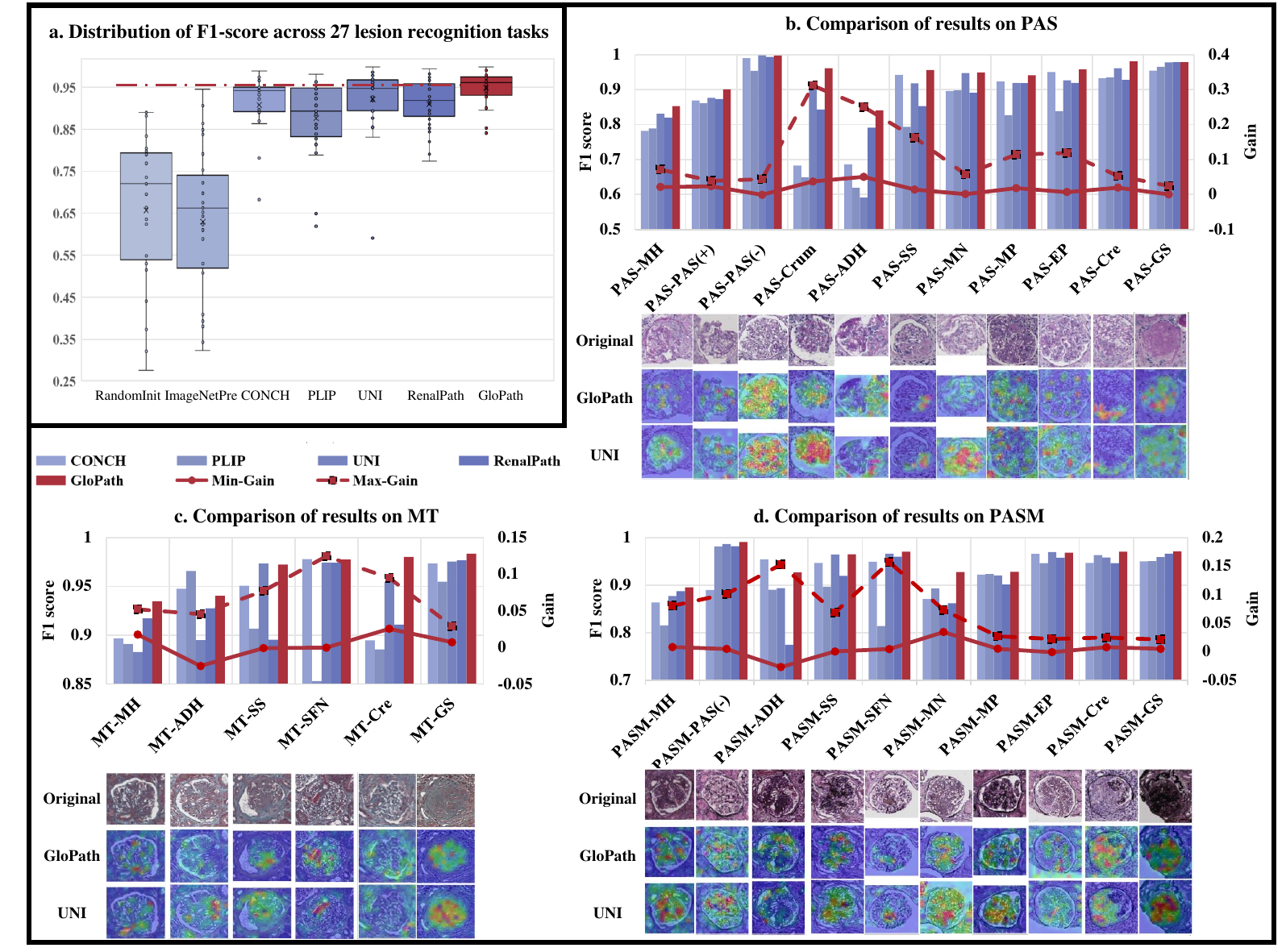}
  \caption{Comparison of results of glomerular lesion analysis. \textbf{a, }Comparison of the distribution of F1 score of lesion recognition. \textbf{b-d, }Comparison of results on PAS, MT and PASM stainings. The primary vertical axis corresponds to the F1 score of each model.
The secondary vertical axis corresponds to Min-Gain and Max-Gain, which are, respectively, the minimum and maximum performance gains of GloPath over all other models. Below the chart are the visualization results, where rows 1-3 are original images and the results of GloPath and UNI, respectively.} \label{la}
\end{figure}

Overall, GloPath significantly outperformed all competing models across the 27 lesion recognition tasks (all p \textless 0.05; Figure \ref{la}a). Both the median and mean F1 scores of GloPath exceeded 0.95, markedly higher than UNI (p = 6.407 × $10^{-7}$). Among the general-purpose foundation models, UNI performed slightly better than CONCH (p = 0.2386) and RenalPath (p = 0.0362), while PLIP showed only marginal improvements. RandomInit and ImageNet exhibited substantial performance deficits compared with all other models (all p \textless 0.05).
Stratified analyses by staining modality further confirmed GloPath’s advantage (Figure \ref{la}b–d). For PAS, MT, and PASM stainings, GloPath achieved the best results in 10 of 11, 3 of 6, and 8 of 10 lesion categories, respectively. Relative to the next-best model for each lesion, the maximum performance gains reached 7.05\% (PAS), 2.53\% (MT), and 3.41\% (PASM). On average, GloPath exceeded general-purpose pathology foundation models by 7.10\%, 3.40\%, and 2.86\% on PAS, MT, and PASM, respectively (p = 0.0010, 0.0313, and 0.0020). Compared with RenalPath, the corresponding gains were 4.62\%, 3.11\%, and 3.57\% (all p \textless 0.05), suggesting that RenalPath performed slightly better than other general-purpose models but remained inferior to GloPath. The performance gap was even more pronounced against non–pathology-based models. Here, GloPath achieved average gains of 34.63\% (PAS), 33.24\% (MT), and 24.39\% (PASM; all p \textless 0.05). Together, these results highlight not only GloPath’s superior diagnostic accuracy across multiple stainings but also its consistent ability to capture a higher proportion of lesion patterns at clinically meaningful performance levels.

GloPath consistently achieved the best performance across PAS, MT, and PASM stainings, with robust recognition accuracy and clinically relevant precision–recall balance (Figure \ref{la}b–d, Tables \ref{pas_27task}–\ref{pasm_27task}).
\begin{itemize}
\item \textbf{PAS}: Figure \ref{la}b and Table \ref{pas_27task} show that on PAS staining, GloPath achieves an overall lead in recognition performance with an overall F1 score of 0.938. Over 0.95 performance was achieved on PAS(-), Crum, SS, MN, ER, Cre, and GS lesions, and the largest gain compared to other models was achieved on Crum. In terms of the more challenging PR-AUC metrics, GloPath achieved about 0.8 and above performance on lesions such as MH, PAS(+), PAS(-), Cre and GS, which fully demonstrates that GloPath has already reached a fairly reliable performance for clinical application in these major lesion modes. In comparison, RenalPath and UNI only achieve about 0.8 and above performance on more typical lesions of MH, PAS(-), and GS, while only the latter of the non-pathology-foundation models RandomInit and ImageNetPre achieves more than 0.8 PR-AUC on GS.
\item \textbf{MT}: Figure \ref{la}c and Table \ref{mt_27task} show that GloPath has the best overall performance on MT staining, with an average F1 score of 0.965 on the diagnostic task for each lesion, and achieves more than 0.97 performance on lesion recognition for SS, SFN, Cre and GS. Compared to other models, GloPath achieves the highest maximum gain in SFN. In terms of PR-AUC, GloPath is able to achieve more than 0.75 performance on MH and GS, which indicates that PR-AUC is able to better balance precision and recall in a clinical setting. In comparison, among the other models, only UNI, PLIP and RenalPath are able to reach the corresponding level on GS.
\item \textbf{PASM}: Figure \ref{la}d and Table \ref{pasm_27task} show that on PASM, GloPath embodies the best performance on the vast majority of lesions, with an average F1 score of 0.951, and achieves more than 0.97 performance on lesions such as PAS(-), SFN, Cre and GS. In terms of PR-AUC, GloPath is the best performer, being able to achieve more than 80\% performance on lesions such as PAS(-), Cre, and GS, a number that is one and two more than UNI and RenalPath, respectively. This shows that GloPath has been able to have strong performance for clinical applications.
\end{itemize}

Grad-CAM–based visualization further underscored GloPath’s interpretability (Figure \ref{la}b–d). Compared with UNI, GloPath localized lesion regions more accurately and comprehensively. For instance, in PAS–MH, GloPath highlighted thylakoid stromal hyperplasia and mesangial proliferation; in MT–SS, it focused precisely on segmental sclerosis regions; and in PASM–Cre, it captured mural epithelial cell proliferation with high fidelity. These interpretable attention maps provide mechanistic plausibility for GloPath’s superior diagnostic accuracy and support its potential for clinical application.

\subsubsection{Performance of Lesion Grading}
Lesion grading provides a more fine-grained assessment of disease progression and carries substantial diagnostic value in renal pathology \cite{bueno2020data, tinawi2020update, moroni2014rapidly, woroniecki2009progression}. This task also enables direct evaluation of model parameter specificity across diverse morphological subtypes. We benchmarked GloPath and six comparative models on 10 lesion grading tasks (Table \ref{grading_10task}).

GloPath achieved the best overall performance, leading in 6 of 10 lesion types, while UNI and CONCH ranked first in 3 and 1 tasks, respectively. On average, GloPath outperformed RenalPath by 1.80\% (p=0.0039), general-purpose foundation models by 2.71\% (p=0.0039), and non-pathology models by 12.96\% (p=0.0020).

Performance gains varied by staining. Compared with RenalPath, general-purpose pathology models, and non-pathology models, GloPath achieved average improvements of 1.15\%, 1.70\%, and 11.28\% on PAS; 1.36\%, 3.87\%, and 19.40\% on MT; and 2.65\%, 3.13\%, and 11.42\% on PASM, respectively. Notably, the largest relative gain was observed for MT and PASM staining, while RenalPath consistently outperformed the general-purpose models.

Lesion-specific analyses further highlighted the fine-grained recognition capacity of GloPath. Relative to RenalPath, general-purpose, and non-pathology models, GloPath achieved average gains of 0.82\%, 2.75\%, and 18.33\% on MH; 2.82\%, 3.94\%, and 13.03\% on Cre; and up to 2.69\%, 1.80\%, and 11.43\% on EP. These results underscore the specificity of GloPath in capturing subtle morphological variations across lesion subtypes, while again indicating that RenalPath performs better overall than general-purpose pathology foundation models.

\begin{figure}
  \centering
  \includegraphics[width=\textwidth]{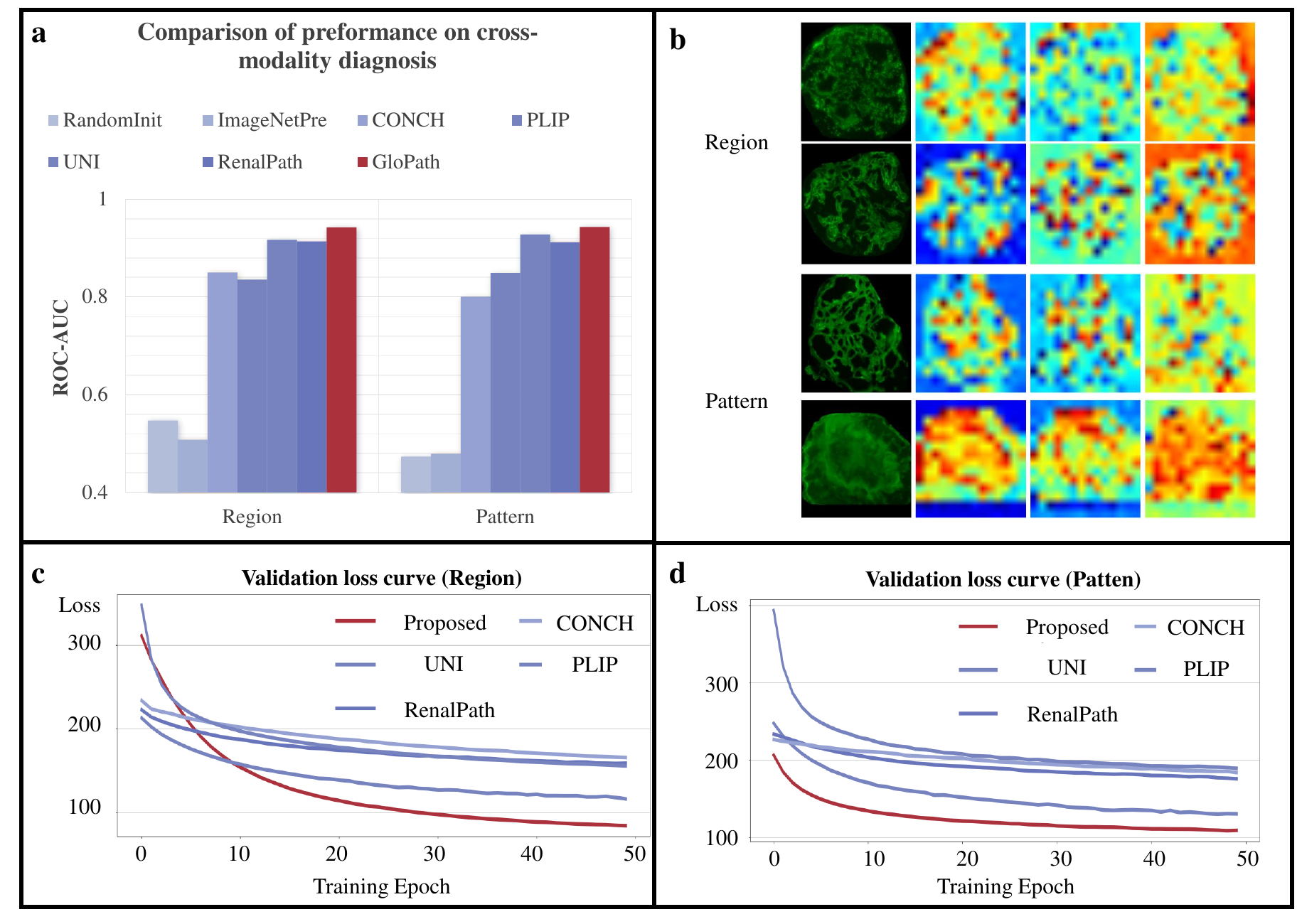}
  \caption{Comparison of results of cross-modality diagnosis. \textbf{a, }Comparison of preformance on cross-modality diagnosis. \textbf{b, }Visualization of attention map of GloPath. Row 1-2 indicate region classification and row 3-4 indicate pattern classification. Column 1 indicates the original IF images and column 2-4 indicate the attention map from three attention heads. \textbf{c-d, } Validation loss curves for region and pattenr respectively.} \label{if}
\end{figure}

\subsubsection{Performance of Cross-Modality Diagnosis}

Immunofluorescence is indispensable for renal pathology, allowing precise classification of immune-mediated glomerular diseases by visualizing immune complex deposition \cite{messias2024immunofluorescence, agrawal2025diagnostic, song5168817immunofluorescence}. Although fluorescence and bright-field images differ markedly in appearance, they share a common glomerular morphological framework. We therefore evaluated GloPath’s ability to generalize across modalities by testing its performance on immunofluorescence (IF)-based lesion classification.

Under full supervision, GloPath achieved the best performance in both deposition region and deposition pattern classification (Figure \ref{if}a; Table \ref{xj_if}). Relative to the suboptimal model UNI, GloPath yielded gains of 2.50\% and 1.57\%, respectively. Larger improvements were observed over PLIP and CONCH (9.94\% and 11.87\%), as well as over RenalPath (2.86\% and 3.14\%). Compared with non-pathology models, GloPath demonstrated substantial gains of 41.44\% and 46.72\%.. Across evaluation metrics, GloPath consistently achieved the highest performance, with an average ROC-AUC of 88.66\% for deposition region classification, exceeding deposition pattern classification by 0.51\%.

Visualization further supported the interpretability of cross-modality performance. Attention maps (Figure \ref{if}b) revealed that different attention heads captured complementary structural features—including glomerular interiors, boundaries, and fluorescent deposition signals—while the integrated attention mechanism localized lesions in a visually consistent manner. Training dynamics also reflected efficient domain adaptation: although initial loss values were higher for deposition region classification, GloPath showed the fastest decline and reached the lowest loss within 10 epochs (Figure \ref{if}c). For deposition pattern classification, GloPath maintained consistently lower loss values than all other models, supporting its capability to capture glomerular-level pathology across imaging modalities (Figure \ref{if}d). To qualitatively assess cross-modality representation alignment,Uniform Manifold Approximation and Projection (UMAP) was used to visualize feature embeddings from brightfield and IF images. The embeddings showed substantial overlap after IF adaptation, suggesting that IF adaptation reduces modality-induced divergence at the feature level (Figure \ref{umap}).

\subsubsection{Performance of Few-Shot based Classification}

\begin{figure}
  \centering
  \includegraphics[width=0.95\textwidth]{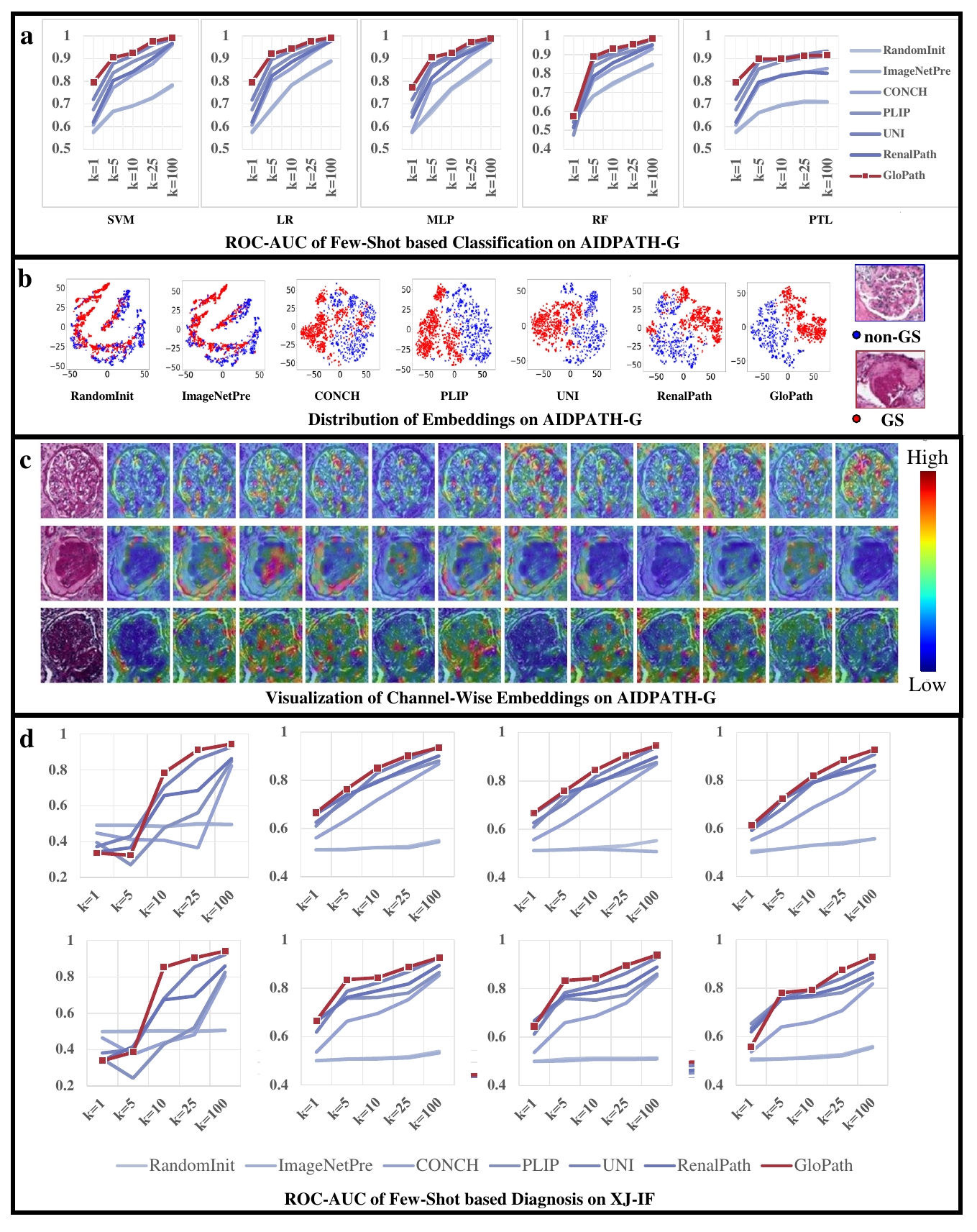}
  \caption{Comparison of results of few-shot based classification. \textbf{a, }Performance of few-shot based classification on AIDPATH-G. \textbf{b, }Distribution of embedding on AIDPATH-G using t-SNE. \textbf{c, }Visualization of Channel-wise embedding on AIDPATH-G. \textbf{d, } Results of few-shot for IF image classification. Row 1-2 indicate results for region, pattern respectively. Column 1-4 indicate results for different methods including LR, MLP, RF, and PTL respectively.} \label{fewshot}
\end{figure}

We first established a binary classification task on the AIDPATH-G dataset to detect the presence of glomerulosclerosis \cite{bueno2020data, bueno2020glomerulosclerosis}. While full supervision with GloPath demonstrated clear advantages, this setting alone was insufficient to fully validate its representational efficiency (Table \ref{aid_linear}). To further evaluate GloPath’s feature expressiveness, we compared few-shot performance across multiple classifiers—including SVM, Linear Regression (LR), Multi-Layer Perceptron (MLP), Random Forest (RF), and Prototype Learning (PTL)—using limited positive and negative samples (Figure \ref{fewshot}).

Across all classifiers, GloPath consistently achieved the best performance (Figure \ref{fewshot}a). With SVM, GloPath maintained the lead for all sample sizes ($k=1$ to 100), with the greatest relative advantage at lower $k$. For instance, with only one positive and one negative sample, GloPath achieved 79.45\% ROC-AUC, a 7.60\% gain over the next-best model (Table \ref{aid_svm}, $p=0.0371$). Similar trends were observed with LR, where GloPath outperformed competitors across all sample sizes, yielding 0.29–7.93\% improvements over the suboptimal model (all $p<0.05$) and 1.57–17.68\% improvements over RenalPath (Table \ref{aid_lr}). GloPath also led consistently in MLP, while in RF the performance gains were modest ($\leq$2.24\%) yet stable across sample sizes (Table \ref{aid_mlp}, \ref{aid_rf}). In PTL, GloPath’s advantage was most pronounced at small $k$, with gains of up to 21.89\% relative to non-foundation models (Table \ref{aid_tpl}).

Feature visualization further highlighted GloPath’s strong representations. Without downstream fine-tuning, t-SNE plots (Figure \ref{fewshot}b) showed that GloPath and PLIP separated normal and sclerotic glomeruli most clearly, while RandomInit and ImageNetPre failed to produce discernible boundaries. Attention maps (Figure \ref{fewshot}c) revealed complementary focus among attention heads: some emphasized glomerular interiors, others the surrounding context or Bowman’s capsule, suggesting biologically interpretable partitioning of structural features during pretraining.

Few-shot classification on IF images posed additional challenges, requiring cross-modality feature transfer and rapid adaptation from limited labeled data. Even under these stringent conditions, GloPath maintained robust performance (Figure \ref{fewshot}d). With LR, GloPath achieved maximum gains of 37.69\% and 42.40\% for deposition region and deposition pattern classification, respectively, with average gains of 24.86\% and 31.49\% across $k$ values (Tables \ref{region_lr}, \ref{pattern_lr}). With MLP, GloPath again achieved the best performance at all $k$. Notably, at $k=1$, GloPath achieved ROC-AUC values of 66.61\% and 66.53\% for region and pattern classification, outperforming UNI by 4.08\% and 4.75\% despite UNI’s substantially larger pretraining scale. As $k$ increased, GloPath’s relative margin narrowed but remained significant. Performance advantages over RF and PTL were smaller but still consistent across modalities and tasks (Tables \ref{region_rf}, \ref{pattern_rf}, \ref{region_tpl}, \ref{pattern_tpl}).

To further assess the stability and data efficiency of different models under few-shot settings, we analyzed the performance distributions across 10 independent repeated experiments at each k. As shown in the box plots (Figure \ref{few-aid-conf}-{few-if-conf}), GloPath achieved comparatively higher performance at low k values and reduced performance variance, particularly in low-data regimes on AIDPATH-G (Figure \ref{few-aid-conf}). In contrast, competing methods exhibited larger interquartile ranges and greater variability across repeated runs, suggesting higher sensitivity to data sampling. These results further indicate the relatively data-efficient and stable learning behavior of GloPath.

Together, these results demonstrate that GloPath achieves superior few-shot generalization in both bright-field and fluorescence modalities. Its robust feature representations enable rapid and accurate classification from extremely limited labeled data, highlighting strong potential for deployment in real-world clinical scenarios where annotations are scarce.

\subsubsection{Large-Scale Real-World Study}

To evaluate the clinical utility of GloPath beyond curated datasets, we constructed XJ-CLI, comprising 13,749 unfiltered PAS-stained glomeruli collected under routine clinical workflows. On this large-scale real-world cohort, GloPath achieved a mean ROC-AUC of 0.913 (as shown in Table \ref{openset}), representing only a 3.39\% decrease compared with the curated test set (XJ-Light-2).

At the lesion level, GloPath maintained high performance across diverse pathological entities. In XJ-CLI, seven lesions achieved ROC-AUCs greater than 0.90, including MH (0.897), Crum (0.983), SS (0.920), MN (0.919), MP (0.982), EP (0.999), and GS (0.997). Importantly, for clinically critical lesions such as MH, Cre, and GS, ROC-AUCs reached 0.897, 0.954, and 0.997, respectively. Notably, GloPath even outperformed its internal benchmark on MP (0.982 vs. 0.969) and EP (0.999 vs. 0.974), underscoring its robustness in clinical settings. Direct comparison with XJ-Light-2 further highlighted the generalization capacity of GloPath. While certain categories, such as PAS(+), Crum, and ADH, showed moderate decreases (–9.9\%, –15.2\%, and –14.9\%, respectively), performance remained stable or improved for others. The overall pattern indicates that GloPath retains strong discriminative ability under real-world data variability, achieving clinically actionable accuracy.

To characterize the failure modes of GloPath, representative false-negative (FN) and false-positive (FP) cases were analyzed for each lesion type in the XJ-CLI cohort (Figure \ref{error_analysis}). Overall, errors were observed across both common and lesion-specific patterns.
Among common error modes, false-positive predictions in membranous nephropathy (MN; FP, p = 0.7686) and mesangial proliferation (MP; FP, p = 0.9433) frequently occurred in glomeruli with relatively homogeneous but non-specific staining. Biopsy-related artifacts further contributed to misclassification, particularly in cases exhibiting elongated glomerular morphology and intensified staining. Such artifacts were associated with increased error rates, including MN FP (p = 0.9636), PAS(–) FN (p = 0.7891), and segmental sclerosis (SS) FN (p = 0.5964), as well as over-prediction in SS FP cases.
Lesion-specific errors were mainly concentrated in histologically borderline or confusable abnormalities. FP cases in ADH (p = 0.8186) and GS (p = 0.9293) showed morphological features overlapping with adjacent structures, leading to incorrect positive predictions. FN cases in mesangial hypercellularity (MH; p = 0.3555) were characterized by mild or early-stage changes, resulting in intermediate prediction scores. Similarly, FN cases in crescentic lesions (Cre) demonstrated partial parietal epithelial proliferation that did not reach diagnostic thresholds, contributing to under-recognition by the model.

To further assess the interpretability of GloPath, we conducted a quantitative analysis of attention–lesion alignment on the XJ-CLI cohort. Specifically, 10 Cre and 10 GS cases were randomly selected, and experienced nephrologists provided coarse bounding-box annotations to localize the dominant lesion regions within each glomerulus. This box-level annotation strategy was adopted to reduce annotation burden and inter-observer variability, while aligning with the classification-focused design of the task. Attention heatmaps were then used to compare attention intensities inside versus outside the annotated lesion boxes. Statistical analysis revealed that attention values within lesion-localized regions were significantly higher than those in non-lesion regions (0.5118 ± 0.0351 vs 0.3261 ± 0.0438 for Cre and 0.4907 ± 0.0534 vs 0.1816 ± 0.0439 for GS), with the distributional difference confirmed by the KS test ($D=1$ and $p < 0.0001$). Representative examples and quantitative summaries are shown in Figure \ref{interpre}. These results provide quantitative evidence that GloPath attention aligns with expert-annotated pathological regions, supporting the interpretability of attention maps beyond qualitative visualization.

In summary, the large-scale real-world study on XJ-CLI demonstrates that GloPath delivers rapid and reliable lesion recognition in routine clinical settings. The model consistently achieves high performance across 11 types of glomerular lesions, with most misclassifications arising in borderline cases or under suboptimal histological conditions, such as staining variability or biopsy-related artifacts. Importantly, these errors are largely confined to diagnostically ambiguous regions rather than overt lesion phenotypes, underscoring the robustness of GloPath’s learned representations. Together, these findings highlight GloPath’s strong translational potential, providing a foundation for timely, scalable, and clinically actionable computational pathology in real-world practice.

\subsection{Clinicopathological Insights}

Quantitative analysis of glomerular morphology parameters in pathology images offers a more objective and reproducible assessment than traditional subjective evaluation \cite{palmer2023cure, van2025unveiling, denic2016structural, li2022epidemiological, denic2018clinical}. Integrating these computationally derived morphological features with clinical data enables the identification of potential clinico-pathologic correlations that may serve as predictive biomarkers for disease progression \cite{holscher2023next, santo2020artificial, border2024investigating, shickel2024dawn}. Furthermore, pathomics-driven data mining based on morphological segmentation facilitates high-throughput, data-driven exploration of renal pathology, providing mechanistic insights into disease pathogenesis and progression and potentially guiding more precise diagnostic and therapeutic strategies.

\begin{figure}
  \centering
  \includegraphics[width=\textwidth]{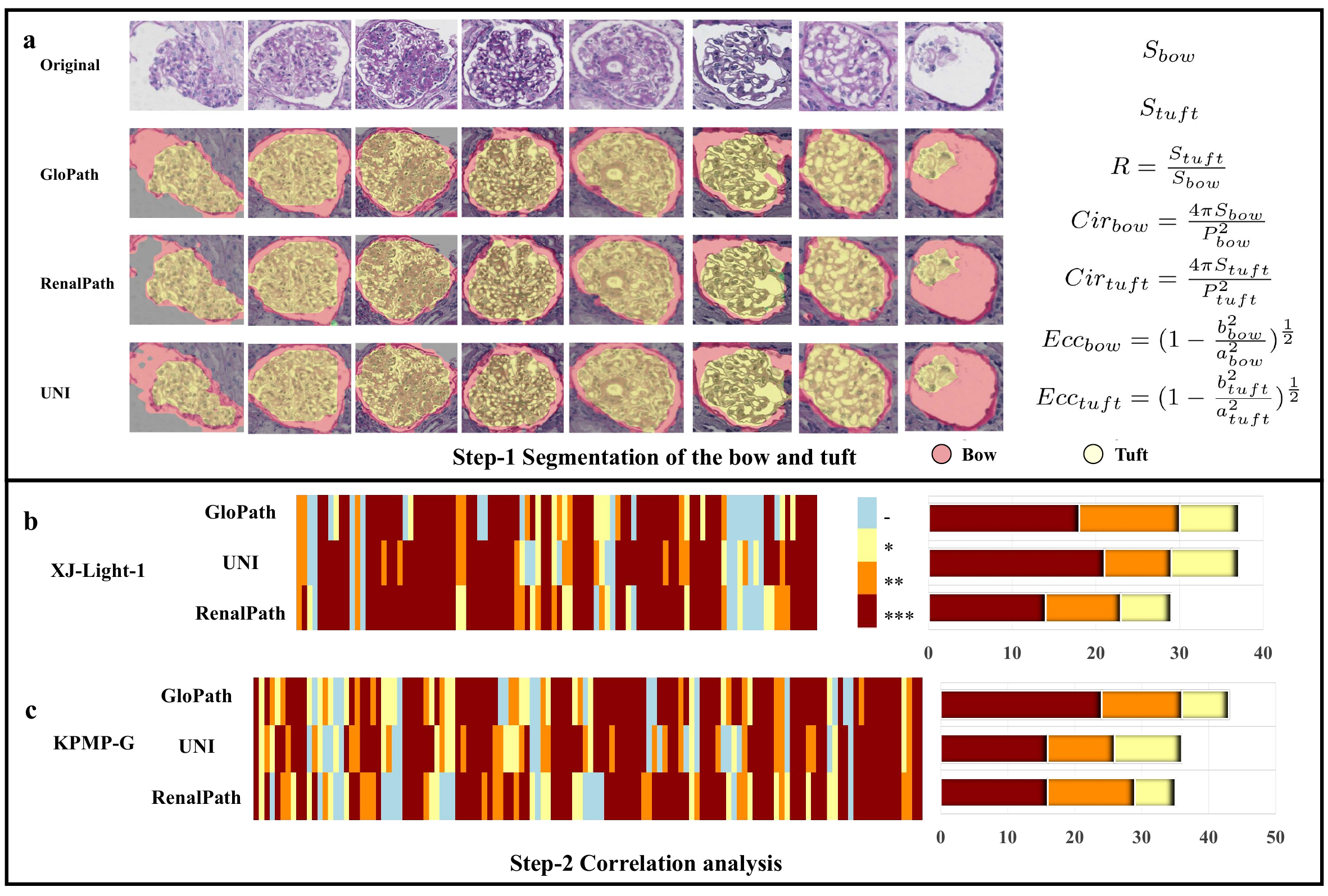}
  \caption{Overview of clinicopathological correlation analysis. \textbf{a, }Visualization of bow and tuft segmentation and morphological parameter definition. \textbf{b-c, }Comparison of distribution of significance levels on XJ-Light-1 and KPMP-G.} \label{omics}
\end{figure}

In this study, GloPath enabled accurate quantification of glomerular morphology (Figure \ref{omics}a), providing a robust foundation for pathomics analyses. Two cohorts with partial clinical information, XJ-Light-1 and KPMP-G, were utilized, and three models—GloPath, UNI, and RenalPath—were evaluated. The Kolmogorov–Smirnov (KS) and Kruskal–Wallis (KW) tests were employed to assess statistical differences in pathological phenotype distributions across clinical parameter groupings, thereby quantifying clinico-pathologic associations \cite{berger2014kolmogorov, mckight2010kruskal}.

As shown in Figure \ref{omics}b-c, morphological parameters quantified by GloPath yielded the most robust pathomics-based insights. In the XJ-Light-1 cohort, GloPath identified significant correlations in 37 of 98 pairs, matching UNI and outperforming RenalPath by 27.6\% (Table \ref{sig-level-private}). In the KPMP-G cohort, GloPath detected significant associations in 43 of 126 pathophysiological–clinical pairs, exceeding UNI by 7 pairs and RenalPath by 8 pairs (Table \ref{sig-level-public}). Among these, correlations were classified as extremely significant (***, n=7), significant (**, n=12), and moderate (*, n=24), underscoring the model’s capacity to uncover clinically relevant morphological patterns.

\subsubsection{Morphological Segmentation}

Quantitative glomerular morphology provides a reproducible basis for clinicopathological analyses, overcoming the subjectivity and labor-intensiveness of manual assessments \cite{kannan2019segmentation, bueno2020glomerulosclerosis, silva2022boundary, altini2020semantic, jiang2021deep, saikia2023mlp, yao2022glo, dimitri2022deep, li2021deep, liu2024hybrid, andreini2024enhancing, bouteldja2021deep, he2024identifying}. Here, we applied GloPath to segment two key glomerular structures, Bowman's capsule and tuft, across three cohorts: XJ-Light, XJ-GIO, and KPMP-G. Segmentor \cite{strudel2021segmenter} was used with GloPath and six comparative models.

As shown in Figure \ref{segmentation}, GloPath consistently achieved the highest IoU. On XJ-Light-1, segmentation of Bowman's capsule reached superior accuracy compared with UNI, despite using only 1\% of the training data and 23.2\% of the parameters. On XJ-GIO, tuft segmentation demonstrated similar gains, with IoU improvements of 8.22–9.17\% over general-purpose models PLIP and CONCH, 1.22–1.23\% over RenalPath, and 8.23–15.10\% over ImageNetPre and RandomInit. Segmentation performance for tuft was generally higher than for Bowman's capsule, with differences up to 7.21\% on PLIP.

Visual inspection further illustrates these results. Figure \ref{segmentation_vis}a shows Bowman's capsule segmentation on XJ-Light, where GloPath accurately delineates the structure with minimal errors, while UNI and RenalPath occasionally mislabel edge regions (e.g., row 3). Figure \ref{segmentation_vis}b shows tuft segmentation on PAS-stained XJ-GIO glomeruli, including challenging cases such as absent Bowman's capsule (row 1) or small tufts (row 2). GloPath maintained accurate segmentation, whereas UNI and RenalPath showed under- or over-segmentation, and PLIP exhibited noticeable edge over-segmentation. Figure \ref{omics}a shows Bow and tuft segmentation on KPMP-G; qualitative analysis indicates GloPath reliably segments extreme cases, such as extrusion deformation with incomplete Bowman's capsule (column 1, 3) and tuft with a small area (column 8), outperforming UNI and RenalPath in challenging scenarios.

In addition, we compared GloPath with the state-of-the-art medical image segmentation framework nnU-Net \cite{isensee2021nnu} under the same data splits. nnU-Net achieved IoU scores of 0.851 for Bowman's capsule and 0.905 for tuft segmentation, representing a strong task-specific upper bound. Notably, GloPath achieved comparable performance while relying on a unified entity-centric backbone rather than a fully supervised, task-specialized architecture.

These results demonstrate that GloPath enables accurate and robust glomerular morphological quantification across cohorts and staining conditions, establishing a solid foundation for downstream pathomics and clinicopathological correlation analyses.

\subsubsection{Insights on XJ-Light-1}

\begin{figure}
  \centering
  \includegraphics[width=0.9\textwidth]{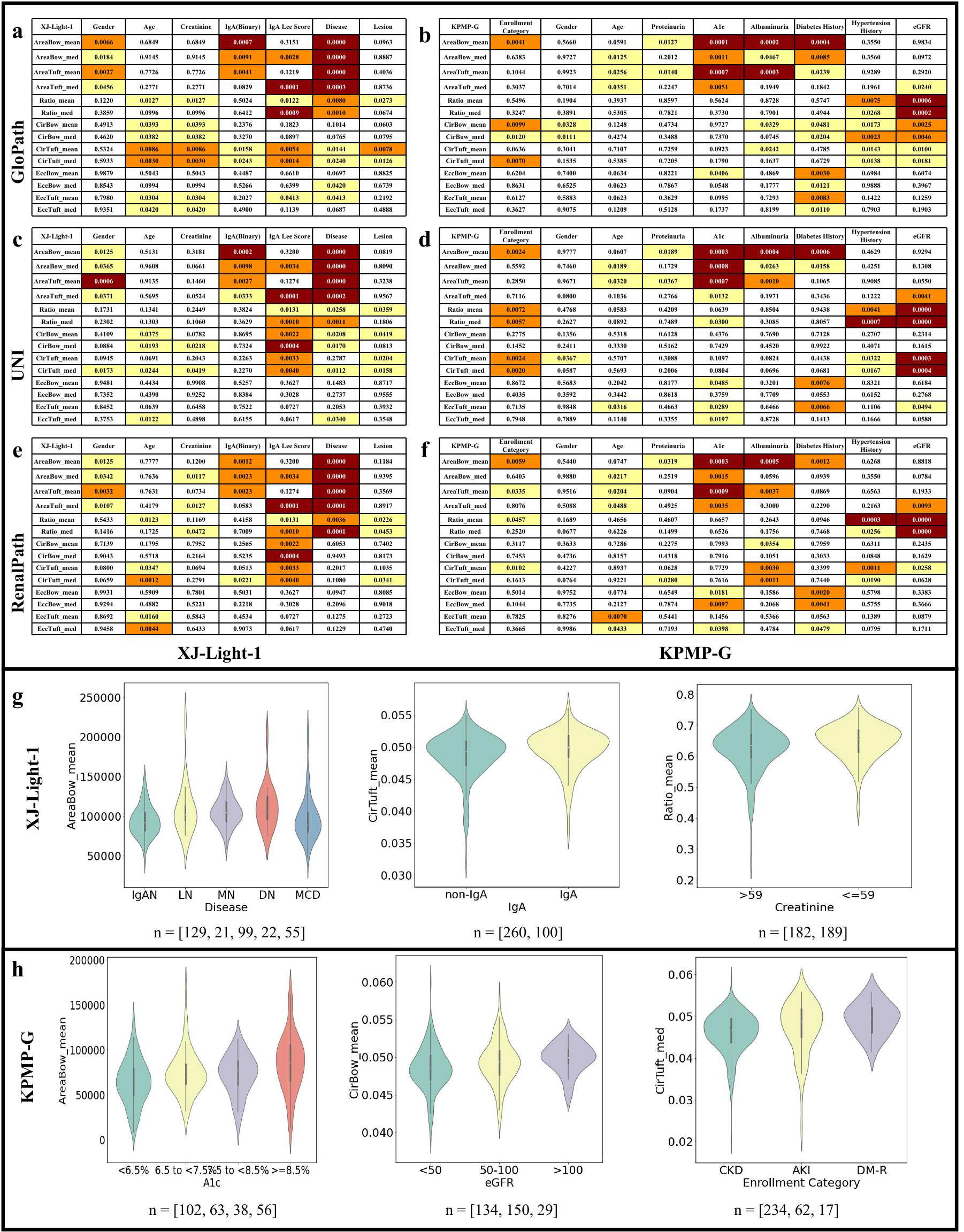}
  \caption{Details of clinicopathological correlation analysis. \textbf{a-f, }The p-values of correlation analysis. \textbf{a-b}, \textbf{c-d} and \textbf{e-f} refer to GloPath, UNI and RenalPath, respectively. The left and right column refer to XJ-Light-1 and KPMP-G respectively. \textbf{g-h} show examples from XJ-Light-1 and KPMP-G.} \label{omics_details}
\end{figure}

In the XJ-Light-1 cohort, GloPath revealed robust associations between glomerular morphological features and clinical parameters (Figure \ref{omics_details}a, c, e; Table \ref{gender-private}–\ref{lesion-private}). Notably, male patients exhibited larger glomerular areas than females, consistent with prior evidence suggesting estrogen’s protective role in limiting compensatory hypertrophy (Table \ref{gender-private}) \cite{messow1980sex}. Age negatively correlated with glomerular roundness, indicating progressive structural simplification, likely due to capillary loop collapse and basement membrane changes (Table \ref{age-private}) \cite{nyengaard1992glomerular}.

Disease-specific patterns were also evident. In IgA nephropathy, glomerular area exhibited a progressive downward shift with disease severity, consistent with a transition from early compensatory hypertrophy to subsequent sclerosis (Table \ref{IgA2-private}, \ref{IgALee-private}). 
Across nephropathy subtypes, glomerular area distributions showed distinct yet partially overlapping profiles: diabetic nephropathy (DN) and membranous nephropathy (MN) tended toward larger glomerular areas, suggestive of compensatory expansion; lupus nephritis (LN) demonstrated pronounced heterogeneity, spanning both hypertrophic and atrophic morphologies; whereas IgAN and minimal change disease (MCD) displayed relatively modest area variation, consistent with early mesangial alterations or largely preserved glomerular structure (Figure \ref{omics_details}g-left; Table \ref{disease-private}). Comparable distributional trends were observed for tuft-related morphological features, including area, roundness, and eccentricity, as well as for the tuft-to–Bowman’s capsule ratio, all of which varied in association with age, serum creatinine, injury severity, and disease category (Figure \ref{omics_details}g-right). A comprehensive summary of statistically significant pathomics–clinical associations is provided in Figure \ref{omics_private}.

To strengthen the clinical insights, we performed multivariate regression analysis (Table \ref{multi_vari_creatine}-\ref{multi_vari_disease}). After adjustment for sex, age, and disease type (IgAN, LN, MN, DN, and MCD), selected GloPath-derived glomerular morphological features were still significantly associated with clinical outcomes. For example, smaller ratio of area of tuft to that of bow was independently associated with higher serum creatinine levels (p = 0.0050 for Proposed\_CirTuft\_mean and p = 0.0006 for Proposed\_CirTuft\_med). Larger eccentricity was also significantly associated with higher severity of injury (p = 0.0244 for Proposed\_EccTuft\_mean and p = 0.0126 for Proposed\_EccTuft\_med). Collectively, these findings suggest that GloPath-derived glomerular morphological features are significantly associated with renal function, lesion severity, and disease status, even after accounting for conventional clinical covariates.

\subsubsection{Insights on KPMP-G}

In the KPMP-G cohort, encompassing a wider spectrum of clinical parameters, GloPath identified pronounced differences in glomerular morphology across patient groups (Figure \ref{omics_details}b, d, f; Table \ref{enrollment-public}–\ref{egfr-public}). CKD patients exhibited the largest glomerular areas and lowest circularity, indicative of compensatory hypertrophy due to nephron loss, whereas diabetes without renal involvement (DM-R) patients had the smallest areas and highest roundness, reflecting preserved structure. AKI patients showed signs of atrophy consistent with acute ischemic or inflammatory injury (Figure \ref{omics_details}h-right; Table \ref{enrollment-public}).

Additional clinicopathological trends were observed across demographic and functional variables. Sex-associated differences in glomerular roundness were detectable, potentially reflecting sex-related hemodynamic or structural variations, while age-related patterns revealed a non-linear distribution, with middle-aged patients exhibiting relatively larger glomerular areas and younger and elderly groups showing reduced values (Tables \ref{gender-public}, \ref{age-public}).
Functional and metabolic indicators also demonstrated consistent distributional associations with glomerular morphology. Proteinuria and HbA1c levels were associated with right-shifted glomerular area distributions, suggestive of compensatory hypertrophic responses under metabolic or filtration stress (Figure \ref{omics_details}h-left; Tables \ref{proteinuria-public}, \ref{a1c-public}). In contrast, albuminuria was linked to increased heterogeneity in glomerular area, indicating variable structural responses across disease states (Table \ref{albuminuria-public}).
Clinical history further modulated morphological patterns: diabetes and hypertension were associated with coordinated changes in both glomerular area and roundness, while eGFR showed a positive association with roundness, consistent with the notion that preserved glomerular geometry supports filtration efficiency (Figure \ref{omics_details}h-middle; Tables \ref{diabetes-public}, \ref{hypertension-public}, \ref{egfr-public}). Tuft-level morphology largely mirrored these distributional trends, with tuft-to–Bowman’s capsule ratios varying across hypertension and eGFR subgroups. A comprehensive summary of pathomics–clinical associations is provided in Figure \ref{omics_public}.

\subsubsection{Summary of Clinicopathological Insights}

In addition to statistical significance, we quantified effect sizes for the identified associations using $\epsilon^2$ (KW test) and the D statistic (KS test) (Supplementary Tables \ref{gender-private}–\ref{egfr-public}). Across both cohorts, most statistically significant clinicopathological associations exhibited small-to-moderate effect sizes (KW $\epsilon^2 < 0.1$ and KS D $\approx 0.1–0.2$). This pattern aligns with prior clinicopathological studies, in which individual histological or clinical variables typically explain only a limited fraction of phenotypic variance due to the indirect, non-linear, and multifactorial relationships between tissue-level pathology and patient-level clinical manifestations. In renal disease specifically, clinical phenotypes arise from the combined effects of structural lesions, molecular alterations, compensatory mechanisms, systemic factors, and their interactions, rather than from isolated pathological features. The observed associations, together with their consistent effect directions, support their robustness and clinical relevance despite the modest magnitude of individual effects.

Collectively, these analyses highlight the complex, multidimensional relationships between glomerular morphological phenotypes and clinical parameters across cohorts. Sex and age exert consistent effects, with males and middle-aged individuals exhibiting larger glomerular areas and structural variations indicative of compensatory or degenerative changes. Chronic conditions—including CKD, diabetes, and hypertension—further modulate glomerular size, roundness, and tuft-to-Bowman ratios, reflecting disease-specific adaptive or pathological remodeling. Moreover, the interplay among these factors—such as age-related vulnerability combined with chronic comorbidities—likely contributes to the observed phenotypic heterogeneity. Markers of renal function and injury, including proteinuria, albuminuria, and eGFR, correlate with structural alterations, emphasizing the functional relevance of these morphometric features.

These findings demonstrate that GloPath can systematically capture subtle yet clinically relevant glomerular changes, enabling high-resolution pathomics analyses. By integrating computationally derived morphology with patient-level clinical data, this framework provides a powerful approach for elucidating disease mechanisms, identifying potential biomarkers, and guiding personalized nephrology care. While the mechanistic underpinnings require further investigation, the results underscore the potential of GloPath to advance next-generation clinicopathological research and multi-factorial, data-driven insights in renal pathology.

\section{Discussion}\label{sec8}

In this study, we present GloPath, a domain-specific, entity-centric foundation model for glomerular pathology, enabling accurate lesion assessment and clinicopathological analysis across diverse cohorts. GloPath demonstrates robust recognition of atypical and variable lesions, clinically relevant performance on representative lesions, and effective generalization across external and cross-modality datasets. GloPath excels in recognizing atypical lesions with limited samples or heterogeneous morphology (e.g., PAS-Crum, PASM-ADH), likely due to its multi-scale, multi-view feature-constrained pretraining that encodes low signal-to-noise patterns within broader contexts. It also handles lesions with broad variability or multiple pathological grades (e.g., PAS-SS, MT-SS, PASM-MH), reflecting its capacity to integrate local and global morphological cues. For both common and diagnostically challenging lesions, GloPath achieves clinically reliable recognition performance. Importantly, multi-head attention maps and t-SNE visualizations of GloPath provide interpretable insights into the structural features driving model predictions, offering a bridge between AI-derived features and human pathology understanding. This demonstrates that GloPath not only quantifies subtle pathological phenotypes but also provides mechanistic insights, supporting translational applications in nephrology research.

Overall, GloPath effectively achieves intrinsic, fine-grained, and cross-modal characterization of glomerular pathology with fewer training samples, reduced model parameters, and limited modalities. A key factor underpinning GloPath’s success is its entity-centric pretraining. By treating detected glomeruli as fundamental units, the model focuses on diagnostic structures rather than heterogeneous tissue regions, embedding strong domain specificity from approximately one million glomerular entities. Multi-scale and multi-view strategies further enrich the representations: multi-scale learning captures features preserved across magnifications, while multi-view integration encodes local morphology alongside global semantic patterns, promoting scale-invariant and semantically consistent representations. Together, these design choices improve robustness to rare, heterogeneous, and multi-grade lesions.

Despite its encouraging generalization, several practical limitations remain. First, variations in scanners, acquisition settings, and staining protocols pose inherent challenges. Although GloPath is robust across institutions and staining types, most pretraining data come from a limited set of scanners and workflows, leaving subtle color or illumination differences underrepresented. While the entity-centric and multi-scale design reduces reliance on low-level appearance features, systematic scanner harmonization (e.g., color normalization or domain adaptation) was not explored and warrants future work. Second, rare or highly specific glomerulonephritis subtypes are underrepresented, limiting performance on uncommon lesions or early transitional forms, especially when they resemble more prevalent patterns. Expanding multi-center collections of rare lesions and incorporating hierarchical or uncertainty-aware labeling could improve reliability in these clinically challenging cases.

Beyond lesion recognition, GloPath enables efficient pathomics-based exploration of clinico-pathological correlations. Across the XJ-Light-1 and KPMP-G cohorts, glomerular area, roundness, and the tuft-to-Bowman’s capsule ratio were influenced by sex, age, disease subtype, and renal function indices such as proteinuria and eGFR. Male patients and individuals with CKD exhibited larger glomerular areas and lower circularity, whereas age-related changes were associated with structural simplification. Diabetes history, hypertension, and elevated HbA1c correlated with compensatory hypertrophy or morphological distortion, reflecting underlying pathophysiological processes. Distinct nephropathy subtypes displayed characteristic glomerular patterns: for example, compensatory hypertrophy in CKD and hyperfiltration-induced morphological changes in metabolic disease. These associations are consistent with established pathophysiological theories and reinforce the relevance of integrating computationally derived morphology with clinical data.

Integration of GloPath-derived morphological features with clinical parameters provides a novel layer of clinicopathological insight. While some associations remain at the research stage, they indicate potential for early, low-cost, non-invasive risk stratification using routine clinical data such as age, sex, proteinuria, and eGFR. Although effect sizes are generally small-to-moderate—reflecting the multifactorial, indirect nature of kidney disease—these systematic shifts are biologically interpretable and clinically meaningful. GloPath descriptors complement conventional clinical variables by capturing latent structural changes that may precede overt functional decline. Future multimodal frameworks combining these features with laboratory, imaging, or molecular data could enable personalized diagnostic and prognostic models, reduce reliance on invasive procedures, and enhance nephropathy monitoring \cite{canney2022risk, kishi2024oxidative}, ultimately improving patient care through timely, informed decision-making.

From a clinical perspective, GloPath provides rapid, reproducible, and accurate lesion recognition on both internal and real-world cohorts, mitigating inter-observer variability and supporting high-throughput analysis. Integration of morphological descriptors with routine clinical parameters enables subtle yet systematic detection of structural alterations, complementing conventional risk assessment. Although individual associations generally exhibit small-to-moderate effect sizes, this is consistent with the multifactorial nature of kidney disease, where multiple morphological and systemic factors collectively influence clinical outcomes. Future multimodal frameworks combining GloPath features with laboratory, imaging, or molecular data could facilitate personalized diagnostics and prognostics, reducing reliance on invasive procedures.

A potential concern is that a substantial portion of the pretraining and internal evaluation data originates from a single institution, raising the possibility that the model may inadvertently capture center-specific artifacts such as staining protocols or scanner characteristics. Several aspects of GloPath’s design and evaluation mitigate this risk. First, GloPath is pretrained at the entity level, focusing exclusively on automatically detected glomeruli rather than whole-slide regions that may encode background, tissue processing, or scanner-specific cues. This entity-centric formulation constrains the model to learn morphology-driven representations that are more tightly coupled to diagnostic structures. Second, the multi-scale and multi-view pretraining strategy encourages the learning of scale-invariant and semantically consistent features, reducing reliance on low-level appearance patterns that are more susceptible to institutional variation. Importantly, the robustness is empirically supported by its consistent performance on external cohorts (AIDPATH-G and KPMP-G), as well as across different imaging modalities. Together, these results suggest that GloPath captures transferable pathological semantics rather than institution-specific imaging artifacts, supporting its generalizability across diverse clinical settings.

Although this study spans multiple cohorts and staining modalities, additional validation across broader multi-center datasets is warranted. The granularity of morphological annotations remains a constraint, particularly in external cohorts lacking detailed labels. Moreover, while GloPath demonstrates strong predictive performance, further studies linking AI-derived morphological features with long-term clinical outcomes are necessary to fully establish clinical utility. Looking ahead, integrating GloPath into multimodal frameworks that combine histopathology, molecular data, and clinical parameters—particularly by leveraging recent advances in integrating morphology with spatial proteomics and spatial transcriptomics \cite{valanarasu2025multimodal,li2026ai}, or even training glomeruli-centric or renal-specific phenotype-protein-gene association models—holds promise for enabling more comprehensive disease modeling. Enhancements in self-supervised learning and domain adaptation may further improve model generalizability across rare pathologies and staining protocols. Ultimately, the deployment of domain-specific foundation models such as GloPath represents a significant step toward scalable, interpretable, and clinically actionable AI in renal pathology.

\section{Methods}\label{sec9}

\subsection{Dataset curation}
High-quality data form the cornerstone of progress in computational renal pathology. The marked heterogeneity of kidney biopsy specimens—arising from variable disease subtypes, staining protocols, and preparation artifacts—necessitates datasets that are both large-scale and well-curated to support the development of robust AI models. In this study, the construction of a comprehensive chronic kidney disease (CKD) pathology database was critical not only for formulating clinically relevant questions but also for driving algorithmic innovation and enabling rigorous validation in real-world settings.

We assembled seven cohorts, comprising five institutional cohorts (XJ-Light-1, XJ-Light-2, XJ-IF, XJ-GIO, and XJ-CLI) and two publicly available cohorts (AIDPATH-G and KPMP-G). Each cohort was assigned to distinct analytic tasks: pretraining of GloPath and RenalPath leveraged portions of XJ-Light-2 (excluding samples reserved for validation); lesion recognition and grading were performed on XJ-Light-1; cross-modality diagnosis and few-shot classification employed XJ-IF and AIDPATH-G; and clinicopathological correlation analysis drew upon morphological segmentation from XJ-GIO and KPMP-G, with quantitative analysis conducted on XJ-Light-1 and KPMP-G. Detailed characteristics of each cohort, including sample size, staining modality, and annotation strategy, are described below.

\subsubsection{XJ-Light-1 and XJ-Light-2}

Recognizing the limitations of existing public renal pathology datasets—namely, limited sample sizes, inconsistent annotations, and insufficient representation of disease heterogeneity—we established a large-scale, high-quality private database in collaboration with the Department of Pathology at Xijing Hospital, a leading tertiary care institution affiliated with the Air Force Medical University (Xi’an, China). This effort involved systematic patient data collection, digitization of pathology slides, preprocessing of whole-slide images, and expert lesion annotation.

\begin{itemize}
\item Patient Data Collection:Renal biopsy specimens were systematically collected at Xijing Hospital to ensure data integrity and consistency. Each biopsy yielded 12 consecutive sections, with two sections mounted per slide, resulting in six slides per case. Sections were stained using four standard histopathological protocols—Hematoxylin and Eosin (H\&E), Periodic Acid-Schiff (PAS), Masson’s Trichrome (MT), and Periodic Acid-Silver Methenamine (PASM)—in a fixed sequence (H\&E, PAS, MT, PASM, H\&E, PAS, H\&E, PAS, MT, PASM, H\&E, PAS). Supplementary clinical information was integrated, including pathology reports, histological findings, and submission reports with patient history, laboratory tests, and imaging results. In addition, IF images were incorporated to identify antibody depositions within glomeruli.
\item Digitization of Pathology Slides: All slides were digitized using the SQS-600P slide scanning imaging system (Shenzhen Shengqiang Technology Co., Ltd.), producing high-resolution whole-slide images (WSIs) suitable for computational analysis (Figure \ref{light_examples}).
\item Entity Extraction: WSIs were initially downsampled using OpenSdpc to 5$\times$ magnification ($approx$1.68 $\mu$m/pixel) and converted into tensor formats compatible with AI models. A pre-trained glomerular detection model from our previous work \cite{he2024identifying} was used to localize glomeruli within renal tissue, after which glomerular regions were extracted at 20$\times$ magnification ($\approx$ 0.42 $\mu$m/pixel) for subsequent analysis.
\item Lesion Annotation: A dedicated software platform, the Kidney Labeling Tool, was developed to support precise annotation (Figure \ref{kidney_label}). Using this platform, all glomerular lesion annotations in XJ-Light-1 were independently performed by two well-trained nephrologists (Yanxia Wang, and Xiaoqin Wang) following standardized diagnostic criteria. In cases of disagreement, a senior renal pathologist (Jing Li) with extensive diagnostic experience reviewed the discrepant cases and provided adjudicated consensus labels, which were used for model training and evaluation.
\end{itemize}

\textbf{XJ-Light-1} represents the inaugural release of light microscopy images from Xijing Hospital, encompassing CKD cases diagnosed between 2019 and 2021. It includes 870 cases, each represented by six slides, totaling 5,220 slides. Slides were stained with H\&E, PAS, MT, and PASM, capturing complementary structural features of glomeruli and tubulointerstitium. Each case was systematically classified into specific CKD subtypes, including diabetic nephropathy (DN), focal segmental glomerulosclerosis (FSGS), anti-neutrophil cytoplasmic antibody-associated glomerulonephritis (AAGN), membranous nephropathy (MN), obesity-related glomerulopathy (ORGN), anti-glomerular basement membrane disease (AGBM), lupus nephritis (LN), endocapillary proliferative glomerulonephritis (EPGN), membranoproliferative glomerulonephritis (MPGN), tubulointerstitial nephritis (TIN), crescentic glomerulonephritis (CrGN), IgA nephropathy (IgAN), and Henoch–Schönlein purpura nephritis (HSP). XJ-Light-1 serves as a foundational dataset for downstream tasks such as lesion identification, classification, and morphological segmentation (Table \ref{27-lesion-anno}).

\textbf{XJ-Light-2}, by contrast, represents the second collection of light microscopy images from Xijing Hospital, encompassing CKD cases archived between 2022 and 2023. Unlike XJ-Light-1, it was curated chronologically at the slide level without detailed CKD subtype annotation, providing a large-scale resource for representation learning. This dataset comprises 14,049 slides stained with H\&E, PAS, MT, and PASM, and was used exclusively for pretraining pathology foundation models, facilitating the development of robust, transferable feature representations for downstream analytic tasks.

To prevent potential data leakage between pretraining and downstream evaluation, the XJ-Light-1 and XJ-Light-2 datasets were constructed from strictly non-overlapping patient cohorts. All dataset splits were performed at the patient level, ensuring that no slides or specimens from the same patient appear in both datasets. This temporal and patient-level separation guarantees the independence of pretraining and downstream fine-tuning and evaluation.

\subsubsection{XJ-IF}

The XJ-IF dataset comprises 3,198 glomerular IF images collected at Xijing Hospital during routine diagnostic evaluations. Images were acquired by renal subspecialty pathologists at a resolution of 0.84 $\mu$m/pixel (10$\times$ magnification), allowing clear visualization of deposition patterns and regional distributions. Among these, 1,722 images were annotated with deposition region or deposition pattern subclasses, while 1,476 remain unannotated, thus providing both labeled and unlabeled data for diverse analytic tasks (Table \ref{markers}, Table \ref{xjif-anno}). Representative examples are provided in the supplementary materials and illustrated in Figure \ref{if_examples}.

\subsubsection{XJ-GIO}

The XJ-GIO dataset was derived from glomerular images in the XJ-Light-1 cohort and constructed through a hybrid approach combining automated and expert annotation. Initially, the pre-trained Glo-In-One-v2 model \cite{yu2024glo} was applied to segment Bowman’s capsules and glomerular tufts. These preliminary masks were then manually refined by senior annotators to ensure accuracy and consistency. The resulting dataset contains 2,675 high-quality annotations of Bowman’s capsules and tufts, serving as a robust resource for downstream morphological and structural analyses.

\subsubsection{AIDPATH-G}

The AIDPATH-G dataset originates from the AIDPATH project \cite{bueno2020data, bueno2020glomerulosclerosis}, which sought to advance digital pathology tools for glomerulosclerosis assessment. Renal biopsy samples were collected at Hospital Universitari de Girona Dr. Josep Trueta (Spain), Hospital de la Plana (Spain), and the National Pathology Center (Lithuania). All samples were PAS-stained and digitized at 20× magnification, producing 47 whole-slide images (WSIs). From these WSIs, 2,340 glomerular images were extracted, evenly divided between normal (n=1,170) and sclerotic (n=1,170) glomeruli. Glomerular contours were annotated, cropped into individual entities, and stored in PNG format. To increase dataset diversity, augmentation techniques such as rotation, vertical flipping, and color jittering were applied. In this study, we refer to this curated dataset as AIDPATH-G. Representative examples are shown in Figure \ref{AIDPATH_examples}.

\subsubsection{KPMP-G}

The Kidney Precision Medicine Project (KPMP) is a multi-institutional, longitudinal initiative aimed at advancing the mechanistic understanding of chronic kidney disease (CKD) and acute kidney injury (AKI) through integrated clinical, molecular, and digital pathology data. For this study, we leveraged a subset of the KPMP dataset comprising 1,352 renal tissue slides with corresponding clinical metadata. From these slides, 28,242 glomeruli were systematically identified and extracted for downstream analyses, enabling clinicopathological correlation studies. Representative examples from KPMP-G are presented in Figure \ref{kpmp_examples}.

\subsubsection{XJ-CLI}
The XJ-CLI dataset was curated to evaluate the clinical applicability of GloPath. It was derived from the open diagnostic stream at the Department of Pathology, Xijing Hospital, reflecting unfiltered real-world biopsy practice. This dataset comprises 695 PAS-stained renal biopsy slides collected during routine clinical workflows, subsequently digitized into WSIs. Glomerular entities were automatically extracted using an entity detection pipeline, after which lesion annotations were generated following the same consensus-based workflow as that used on XJ-Light-1: two nephrologists (Yanxia Wang, and Xiaoqin Wang) independently annotated each glomerular entity, and a senior renal pathologist (Jing Li) adjudicated discrepant cases to produce the final reference labels. By capturing the heterogeneity and complexity of everyday clinical cases, XJ-CLI provides a stringent, forward-looking testbed for validating model robustness and translational value in real-world clinical environments (Table \ref{xjcli_dataset}).

\subsection{Model Training}

\subsubsection{Problem setting}

Let $\mathcal{D}_{pre} = \{X_i\}_{i=1}^{N}$ be a pretraining large-scale glomerular pathology image dataset, where denotes the first image. The goal of this study is to compute the mapping function $f:\mathcal{X}\rightarrow\mathcal{Z}$, where $\mathcal{X}$ is the image space and $\mathcal{Z}$ is the feature space. In the self-supervised pre-training phase, a training loss function $\mathcal{L}_{pre}$ is defined to train the model to learn useful feature representations from glomerular pathology images, which can be expressed as:
\begin{equation}
\mathcal{L}_{pre}(\theta) = \mathbb{E}_{\mathcal{X}\sim\mathcal{D}_{pre}}[\ell(f_{\theta}(X), g(X))].
\end{equation}
Where $f_{\theta}$ is a pre-trained model with parameter ${\theta}$, $g(X)$ is an auxiliary function used to generate the objective for the self-supervised task, and $\ell$ is a function of the loss of the contrast employed by the self-supervision, which is used to measure the discrepancy between the model output and the objective.

After training is completed, the optimal model $f_{\theta^{*}}$ is obtained, where $\theta^{*} = argmin_{\theta}\mathbb{L}_{pred}(\theta)$. On the downstream task, let $D_{down} = \{(X_i, Y_i)\}_{i=1}^M$ be the dataset used for model performance validation, where $(X_i, Y_i)$ denotes the first sample and its corresponding label. The goal is to fine-tune the model parameters so as to ensure that the model can be adapted to downstream tasks, which can be expressed as:
\begin{equation}
\theta_{down}^{*} = argmin_{\theta}\mathbb{L}_{down}(\theta) = argmin_{\theta}\mathbb{E}_{{x,y}\sim\mathcal{D}_{down}}[\ell^{`}(f_{\theta^{*}}^{`}(x), y)].
\end{equation}
where $f_{\theta^{*}}^{`}$ is the fine-tuned model and $\ell^{`}$ is a loss function that measures the difference between output probability and the ground truth.

\subsubsection{Entity-centric self-supervised pretraining}\label{entity_centric_pretraining}

Glomeruli are intrinsically entity-based structures, in which local components (e.g., endothelial cells, basement membranes, podocytes) function synergistically to maintain blood filtration. From a pathological perspective, localized damage accumulates to impair overall glomerular function. This consistency between global and local semantics provides a biologically grounded rationale for designing entity-centric self-supervised pretraining.

Rather than relying on reconstruction-based methods \cite{he2022masked, bao2021beit, li2021lomar}, we employed contrastive learning \cite{he2020momentum, chen2021exploring, grill2020bootstrap, caron2020unsupervised, oquab2023dinov2}, which has proven effective in computational pathology \cite{koohbanani2021self, wang2021transpath}. Specifically, we adopted the DINO framework, utilizing a teacher-student architecture with exponential moving average (EMA) updates. Both networks share a ViT-Base backbone. The teacher network maintains stable representations by slowly updating its weights using EMA, while applying a sharpening operation to prevent representation collapse, thereby stabilizing the student’s convergence. In the context of GloPath, this self-supervised pretraining allows the model to learn robust, invariant representations of glomerular pathology. By maximizing the similarity between augmented views of the same glomerulus, such as different lesions or whole glomerulus morphology, the model learns to capture fine-grained pathological features that are crucial for accurate diagnosis and disease progression analysis.

During pretraining, our framework introduces multi-scale and multi-view constraints (Figure \ref{overview}b). Random augmentations generate global views (whole-glomerulus morphology) and local views (substructural or lesion-level features), enabling the model to integrate multi-scale pathological cues. Formally, given a glomerulus image $I$, we define two augmentation operators: a global-view augmentation $\mathcal{A}_G$ and a local-view augmentation $\mathcal{A}_L$. The global-view augmentation $\mathcal{A}_G$ preserves the entire glomerular entity and applies stochastic appearance transformations, while the local-view augmentation $\mathcal{A}_L$ additionally incorporates random spatial cropping to capture substructural details using random cropping scales. In contrast to $\mathcal{A}_G$, which operates on the full glomerulus image, $\mathcal{A}_L$ samples a spatial crop with a random scale factor $s \sim \mathcal{U}(s_{\min}, s_{\max})$,. Based on these operators, we generate two global views $\{g^{(1)}, g^{(2)} \}$ via $\mathcal{A}_G(I)$, representing different whole-glomerulus views with varying appearance perturbations, and six local views $\{l^{(k)} \}_{k=1}^{6}$ via $\mathcal{A}_L(I)$, capturing diverse substructural or lesion-level patterns.

Training is guided by a cross-entropy–based contrastive loss:
\begin{equation}
\mathbb{L}{pre} = \frac{1}{2n}\sum{i=1}^{n}\Big(-\sum_{j=1}^{m}\sum_{k=1}^{d}g_{m,k}\log(l_{i,k})\Big),
\end{equation}
where $n$, $m$, and $d$ denote the number of local views, global views, and feature dimensions, respectively. This formulation encourages discriminative representation learning and mitigates feature collapse, yielding embeddings suitable for clinical downstream tasks.

\subsubsection{Model structure}

Given the need for effective feature extraction and classification of glomerular pathology images, this study employs the ViT as the foundational structure for self-supervised pretraining \cite{dosovitskiy2020image}. To accommodate the processing of image data, the ViT model divides the input image into multiple fixed-size patches. These local patches are assigned patch tokens and linearly embedded into a high-dimensional space. For each glomerular pathology image, the classification token (CLS token) is appended, serving as the carrier for the global pathological features of the glomerulus. This study utilizes the ViT-Base model, which comprises 12 Transformer encoder layers, each consisting of multi-head self-attention mechanisms and feed-forward neural networks. 

This architecture enables the model to capture long-range dependencies when processing image data, thereby allowing it to identify complex patterns and structures within glomerular pathology images. Specifically, to enhance the ability to learn features at different scales, the ViT-Base model employs multi-head attention mechanisms. This allows the model to simultaneously focus on both local and global interactions between image patches. Since multi-head attention mechanisms can process features in multiple subspaces in parallel, this study leverages this mechanism to augment the model's capability to capture multi-scale features in glomerular pathology images.

\subsubsection{Parameter setting}

Training was conducted with the following hyperparameters: momentum for EMA encoder = 0.996; temperature = 0.04; mixed-precision (16-bit) training; AdamW optimizer with learning rate $5^{-4}$ (minimum $1^{-6}$) and 10-epoch warm-up. Weight decay was scheduled up to 0.4, with gradient clipping applied. For data augmentation, we adopted global cropping (two entity-level cropping, covering the full glomerulus) and local cropping (six crops; $s_{min} = 0.4$ and $s_{max} = 1$), combined with DropPath (rate 0.1). For data augmentation during pretraining, both $\mathcal{A}_G$ and $\mathcal{A}_L$ include a shared set of photometric and geometric transformations, such as random horizontal and vertical flipping, color jittering (brightness, contrast, saturation, hue), grayscale conversion, and Gaussian blur. Batch size was 64 per GPU, and training was performed for 200 epochs. These settings stabilized convergence and promoted generalization, enabling the pretrained GloPath model to deliver robust feature representations for downstream clinical applications. 

\subsection{Evaluation}
\subsubsection{Benchmark Models}\label{models_being_compared}

To evaluate the parameter specificity and feature representation ability of the proposed GloPath, we benchmarked it against six representative models spanning three categories: (1) non-pathology foundation models, (2) general-purpose pathology foundation models, and (3) a renal pathology domain-specific foundation model.

The non-pathology foundation models include \textbf{RandomInit} and \textbf{ImageNetPre}. Both models are based on the ViT-Base architecture, which comprises 86.8 million parameters. RandomInit applies He initialization for linear layers and truncated normal initialization for positional encodings without pretraining \cite{he2015delving}. ImageNetPre is pretrained on the large-scale ImageNet dataset \cite{deng2009imagenet}, enabling transfer of generic visual features but without pathology-specific adaptation.

Three general-purpose pathology foundation models were selected: UNI, PLIP, and CONCH \cite{chen2024towards, huang2023visual, lu2024visual}.
\begin{itemize}
\item \textbf{UNI} is a ViT-Large–based model (380 million parameters) pretrained with the DINOv2 framework \cite{oquab2023dinov2} on the Mass-100K dataset, comprising more than 100 million tissue patches across 20 tissue types. This breadth enables rich feature learning, though not tailored to renal pathology.
\item \textbf{PLIP} adapts the CLIP framework \cite{radford2021learning} to pathology by jointly training an image encoder and text encoder. Its training set includes 208,414 image–text pairs curated from social media and other sources, supporting multimodal alignment between histopathology images and natural language descriptions.
\item \textbf{CONCH} also leverages contrastive vision–language learning, trained on 1.17 million image–caption pairs spanning H\&E and immunohistochemistry (IHC) slides. This cross-modal setup allows the model to capture associations between morphological patterns and textual descriptors, providing broad but non-specialized pathology representations.
\end{itemize}

To provide a domain-specific baseline, we developed \textbf{RenalPath} using the same ViT-Base architecture and identical self-supervised pretraining strategy as GloPath. Pretraining was conducted on 14,049 renal biopsy WSIs, from which 224$\times$224 image patches were extracted at 0.42 $\mu$m/pixel, yielding 1,017,879 training samples. These patches encompassed diverse renal tissue components, including glomeruli, interstitial regions, and other renal structures, thereby reflecting the heterogeneity of renal pathology. Importantly, the only difference between RenalPath and GloPath lies in the definition of the pretraining unit and data organization: RenalPath follows a conventional patch-based paradigm over heterogeneous renal regions, whereas GloPath is pretrained exclusively on explicitly detected glomerular entities. This design enables a controlled comparison to assess the impact of entity-level pretraining on downstream performance.

\subsubsection{Evaluation metrics}\label{evaluation_metrics}

To ensure rigorous and clinically meaningful evaluation, task-specific metrics were applied according to dataset characteristics and prediction goals. For lesion recognition (Task A) and lesion grading (Task B), class imbalance posed a major challenge. We therefore prioritized the F1 score, which balances precision and recall, as the primary metric. In lesion recognition, we further incorporated the area under the precision–recall curve (PR-AUC) \cite{sofaer2019area} and Receiver Operating Characteristic Curve (ROC-AUC) as a complementary measure. For cross-modality diagnosis (Task C) and few-shot based classification (Task D), the datasets were relatively balanced. Here, the emphasis was on discriminability across classes and ranking ability. Accordingly, the ROC-AUC was adopted as the primary evaluation metric, reflecting the models’ ability to differentiate pathological categories with high sensitivity and specificity.

\subsubsection{Fully supervied fine-tuning}

Tasks A (lesion recognition), B (lesion grading), and C (cross-modality diagnosis) were implemented as full-parameter fine-tuning tasks using a pre-trained glomerular pathology foundation model as the backbone. Task-specific output layers were defined according to clinically relevant categories. For lesion recognition (Task A), the classifier mapped features to binary outputs (lesion vs. non-lesion). For lesion grading (Task B), subclass-specific output nodes reflected established pathological categories, including four outputs for mesangial lesions (normal, mild, moderate, severe) and crescents (normal, cellular, cellular-fibrous, fibrous), and three outputs for membranoproliferative glomerulonephritis and endocapillary proliferation (normal, segmental, global). For cross-modality diagnosis (Task C), fluorescence-based outputs corresponded to capillary vs. mesangial deposition and focal vs. diffuse distribution (two outputs each).

Training was performed with cross-entropy loss and parameter updates by backpropagation. To account for class imbalance in lesion recognition, an imbalanced sampling strategy was applied to balance positive and negative samples within each epoch. To further align the pathology foundation model with fluorescence data, an additional light-weight fine-tuning step was conducted on 1,476 unlabeled IF images prior to formal supervised training. Specifically, starting from the brightfield-pretrained GloPath, we performed a self-supervised adaptation using the same pretraining objective, during which only a subset of higher-level layers was updated while the core representation was largely preserved. This procedure resulted in an IF-finetuned GloPath that reduced modality-induced distribution shifts and enabled more consistent embedding alignment between brightfield and IF images.

All models were optimized with the Adam algorithm and trained for 50 epochs with standard data augmentation (random flipping, rotation, shearing, brightness/contrast adjustment) to enhance generalization. Model selection was based on the highest F1 score achieved on the validation set. To assess robustness, lesion recognition models were trained with three independent random seeds, while lesion subclassification and cross-modality tasks were evaluated with three-fold cross-validation. These procedures minimized evaluation bias and ensured reproducibility.

\subsubsection{Few-shot based fine-tuning}

To further validate the model's parameter specificity for glomerular structures, this chapter conducts few-shot based tasks D, targeting the classification of normal and sclerotic glomeruli based on light microscopy images and the classification of deposits based on IF images. Under this setup, on the one hand, with only a very limited number of samples, the model needs to efficiently grasp the morphological essence of different categories based on the few perceived clues. On the other hand, the entity features of the glomeruli extracted by the model are fixed, and only the parameters of the classification end of the model are updated, which poses high demands on the accuracy of the initial feature construction of the model. First, different models extract the CLS token of the glomerular entity as the global deep semantic feature, which is only used as the input feature in the subsequent classification process and its value is no longer updated. Second, different methods are used to construct the classification end to classify the fixed input features. The experiment was repeated 10 times under different dataset splits. This study constructs five machine learning algorithms to build a mapping from the initial feature space to the specific category space (Figure \ref{overview}c):
\begin{itemize}
\item Support Vector Machine (SVM): The core idea of SVM is to find an optimal hyperplane to separate data of different categories. It uses the radial basis function as the kernel function to enhance the model's nonlinear classification capability.
\item Logistic Regression (LR): LR maps the output of the linear combination to the interval (0,1) through the Sigmoid function to achieve probability estimation. The quasi-Newton method is used to optimize the logistic regression process.
\item Multilayer Perceptron (MLP): MLP is a feedforward neural network composed of multiple hidden layers, which in this study includes hidden layers with 256 and 128 neurons. MLP uses ReLU as the activation function to introduce nonlinearity and Adam as the adaptive learning rate optimization algorithm, with the maximum number of iterations set to 200.
\item Random Forest (RF): RF achieves glomerular classification by constructing multiple decision trees and aggregating their prediction results. This study uses Classification and Regression Trees (CART) to build decision trees and sets the number of decision trees in the forest to 100, which can balance underfitting and overfitting.
\item Prototype Learning (PTL): PTL calculates the feature centers of each category based on the known k samples of each category in the training set, and then determines the category of each sample in the test set according to the distance between each sample and the feature centers of each category. Specifically, the feature center is usually obtained by calculating the mean of the samples of each category. During the test phase, for each test sample, the Euclidean distance between it and the feature centers of each category is calculated, and the sample is assigned to the nearest category. The characteristics of prototype learning are its simplicity, efficiency, and ease of understanding and implementation.
\end{itemize}

Importantly, the datasets used for self-supervised pretraining (XJ-Light-2) and few-shot evaluation (AIDPATH-G and XJ-IF) are fully independent at the patient level. XJ-Light-2 was collected between 2022–2023, whereas XJ-IF was acquired in 2024, and AIDPATH-G is a public dataset from external institutions. No slides or glomeruli from the same patient appear across pretraining and few-shot evaluation, ensuring the absence of data leakage.

\subsubsection{Large-Scale Real-World Study}
To evaluate the clinical utility of GloPath in a real-world diagnostic context, we performed lesion-recognition task on the XJ-CLI dataset. For each lesion category, the model initialized with the best-performing weights from prior experiments was deployed without further fine-tuning, thereby ensuring independence between model development and real-world testing. Inference was then conducted across all cases in the XJ-CLI cohort, which consists of de-identified renal biopsy slides collected during routine diagnostic workflow.

Performance was primarily assessed using ROC-AUC, a metric robust to pronounced class imbalance and particularly informative in open-set scenarios where rare but clinically critical lesions may otherwise be underrepresented. By design, this evaluation simulates the conditions under which nephropathologists encounter heterogeneous biopsy material in daily practice. Importantly, all annotations were generated and verified by senior nephropathologists under institutional review board approval, further ensuring the reliability and translational relevance of the evaluation.

\subsubsection{Morphological segmentation}

The global feature representations learned by GloPath encode both overall and local structural information, which can be leveraged to generate fine-grained semantic segmentation of glomerular compartments. In this study, we adopted the Segmenter framework \cite{strudel2021segmenter}, which comprises an encoder and a decoder. The encoder employs the backbone of GloPath (and other models under comparison), with parameters frozen during training to retain the pre-trained feature representations. The decoder follows a MaskTransformer design, capturing global dependencies via multi-head self-attention and applying nonlinear transformations through feed-forward networks. A masking mechanism is applied to constrain attention to specific pixel regions, enhancing the focus on local structural details. In our experiments, the decoder was configured with 12 layers and 12 attention heads, and the model and feed-forward network dimensions were set to 192 and 768, respectively. This configuration allows Segmenter to effectively integrate global context and local features, dynamically weighting them to achieve precise segmentation of complex glomerular structures. Training was performed with a batch size of 32 using the Adam optimizer (learning rate = $1^{-4}$; weight decay = $1^{-4}$) for up to 100 epochs. 

In addition to comparing Segmentors based on different weights, we also compared GloPath with nnU-Net, the state-of-the-art method for medical image segmentation \cite{isensee2021nnu}. To ensure experimental fairness, nnU-Net adopted the exact same dataset split and training/validation strategy as GloPath, without introducing any additional prior information or manual parameter tuning. All model architecture configurations, preprocessing pipelines, and training hyperparameters were automatically determined by nnU-Net's automated planning and configuration mechanism, strictly following its official recommended settings. This ensures that the comparison reflects the intrinsic performance of a specialized supervised segmentation framework rather than benefits from task-specific manual optimization.

The model achieving the highest intersection-over-union (IoU) on the validation set was selected for performance evaluation. To assess robustness and reproducibility, three independent training runs were conducted with identical dataset splits but different random seeds.

\subsubsection{Clinicopathological correlation analysis}

Morphological alterations in glomeruli, including glomerulosclerosis, mesangial proliferation, and podocyte injury, reflect underlying renal pathophysiology and directly influence kidney filtration and metabolic waste clearance. These structural features establish a clinically relevant link between renal histopathology and patient characteristics. In this study, we examined clinicopathological correlations using morphological segmentation derived from GloPath, RenalPath, and UNI. Clinical variables for the XJ-Light-1 and KPMP-G cohorts are summarized in Tables \ref{clinic_private} and \ref{clinic_public}, while pathological indicators were constructed via semantic segmentation of Bowman's capsule and glomerular tuft, entity-level parameter quantification, and case-level aggregation (Table \ref{morphology}).

Segmentor models based on GloPath, RenalPath, and UNI were applied to generate Bowman's capsule and tuft masks for each glomerulus. From these masks, seven glomerular-level parameters were computed, and their median and mean values across all glomeruli in a case were used to obtain 14 aggregated case-level features per sample.

To evaluate the relationships between clinical and pathological indicators, we employed non-parametric statistical tests. The Kolmogorov–Smirnov (KS) test was used to compare the distributions of two independent samples, quantifying differences in their empirical cumulative distribution functions. The Kruskal–Wallis (KW) test was used to assess differences among three or more independent samples based on rank sums. Pathological parameters were treated as continuous variables, while clinical variables were categorized according to relevant clinical criteria. Importantly, because both the KS and KW tests are non-parametric, they do not require the data to satisfy normality assumptions or homogeneity of variance, making them suitable for clinical and pathological datasets that may be skewed or contain outliers. Statistical significance for each pairwise comparison was determined based on the corresponding p-values. The quantified effect sizes using epsilon-squared ($\epsilon^2$) for the KW test and the KS statistic (D) for the KS test are reported to interprete the magnitude of clinicopathological associations.

In addition, to account for potential confounding factors, patient-level multivariate regression analyses were performed in the XJ-Light-1 cohort. For each patient, GloPath-derived glomerular morphological features were aggregated across all detected glomeruli using summary statistics. Three clinically relevant outcomes were analyzed: creatinine as a continuous variable, and lesion severity categorized as non-severe versus severe.
Multivariate linear regression was used for continuous outcomes, while multivariate logistic regression was applied for binary outcomes. All models included sex, age, and disease type (IgAN, LN, MN, DN, and MCD) as covariates. Regression analyses were implemented using standard Python statistical libraries with default settings. Statistical significance was assessed using two-sided tests.

\subsubsection{Visualization}

To evaluate the interpretability of GloPath and the rationality of its predictions, attention score and feature embedding visualizations were performed. For attention visualization, the multi-head self-attention mechanism in the ViT-Base backbone was utilized. Attention weights for each patch token were aggregated across all heads and visualized according to their spatial positions, highlighting regions of the glomerulus that contribute most to the model’s decision. This approach provides insight into GloPath’s focus on lesion-related structures and supports the biological plausibility of its predictions. These visualizations provide saliency-level interpretability by highlighting diagnostically relevant regions, rather than explicit semantic reasoning over predefined pathological concepts. Future work may explore concept-level interpretability by integrating human-defined pathological concepts or structured lesion annotations, enabling a more explicit linkage between model decisions and established diagnostic criteria. 

To further assess the discriminability of learned features across lesion categories, t-distributed stochastic neighbor embedding (t-SNE) was applied to the high-dimensional feature representations extracted from the CLS token. t-SNE reduces these features to a two-dimensional space, preserving the relative similarity between samples and enabling intuitive visualization of clustering patterns. This method allows qualitative evaluation of whether the model captures meaningful distinctions among different lesion types. All visualizations were generated using standard Python packages, ensuring reproducibility and clarity of the results.

To qualitatively assess cross-modality feature alignment, feature embeddings extracted from the CLS token were projected into a two-dimensional space using UMAP \cite{mcinnes2018umap} approach with 50 nearest neighbors and a minimum inter-point distance of 0.1. This configuration balances local continuity with global structure preservation.

\subsection{Statistical Analysis}

All statistical analyses were conducted to ensure the scientific rigor and reliability of the reported results. For lesion assessment, paired measurements were analyzed using the two-tailed Wilcoxon signed-rank test, a non-parametric method suitable for assessing differences between related samples without assuming normality. A p-value less than 0.05 was considered indicative of statistical significance.

For clinicopathological correlation analysis, all tests were performed in Python (version 3.12.4) using the scipy.stats library. The choice of statistical tests was based on data distribution characteristics and standard methodological assumptions; non-parametric tests were applied when normality assumptions were not met. Continuous variables were first assessed for normality using the Shapiro-Wilk test. When distributions deviated from normality, non-parametric tests were employed. Specifically, the KS test was used to evaluate differences in the empirical distributions of continuous pathological parameters between two groups, while the KW test was applied to compare multiple independent groups. For post-hoc pairwise comparisons following the KW test, Dunn’s test with Bonferroni correction was used to control for type I error. Sample size for each statistical analysis is reported in the corresponding figure legends and tables.

All statistical tests were two-sided, with a significance threshold set at $\alpha$ = 0.05. Exact p-values are reported wherever applicable. Summary statistics are presented as mean $\pm$ standard deviation for normally distributed variables, or median with interquartile range for non-normally distributed variables. This approach ensures transparency, reproducibility, and robust interpretation of the associations between model-derived pathological features and clinical variables.

\subsection{Computing harware and software}

All experiments were implemented using PyTorch (v2.5.0) with CUDA 12.4 (https://pytorch.org
). Pretraining was performed on eight NVIDIA A100 GPUs (80 GB each) using a distributed data parallel configuration for multi-GPU, multi-node training. Downstream validation experiments utilized four NVIDIA GeForce RTX 3090 GPUs (20 GB each) and eight NVIDIA A6000 GPUs (40 GB each).

Whole-slide images in SDPC format were processed using the openSdpc library (https://github.com/WonderLandxD/opensdpc). Comparative models, including UNI, PLIP, and CONCH, were obtained from their respective public repositories. Image processing tasks were conducted using OpenCV-Python (v4.8.0) and Pillow (v9.5.0), while data manipulation was performed with NumPy (v1.22.4) and Pandas (v1.4.3). Visualization was achieved using Matplotlib (v3.7.2), Seaborn (v0.12.1), and OriginLab 2025.

For model construction, timm (v0.9.7; https://huggingface.co) was employed for classification networks, and the Segmenter library (https://github.com/rstrudel/segmenter) and nnU-Netv2 library(https://github.com/MIC-DKFZ/nnUNet) were used for segmentation networks. Clinical-pathological correlation analyses and machine learning-based classifiers (SVM, LR, MLP, and RF) were implemented using scikit-learn (v1.3.2). The multivariate regression analyses were performed using python-statsmodels (https://pypi.org/project/statsmodels/). All software versions and hardware configurations are reported to ensure reproducibility.

\section*{Acknowledgements}

We acknowledge the AIDPATH project for providing access to digital pathology resources that facilitated this study. We also thank the Kidney Precision Medicine Project (KPMP), which is funded by the grants from the NIDDK, for providing freely available clinical and histology slide data upon which parts of the results in this study are based. The work was supported by National Natural Science Foundation of China (NSFC) (82430062), the Shenzhen Engineering Research Centre (XMHT20230115004), and Tsinghua Shenzhen International Graduate School Cross-disciplinary Research and Innovation Fund Research Plan (JC2024002). The work was also supported by Key Research and Development Program of Shaanxi Province (2024GX-YBXM-123). We also thank the Jilin FuyuanGuan Food Group Co., Ltd.

\section*{Author contributions}

Qiming He, Jing Li, Tian Guan, Chao He, Zhe Wang, and Yonghong He conceived and designed the research project and experiments.
Chao He, Zhe Wang, and Yonghong He coordinated the project and provided overall guidance.
Qiming He, Jing Li, and Tian Guan drafted the manuscript.
Qiming He performed data processing and model construction.
Jing Li was responsible for dataset construction and annotation, as well as result analysis.
Tian Guan completed the polishing of the manuscript.
Shuang Ge, Yifei Ma, and Zimo Zhao contributed to the development of part of the code and analysis of experimental results.
Yanxia Wang and Hongjing Chen completed dataset construction and partial annotation.
Yingming Xu, Yizhi Wang, Xinrui Chen, Lianghui Zhu, Junru Cheng, Yexing Zhang, and Yiqing Liu assisted with some experiments and organized the code.
Qingxia Hou, Shuyan Zhao, Xiaoqin Wang, and Lili Ma aided in dataset curation.
Qiang Huang, Zihan Wang, Zhiyuan Shen, Siqi Zeng, Peizhen Hu, Jiurun Chen, and Zhen Song provided equipment and resource support.

\section*{Competing interests}

The authors declare no competing interests.

\section*{Data availability}

The data that support the findings of this study are available on request from the corresponding author. The data are not publicly available due to privacy or ethical restrictions. The public external image and clinical data used in this study are available in the KPMP (\href{https://atlas.kpmp.org/repository}{https://atlas.kpmp.org/repository}) and AIDPATH (\href{https://aidpath.eu}{https://aidpath.eu}).

\section*{Code availability}

Code and model weights for GloPath can be accessed for academic research purposes at 

\href{https://github.com/jiangnansss/Code-for-GloPath}{https://github.com/jiangnansss/Code-for-GloPath}. 

\section*{Supplementary information}
See  Fig. S1-S14 and Table S1 - S49 and for details.

\bibliographystyle{MSP}
\bibliography{sn-bibliography}

\appendix
\newpage
\clearpage
\section*{Supplementary materials}
\setcounter{figure}{0}
\setcounter{table}{0}
\renewcommand{\thefigure}{S\arabic{figure}}
\renewcommand{\thetable}{S\arabic{table}}
\makeatletter
\let\theHfigure\thefigure
\makeatother

\subsection*{Menu of supplementary content}
\setlength{\parindent}{0pt}

\subsubsection*{1. Supplementary Figures}

\begin{figure}[H]
  \centering
  \includegraphics[width=0.8\textwidth]{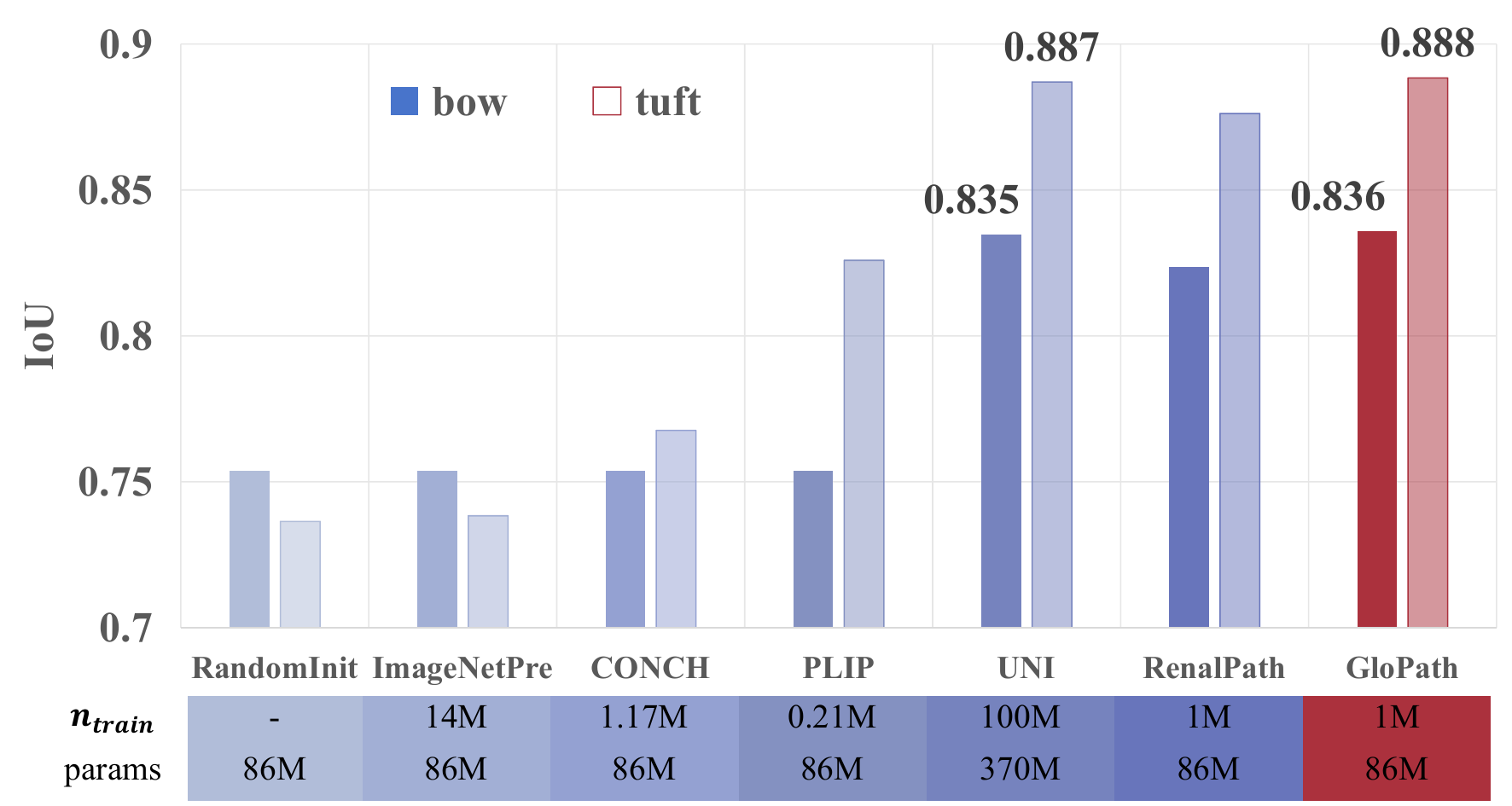}
  \caption{Comparison of IoU of all the models on morphological segmentation, which shows the results of bow segmentaion on XJ-Light and tuft segmentation on XJ-GIO.} \label{segmentation}
\end{figure}

Fig. \ref{segmentation} Comparison of IoU of all the models on morphological segmentation, which shows the results of bow segmentation on XJ-Light and tuft segmentation on XJ-GIO.

\begin{figure}[H]
  \centering
  \includegraphics[width=0.8\textwidth]{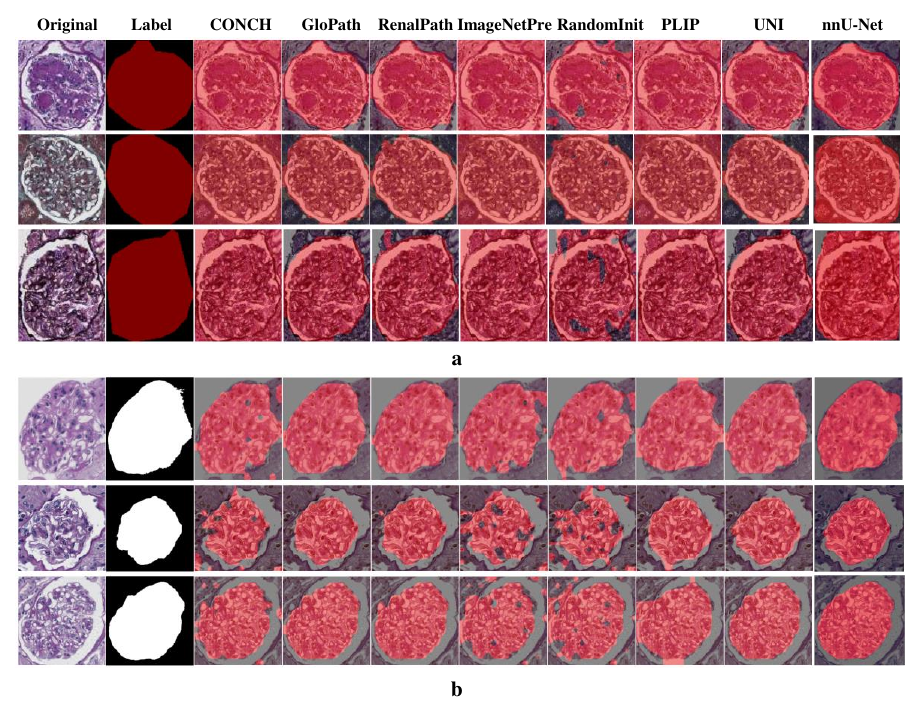}
  \caption{Comparison of segmentation visulization of all the models. \textbf{a, }Performance of bow segmentaion. Column 1-10 indicate original iamegs, ground truth, results of CONCH, GloPath, RenalPath, ImageNetPre, RandomInit, PLIP, UNI and nnU-Net. The masks represent the target areas. \textbf{b, }Performance of tuft segmentaion. } \label{segmentation_vis}
\end{figure}

Fig. \ref{segmentation_vis} Comparison of segmentation visualization of all the models. a, Performance of bow segmentation. Column 1-9 indicate original images, ground truth, results of CONCH, GloPath, RenalPath, ImageNetPre, RandomInit, PLIP and UNI. The masks represent the target areas. b, Performance of tuft segmentation.

\begin{figure}[H]
  \centering
  \includegraphics[width=\textwidth]{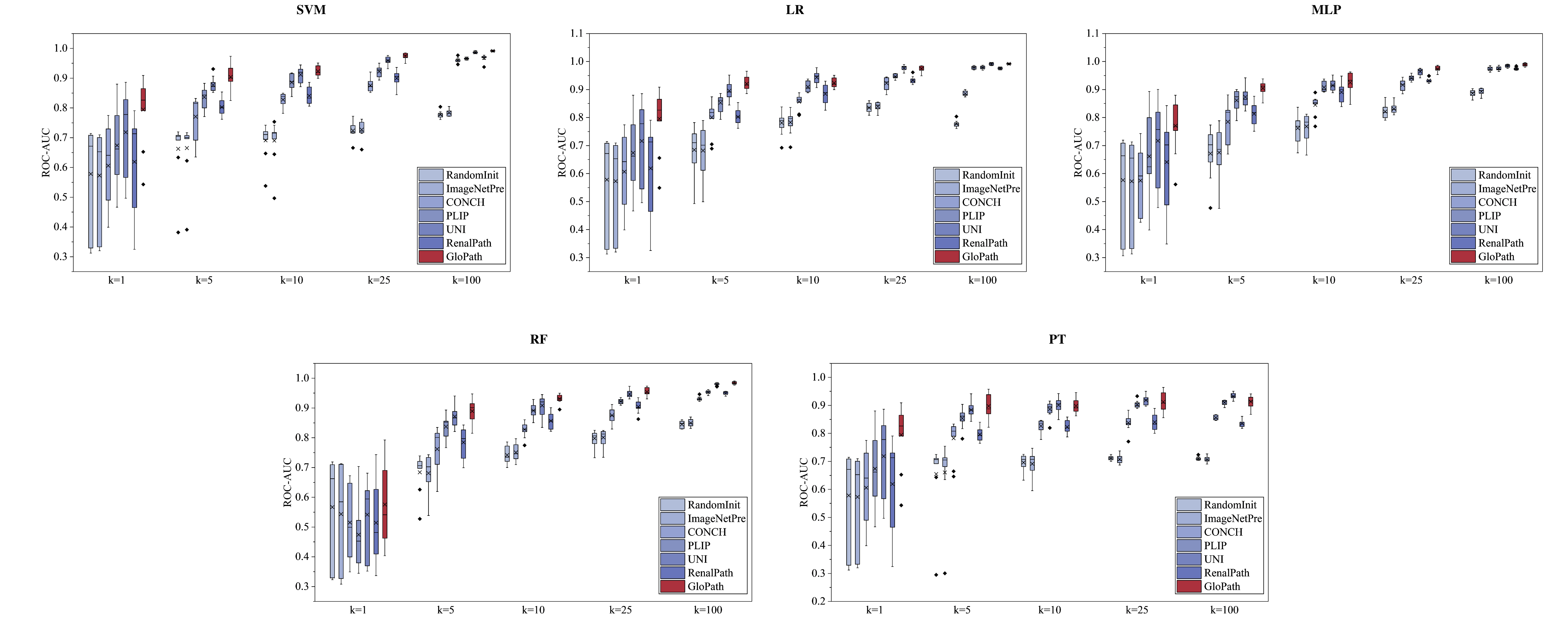}
  \caption{Performance of all the compared models on few-shot leanring on AIDPATH-G with confidence intervals.} \label{few-aid-conf}
\end{figure}

Fig. \ref{few-aid-conf} Performance of all the compared models on few-shot leanring on AIDPATH-G with confidence intervals.

\begin{figure}[H]
  \centering
  \includegraphics[width=\textwidth]{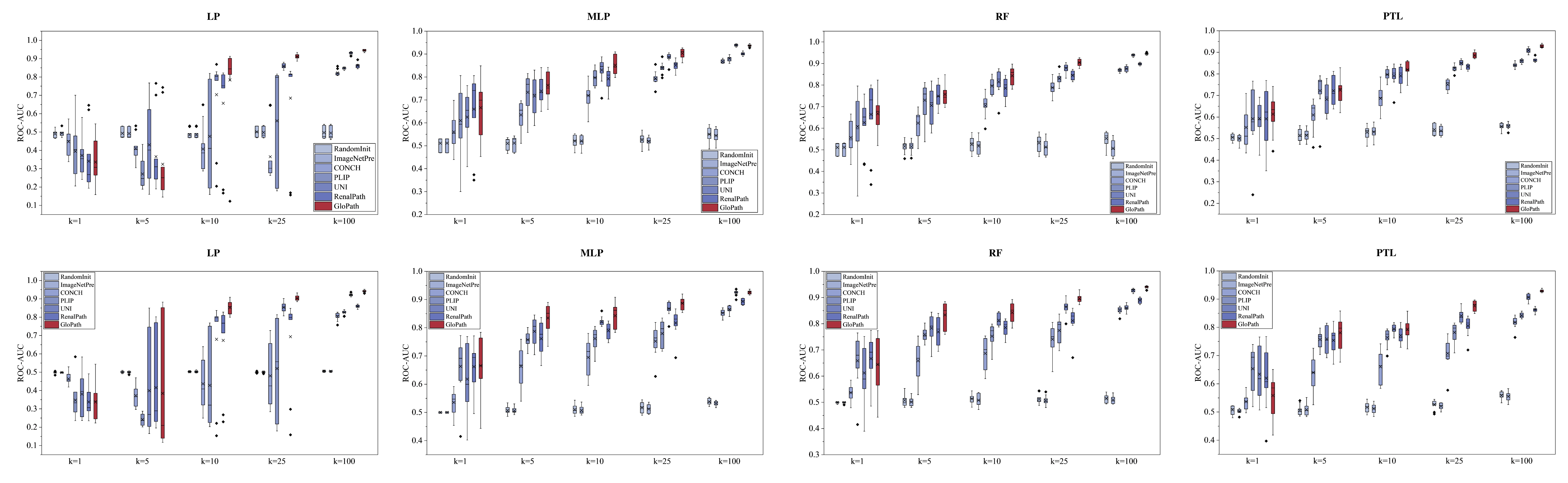}
  \caption{Performance of all the compared models on few-shot leanring on XJ-IF with confidence intervals. Row 1-2 indicate region and pattern classification tasks respectively.} \label{few-if-conf}
\end{figure}

Fig. \ref{few-if-conf} Performance of all the compared models on few-shot leanring on XJ-IF with confidence intervals.

\begin{figure}[H]
  \centering
  \includegraphics[width=\textwidth]{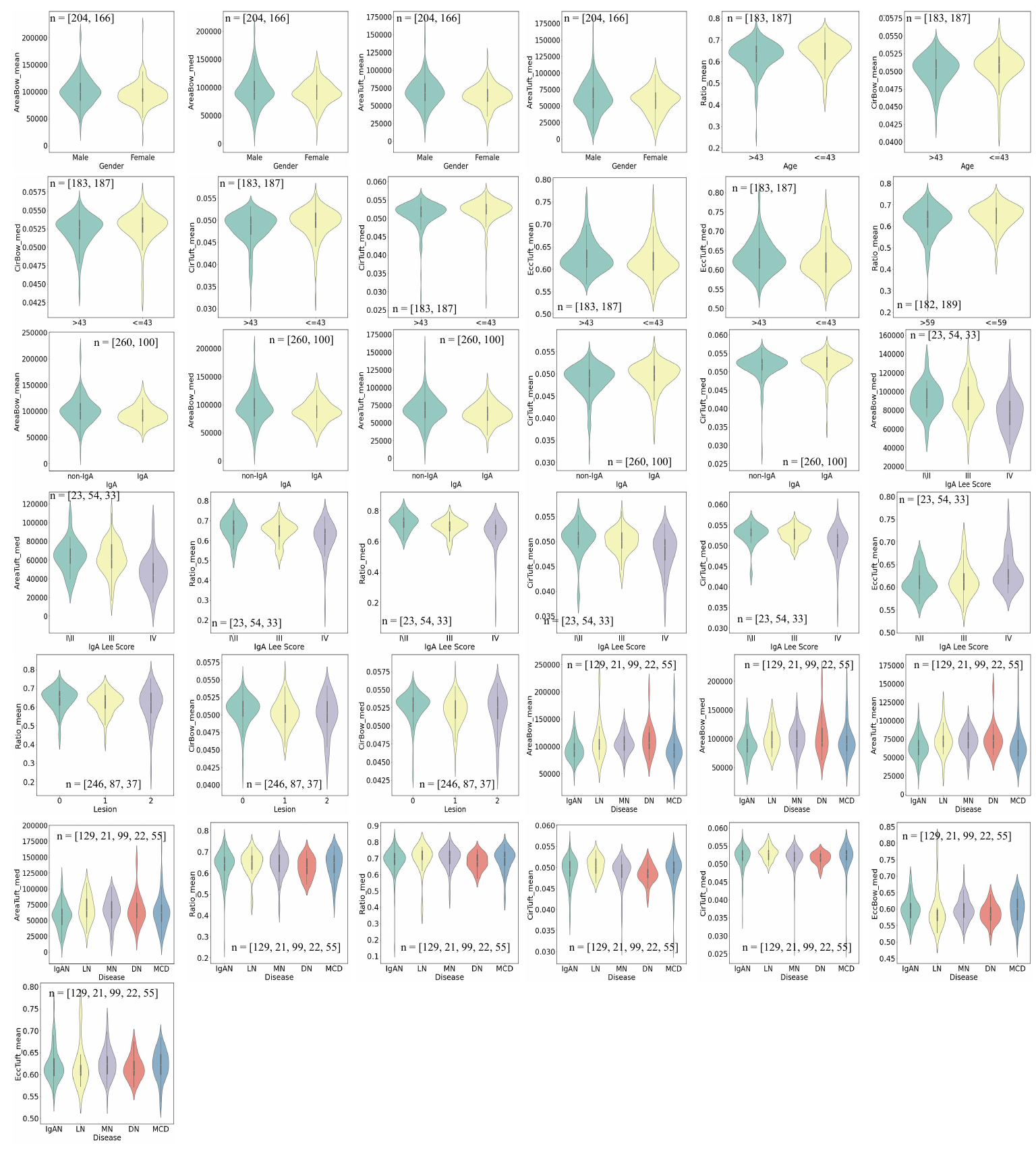}
  \caption{Clinicopathological correlation analysis on XJ-Light-1. The x-axis represents clinical parameters and their respective groups, while the y-axis shows the statistical distribution of pathological phenotypes within each group. Exact P values are provided in Table S16-S22.} \label{omics_private}
\end{figure}

Fig. \ref{omics_private} Clinicopathological correlation analysis on XJ-Light-1. The x-axis represents clinical parameters and their respective groups, while the y-axis shows the statistical distribution of pathological phenotypes within each group.

\begin{figure}
  \centering
  \includegraphics[width=\textwidth]{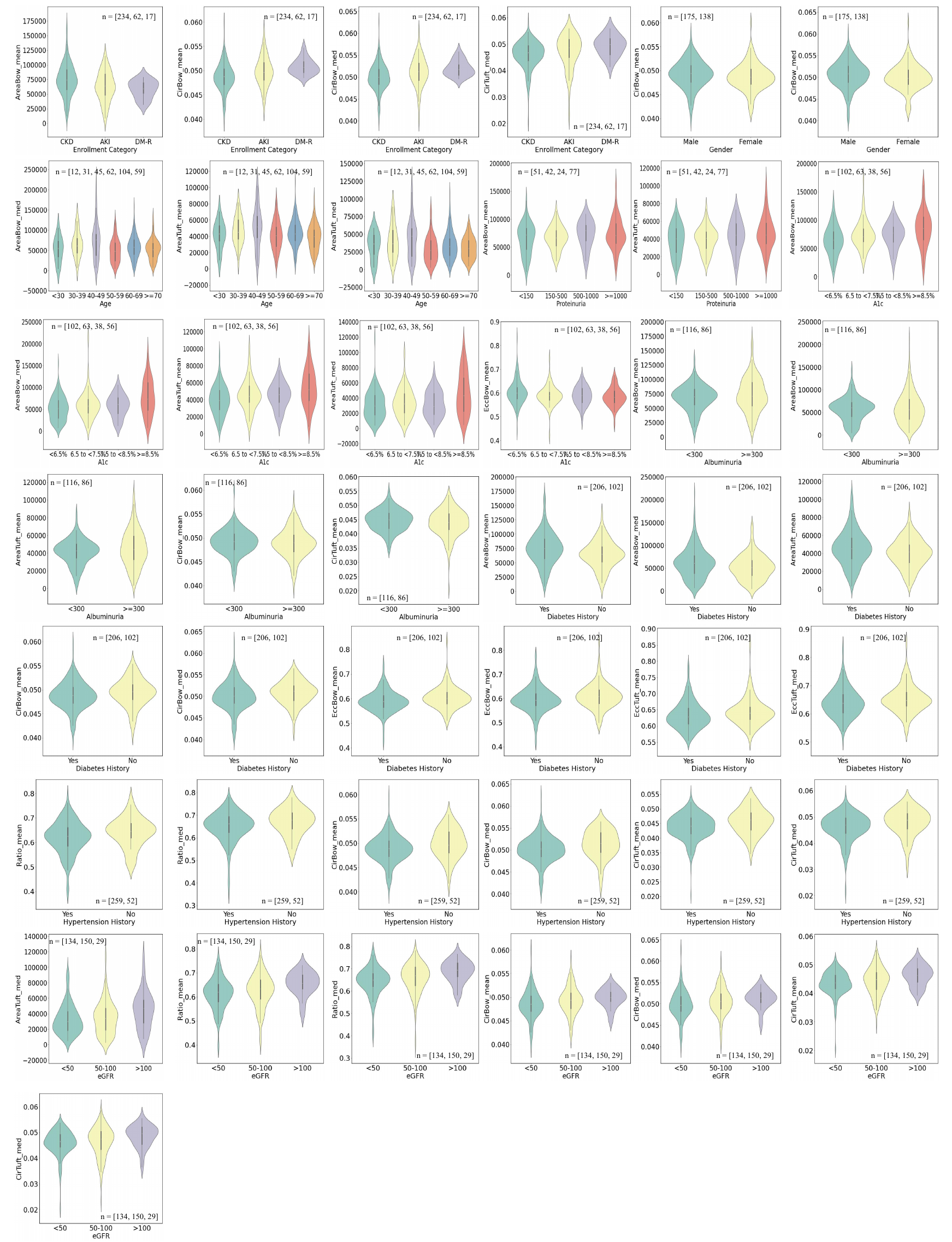}
  \caption{Clinicopathological correlation analysis on KPMP-G. The definitions of the x-axis and y-axis are the same as in Fig. \ref{omics_private}.  Exact P values are provided in Table S25-S33.} \label{omics_public}
\end{figure}

Fig. \ref{omics_public} Clinicopathological correlation analysis on KPMP-G. The definitions of the x-axis and y-axis are the same as in Fig. S\ref{omics_private}.

\begin{figure}
  \centering
  \includegraphics[width=0.6\textwidth]{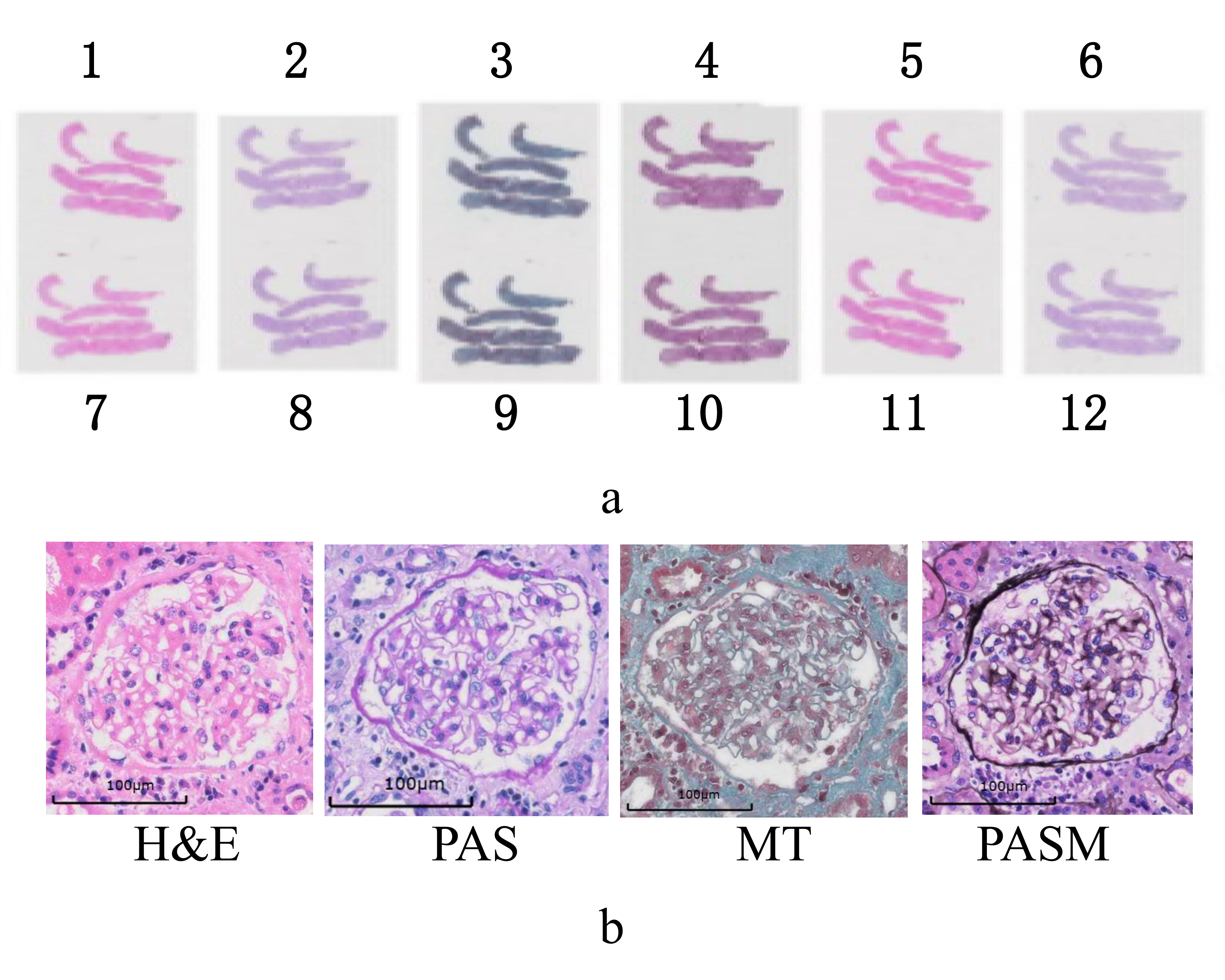}
  \caption{Examples of the XJ-Light dataset. \textbf{a, }Examples of slides from 12 levels. \textbf{b, }Examples of glomeruli from four stainings.} \label{light_examples}
\end{figure}

Fig. \ref{light_examples} Examples of the XJ-Light dataset. a, Examples of slides from 12 levels. b, Examples of glomeruli from four stainings.

\begin{figure}
  \centering
  \includegraphics[width=0.8\textwidth]{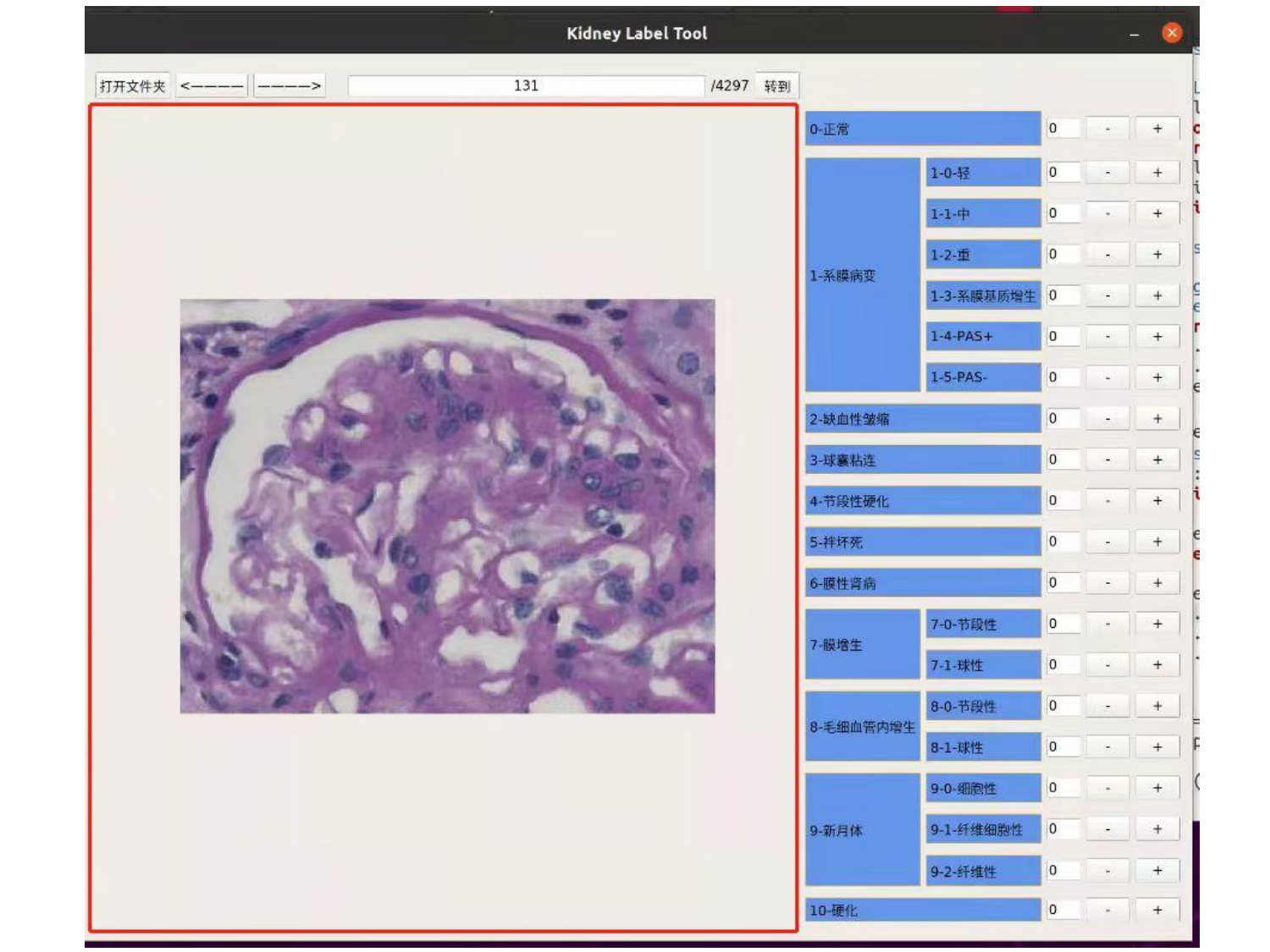}
  \caption{Interface of the Kidney Label Tool which is used for lesion annotation.} \label{kidney_label}
\end{figure}

Fig. \ref{kidney_label} Interface of the Kidney Label Tool which is used for lesion annotation.

\begin{figure}
  \centering
  \includegraphics[width=0.8\textwidth]{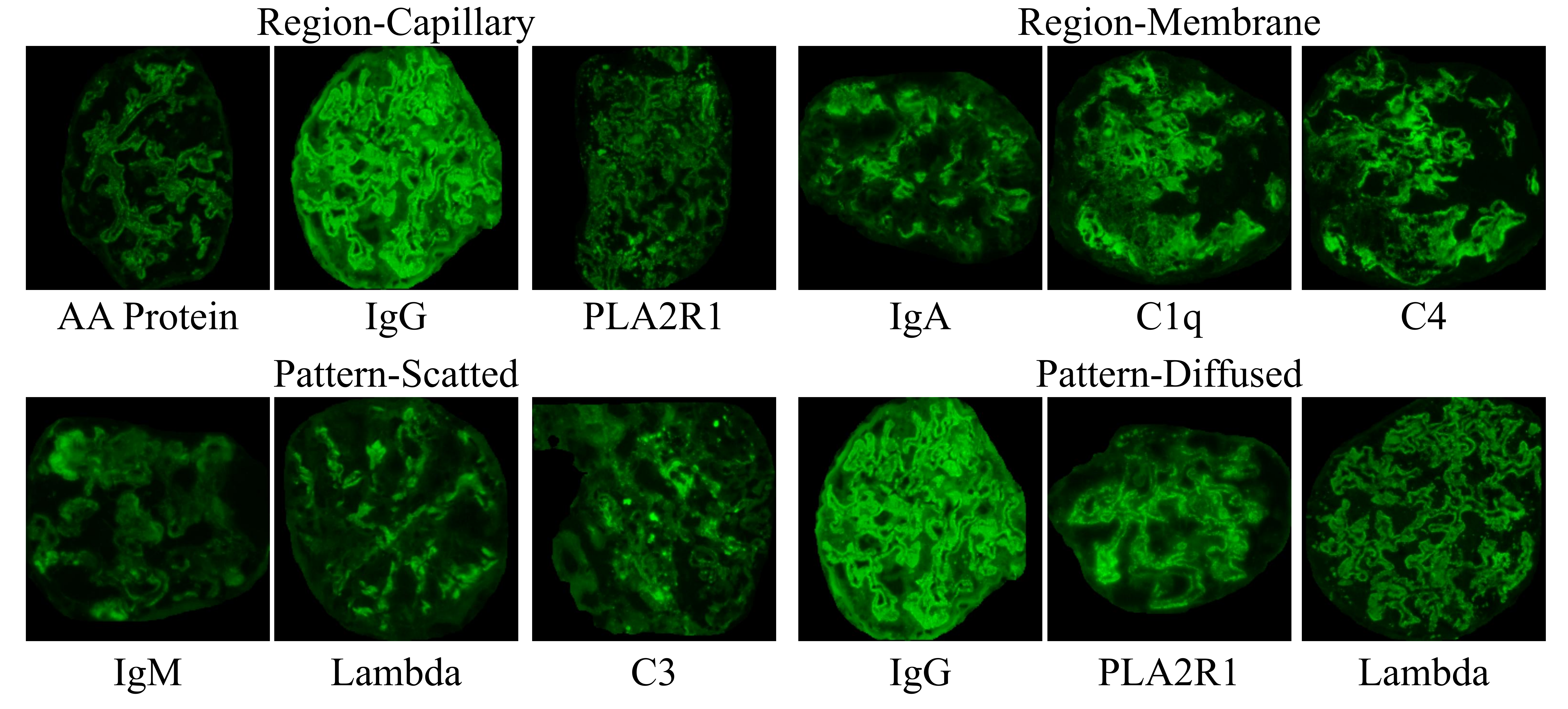}
  \caption{Examples of the XJ-IF dataset with different deposition regions and patterns and with diverse markers.} \label{if_examples}
\end{figure}

Fig. \ref{if_examples} Examples of the XJ-IF dataset with different deposition regions and patterns and with diverse markers.

\begin{figure}
  \centering
  \includegraphics[width=\textwidth]{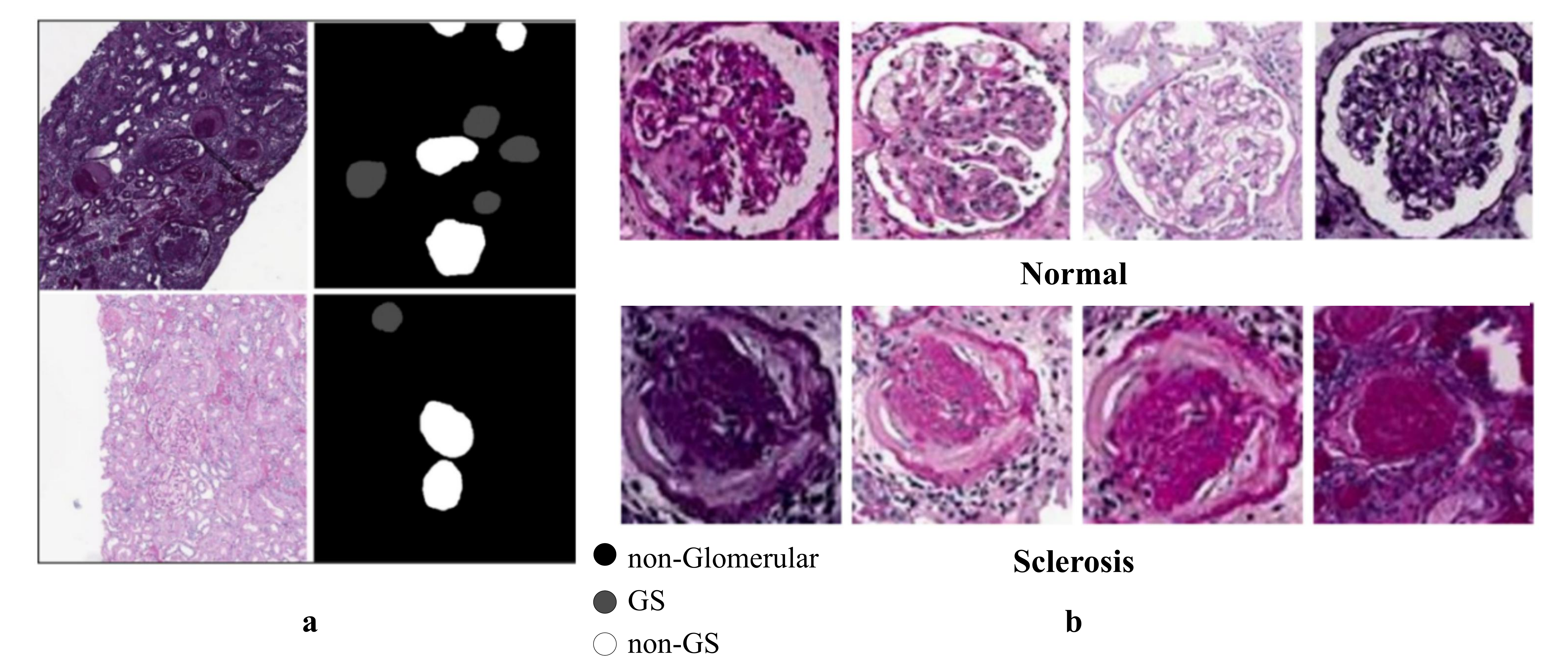}
  \caption{Examples of the AIDPATH-G dataset. \textbf{a, } Renal pathology images and their annotations. \textbf{b, } Examples of normal and sclerotic glomeruli.} \label{AIDPATH_examples}
\end{figure}

Fig. \ref{AIDPATH_examples} Examples of the AIDPATH-G dataset. a, Renal pathology images and their annotations. b, Examples of normal and sclerotic glomeruli.

\begin{figure}
  \centering
  \includegraphics[width=\textwidth]{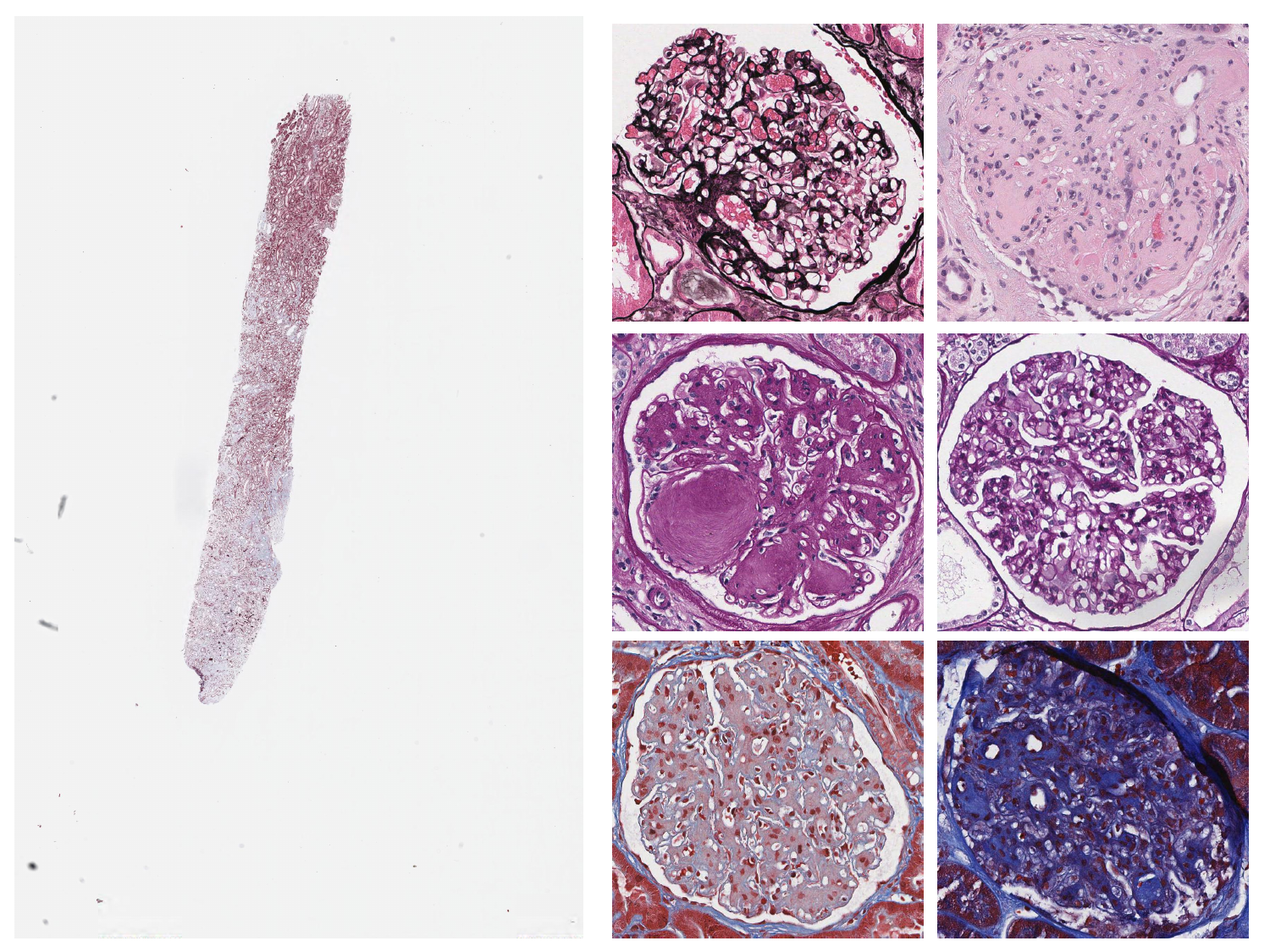}
  \caption{Examples of slides and extracted glomeruli with different stainings in the KPMP-G dataset.} \label{kpmp_examples}
\end{figure}

Fig. \ref{kpmp_examples} Examples of slides and extracted glomeruli with different stainings in the KPMP-G dataset.

\begin{figure}
  \centering
  \includegraphics[width=0.8\textwidth]{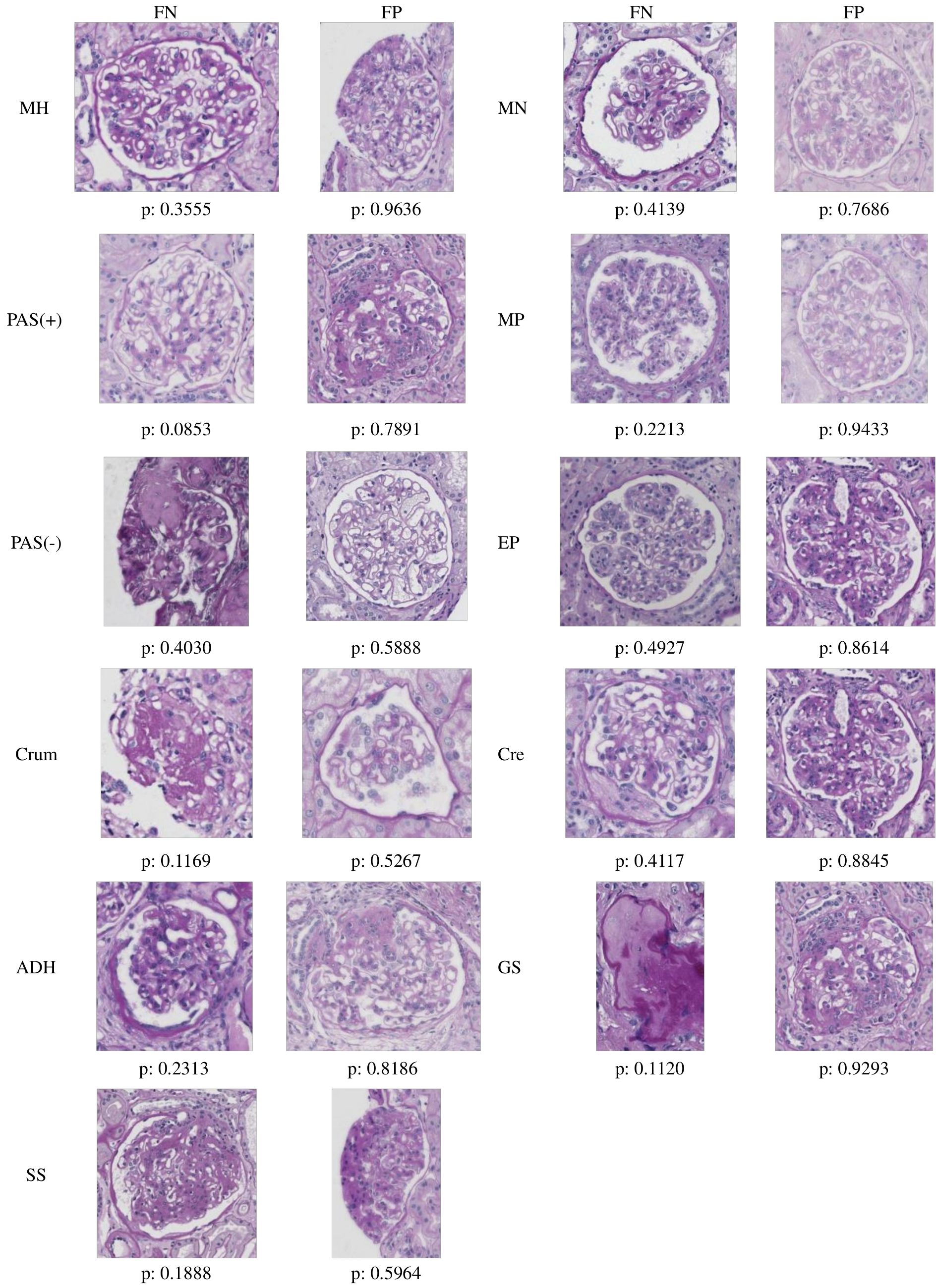}
  \caption{Error analysis of GloPath on XJ-CLI. For each lesion type, we present one false negative and one false positive example, with GloPath's predicted probability for the lesion displayed below each glomerular image (a probability $p > 0.5$  indicates the presence of the lesion).} \label{error_analysis}
\end{figure}

Fig. \ref{error_analysis} Error analysis of GloPath on XJ-CLI.

\begin{figure}
  \centering
  \includegraphics[width=\textwidth]{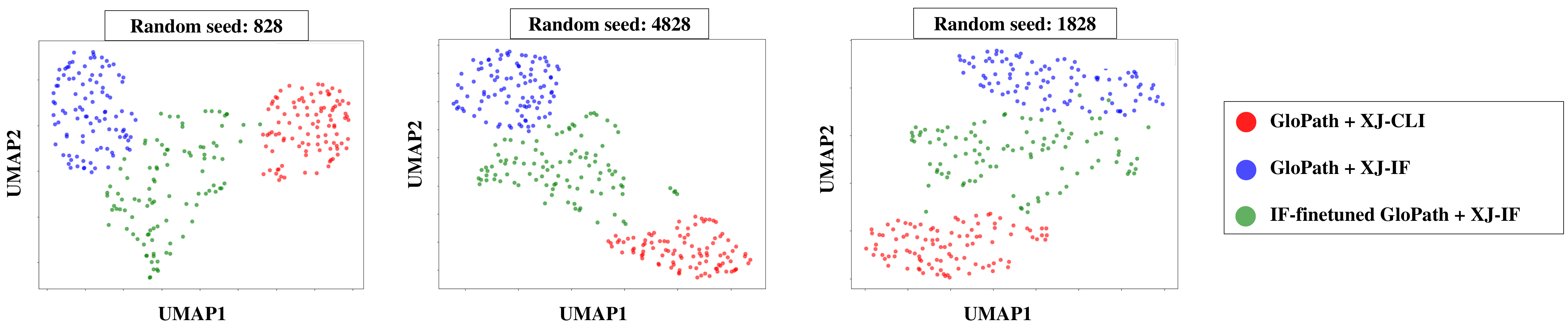}
  \caption{UMAP projection across different weights and imaging domains.} \label{umap}
\end{figure}

Fig. \ref{umap} UMAP projection across different weights and imaging domains.

\begin{figure}
  \centering
  \includegraphics[width=\textwidth]{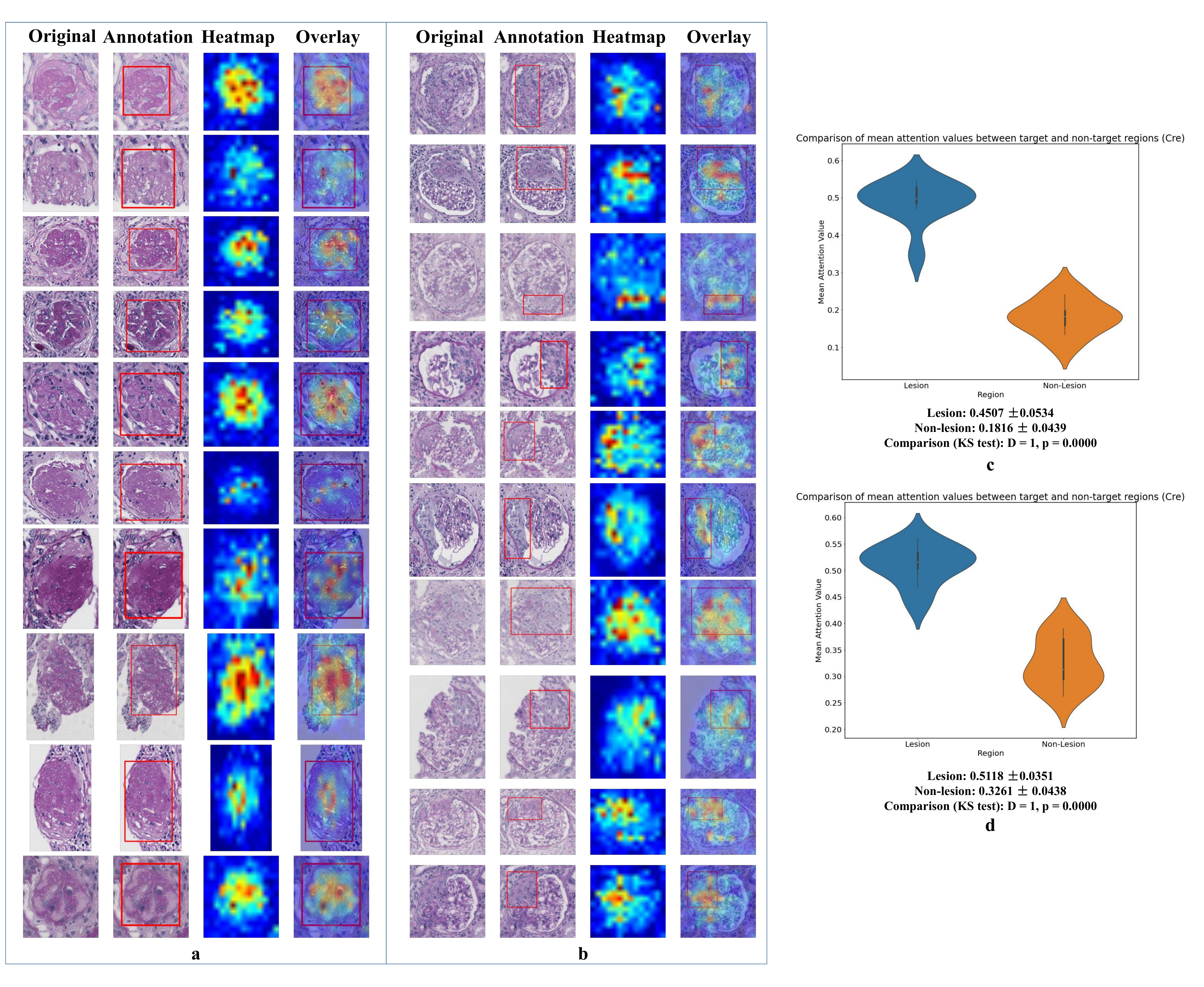}
  \caption{Interpretability study of GloPath. \textbf{a, }Comparison between heatmaps generated from GloPath predictions and annotations by nephrologists (GS). Columns 1-4 indicate original images, original images with annotations, heatmaps, and overlays. \textbf{b, }Comparison between heatmaps generated from GloPath predictions and annotations by nephrologists (Cre). \textbf{c-d, }Comparison of mean attention values between target (lesion) and non-target (non-lesion) regions for GS and Cre, respectively. KS-test results (KS D and p-value) are shown below the violin plots.} \label{interpre}
\end{figure}

Fig. \ref{interpre} Interpretability study of GloPath.

\subsubsection*{2. Supplementary Tables}

Table \ref{54tasks} Details of the 52 lesion assessment tasks.

Table \ref{xjcli_dataset} Number of glomeruli with each type of lesion in XJ-CLI.

Table \ref{pas_27task} Comparison of methods on lesion recognition in PAS staining (F1 score). The bold value indicates the best model.

Table \ref{mt_27task} Comparison of methods on lesion recognition in MT staining (F1 score). The bold value indicates the best model.

Table \ref{pasm_27task} Comparison of methods on lesion recognition in PASM staining (F1 score). The bold value indicates the best model.

Table \ref{aid_linear} Comparison of methods on AIDPATH-G under full supervision. The bold value indicates the best model.

Table \ref{aid_svm} Comparison of methods on AIDPATH-G using SVM-based few-shot learning. The bold value indicates the best model.

Table \ref{aid_lr} Comparison of methods on AIDPATH-G using LR-based few-shot learning. The bold value indicates the best model.

Table \ref{aid_mlp} Comparison of methods on AIDPATH-G using MLP-based few-shot learning. The bold value indicates the best model.

Table \ref{aid_rf} Comparison of methods on AIDPATH-G using RF-based few-shot learning. The bold value indicates the best model.

Table \ref{aid_tpl} Comparison of methods on AIDPATH-G using PTL-based few-shot learning. The bold value indicates the best model.

Table \ref{grading_10task} Comparison of methods on lesion grading (ROC-AUC). The bold value indicates the best model.

Table \ref{seg} Comparison of methods on semantic segmentation of the glomerular bow and tuft. The bold value indicates the best model.

Table \ref{sig-level-private} Comparison of significance level of pathomics-data mining of XJ-Light-1 for GloPath, UNI, and RenalPath.

Table \ref{sig-level-public} Comparison of significance level of pathomics-data mining of KPMP-G for GloPath, UNI, and RenalPath.

Table \ref{gender-private} Significance of clinicopathological correlation analysis for Gender on XJ-Light-1 (p-value).

Table \ref{age-private} Significance of clinicopathological correlation analysis for Age on XJ-Light-1 (p-value).

Table \ref{creatinine-private} Significance of clinicopathological correlation analysis for Creatinine on XJ-Light-1 (p-value).

Table \ref{IgA2-private} Significance of clinicopathological correlation analysis for IgA (Binary) on XJ-Light-1 (p-value).

Table \ref{IgALee-private} Significance of clinicopathological correlation analysis for IgA Lee Score on XJ-Light-1 (p-value).

Table \ref{disease-private} Significance of clinicopathological correlation analysis for Disease on XJ-Light-1 (p-value).

Table \ref{lesion-private} Significance of clinicopathological correlation analysis for Lesion on XJ-Light-1 (p-value).

Table \ref{multi_vari_creatine} Multivariate Regression Analysis of Pathology Phenotypes Associated with Creatinine.

Table \ref{multi_vari_disease} Multivariate Regression Analysis of Pathology Phenotypes Associated with Disease Type.

Table \ref{enrollment-public} Significance of clinicopathological correlation analysis for Enrollment Category on KPMP-G (p-value).

Table \ref{gender-public} Significance of clinicopathological correlation analysis for Gender on KPMP-G (p-value).

Table \ref{age-public} Significance of clinicopathological correlation analysis for Age on KPMP-G (p-value).

Table \ref{proteinuria-public} Significance of clinicopathological correlation analysis for Proteinuria on KPMP-G (p-value).

Table \ref{a1c-public} Significance of clinicopathological correlation analysis for A1c on KPMP-G (p-value).

Table \ref{albuminuria-public} Significance of clinicopathological correlation analysis for Albuminuria on KPMP-G (p-value).

Table \ref{diabetes-public} Significance of clinicopathological correlation analysis for Diabetes History on KPMP-G (p-value).

Table \ref{hypertension-public} Significance of clinicopathological correlation analysis for Hypertension History on KPMP-G (p-value).

Table \ref{egfr-public} Significance of clinicopathological correlation analysis for eGFR on KPMP-G (p-value).

Table \ref{xj_if} Comparison of methods on XJ-IF for cross-modality diagnosis based on full supervision.

Table \ref{region_lr} Comparison of methods on deposition region classification on XJ-IF using LR-based few-shot learning. The bold value indicates the best model and the hyphen mean the metrics value is lower than 0.5 and not applicable.

Table \ref{pattern_lr} Comparison of methods on deposition pattern classification on XJ-IF using LR-based few-shot learning. The bold value indicates the best model and the hyphen mean the metrics value is lower than 0.5 and not applicable.

Table \ref{region_mlp} Comparison of methods on deposition region classification on XJ-IF using MLP-based few-shot learning. The bold value indicates the best model and the hyphen mean the metrics value is lower than 0.5 and not applicable.

Table \ref{pattern_mlp} Comparison of methods on deposition pattern classification on XJ-IF using MLP-based few-shot learning. The bold value indicates the best model and the hyphen mean the metrics value is lower than 0.5 and not applicable.

Table \ref{region_rf} Comparison of methods on deposition region classification on XJ-IF using RF-based few-shot learning. The bold value indicates the best model and the hyphen mean the metrics value is lower than 0.5 and not applicable.

Table \ref{pattern_rf} Comparison of methods on deposition pattern classification on XJ-IF using RF-based few-shot learning. The bold value indicates the best model and the hyphen mean the metrics value is lower than 0.5 and not applicable.

Table \ref{region_tpl} Comparison of methods on deposition region classification on XJ-IF using PTL-based few-shot learning. The bold value indicates the best model and the hyphen mean the metrics value is lower than 0.5 and not applicable.

Table \ref{pattern_tpl} Comparison of methods on deposition pattern classification on XJ-IF using PTL-based few-shot learning. The bold value indicates the best model and the hyphen mean the metrics value is lower than 0.5 and not applicable.

Table \ref{openset} Performance of GloPath in large-scale real-world study.

Table \ref{27-lesion-anno} Details of the glomerular lesion annotation on XJ-Light-1.

Table \ref{markers} Details of the IF markers in XJ-IF.

Table \ref{xjif-anno} Details of the annotated images on XJ-IF.

Table \ref{clinic_private} Details of the clinical variables on XJ-Light-1.

Table \ref{clinic_public} Details of the clinical variables on KPMP-G.

Table \ref{morphology} Details of the morphological variables.

\FloatBarrier

\newpage
\section*{1. Supplementary Figures}

\newpage
\clearpage
\section*{2. Supplementary Tables}
\FloatBarrier

\begin{table}[htbp]
\caption{Details of the 52 lesion assessment tasks.}
\vspace{-1.2ex}
\label{54tasks}
\centering
\scriptsize
\begin{tabular}{p{3.5cm}p{10cm}}
\hline
\multicolumn{2}{c}{Lesion Recognition}                                                                                               \\ \hline
PAS-MH recognition      & Recognition of thylakoid hyperplasia in PAS staining                                                       \\
PAS-PAS(+) recognition  & Recognition PAS positivity in PAS staining                                                                 \\
PAS-PAS(-) recognition  & Recognition PAS negativity in PAS staining                                                                 \\
PAS-Crum recognition    & Recognition of the presence of ischemic crumpling in PAS staining                                          \\
PAS-ADH recognition     & Recognition the presence of adhesions in PAS staining                                                      \\
PAS-SS recognition      & Recognition the presence of segmental sclerosis in PAS staining                                            \\
PAS-MN recognition      & Recognition the presence of glomerular lesions in membranous nephropathy in PAS staining                   \\
\hline
\end{tabular}
\end{table}

\begin{table}[]
\caption{Number of glomeruli with each type of lesion in XJ-CLI.}\label{xjcli_dataset}
\centering
\begin{tabular}{ccccccccccccl}
\hline
Lesion & MH   & PAS(+) & PAS(-) & Crum & ADH & SS & MN  & MP & EP & Cre & GS  & None of above lesions    \\ \hline
n      & 1987 & 1441   & 24     & 119  & 511 & 90 & 611 & 5  & 1  & 132 & 954 & \multicolumn{1}{c}{9330} \\ \hline
\end{tabular}
\end{table}

\begin{table}[]
\centering
\caption{Comparison of methods on lesion recognition in PAS staining (F1 score). The bold value indicates the best model.}\label{pas_27task}
\begin{tabularx}{1.02\textwidth}{cXXXXXXXXXXX}
\hline
Method      & MH          & PAS(+)      & PAS(-)      & Crum        & ADH         & SS          & MN          & MP          & EP          & Cre         & GS          \\ \hline
RandomInit  & 0.720          & 0.629          & 0.374          & 0.803          & 0.275          & 0.324          & 0.834          & 0.664          & 0.321          & 0.720          & 0.882          \\ 
ImagenetPre & 0.653          & 0.675          & 0.643          & 0.699          & 0.343          & 0.323          & 0.722          & 0.727          & 0.154          & 0.588          & 0.945          \\ 
CONCH       & 0.782          & 0.870          & 0.990          & 0.683          & 0.686          & 0.942          & 0.896          & 0.923          & 0.950          & 0.932          & 0.955          \\ 
PLIP        & 0.789          & 0.862          & 0.954          & 0.649          & 0.619          & 0.794          & 0.898          & 0.827          & 0.839          & 0.935          & 0.966          \\ 
UNI         & 0.832          & 0.877          & \textbf{0.999} & 0.924          & 0.591          & 0.919          & 0.948          & 0.919          & 0.927          & 0.962          & 0.978          \\ 
RenalPath   & 0.820          & 0.873          & 0.994          & 0.844          & 0.791          & 0.852          & 0.891          & 0.919          & 0.920          & 0.928          & 0.979          \\ 
GloPath     & \textbf{0.853} & \textbf{0.901} & 0.998          & \textbf{0.961} & \textbf{0.841} & \textbf{0.956} & \textbf{0.949} & \textbf{0.942} & \textbf{0.957} & \textbf{0.981} & \textbf{0.979} \\ \hline
\end{tabularx}
\end{table}

\begin{table}[]
\caption{Comparison of methods on lesion recognition in MT staining (F1 score). The bold value indicates the best model.}\label{mt_27task}
\begin{tabularx}{\textwidth}{cXXXXXXXXXXX}
\hline
Method      & MH           & ADH          & SS           & SFN          & Cre          & GS           \\ \hline
RandomInit  & 0.735         & 0.441         & 0.515         & 0.768         & 0.636         & 0.790         \\
ImagenetPre & 0.663         & 0.623         & 0.393         & 0.725         & 0.508         & 0.791         \\
CONCH       & 0.897         & 0.948         & 0.951         & \textbf{0.978} & 0.895         & 0.974         \\
PLIP        & 0.891         & \textbf{0.966} & 0.907         & 0.853         & 0.885         & 0.955         \\
UNI         & 0.883         & 0.895         & \textbf{0.973} & 0.974         & 0.955         & 0.975         \\
RenalPath   & 0.917         & 0.928         & 0.895         & 0.974         & 0.911         & 0.977         \\
GloPath     & \textbf{0.935} & 0.940         & 0.972         & 0.977         & \textbf{0.980} & \textbf{0.984} \\ \hline
\end{tabularx}
\end{table}

\begin{table}[]
\caption{Comparison of methods on lesion recognition in PASM staining (F1 score). The bold value indicates the best model.}\label{pasm_27task}
\begin{tabularx}{\textwidth}{cXXXXXXXXXXX}
\hline
Method      & MH         & PAS(-)     & ADH        & SS         & SFN        & MN         & MP         & EP         & Cre        & GS         \\ \hline
RandomInit  & 0.530       & 0.548       & 0.624       & 0.695       & 0.738       & 0.791       & 0.805       & 0.890       & 0.798       & 0.890       \\
ImagenetPre & 0.610       & 0.381       & 0.530       & 0.863       & 0.753       & 0.408       & 0.698       & 0.848       & 0.907       & 0.838       \\
CONCH       & 0.864       & 0.890       & \textbf{0.954} & 0.947       & 0.949       & 0.871       & 0.922       & 0.966       & 0.946       & 0.950       \\
PLIP        & 0.815       & 0.981       & 0.890       & 0.896       & 0.814       & 0.893       & 0.923       & 0.946       & 0.963       & 0.950       \\
UNI         & 0.877       & 0.986       & 0.894       & 0.964       & 0.966       & 0.855       & 0.920       & \textbf{0.969} & 0.958       & 0.959       \\
RenalPath   & 0.887       & 0.982       & 0.774       & 0.919       & 0.960       & 0.862       & 0.901       & 0.957       & 0.946       & 0.966       \\
GloPath     & \textbf{0.895} & \textbf{0.990} & 0.927       & \textbf{0.965} & \textbf{0.970} & \textbf{0.927} & \textbf{0.928} & 0.968       & \textbf{0.970} & \textbf{0.971} \\ \hline
\end{tabularx}
\end{table}

\begin{table}[]
\caption{Comparison of methods on AIDPATH-G under full supervision. The bold value indicates the best model.}\label{aid_linear}
\centering
\begin{tabularx}{0.8\textwidth}{cccccccc}
\hline
Method  & RandomInit & ImageNetPre & CONCH  & PLIP   & UNI    & RenalPath & GloPath         \\ \hline
ROC-AUC & 0.878     & 0.893      & 0.992 & 0.990 & 0.999 & 0.995    & \textbf{0.999} \\ \hline
\end{tabularx}
\end{table}

\begin{table}[]
\caption{Comparison of methods on AIDPATH-G using SVM-based few-shot learning. The bold value indicates the best model.}\label{aid_svm}
\centering
\begin{tabularx}{0.55\textwidth}{cccccc}
\hline
Method      & k=1             & k=5             & k=10            & k=25            & k=100           \\ \hline
RandomInit  & 0.578           & 0.667           & 0.691           & 0.724           & 0.777           \\
ImageNetPre & 0.573           & 0.665           & 0.690           & 0.727           & 0.783           \\
CONCH       & 0.606           & 0.771           & 0.828           & 0.876           & 0.960           \\
PLIP        & 0.674           & 0.836           & 0.886           & 0.923           & 0.966           \\
UNI         & 0.719           & 0.877           & 0.912           & 0.959           & 0.987           \\
RenalPath   & 0.619           & 0.803           & 0.841           & 0.899           & 0.968           \\
GloPath     & \textbf{0.795}  & \textbf{0.904}  & \textbf{0.924}  & \textbf{0.974}  & \textbf{0.992}  \\ \hline
\end{tabularx}
\end{table}

\begin{table}[]
\caption{Comparison of methods on AIDPATH-G using LR-based few-shot learning. The bold value indicates the best model.}\label{aid_lr}
\centering
\begin{tabularx}{0.55\textwidth}{cccccc}
\hline
Method      & k=1             & k=5             & k=10            & k=25            & k=100           \\ \hline
RandomInit  & 0.578           & 0.677           & 0.781           & 0.835           & 0.886           \\
ImageNetPre & 0.573           & 0.682           & 0.783           & 0.839           & 0.890           \\
CONCH       & 0.607           & 0.801           & 0.857           & 0.923           & 0.978           \\
PLIP        & 0.674           & 0.853           & 0.910           & 0.946           & 0.979           \\
UNI         & 0.716           & 0.896           & 0.934           & 0.969           & 0.989           \\
RenalPath   & 0.619           & 0.823           & 0.884           & 0.933           & 0.976           \\
GloPath     & \textbf{0.796}  & \textbf{0.921}  & \textbf{0.944}  & \textbf{0.976}  & \textbf{0.992}  \\ \hline
\end{tabularx}
\end{table}

\begin{table}[]
\caption{Comparison of methods on AIDPATH-G using MLP-based few-shot learning. The bold value indicates the best model.}\label{aid_mlp}
\centering
\begin{tabularx}{0.55\textwidth}{cccccc}
\hline
Method      & k=1             & k=5             & k=10            & k=25            & k=100           \\ \hline
RandomInit  & 0.577           & 0.662           & 0.762           & 0.821           & 0.887           \\
ImageNetPre & 0.573           & 0.675           & 0.768           & 0.831           & 0.893           \\
CONCH       & 0.575           & 0.785           & 0.845           & 0.917           & 0.973           \\
PLIP        & 0.662           & 0.861           & 0.908           & 0.941           & 0.976           \\
UNI         & 0.717           & 0.873           & 0.915           & 0.964           & 0.984           \\
RenalPath   & 0.641           & 0.813           & 0.891           & 0.933           & 0.975           \\
GloPath     & \textbf{0.771}  & \textbf{0.904}  & \textbf{0.925}  & \textbf{0.974}  & \textbf{0.989}  \\ \hline
\end{tabularx}
\end{table}

\begin{table}[]
\caption{Comparison of methods on AIDPATH-G using RF-based few-shot learning. The bold value indicates the best model.}\label{aid_rf}
\centering
\begin{tabularx}{0.55\textwidth}{cccccc}
\hline
Method      & k=1             & k=5             & k=10            & k=25            & k=100           \\ \hline
RandomInit  & 0.567           & 0.681           & 0.742           & 0.798           & 0.845           \\
ImageNetPre & 0.544           & 0.681           & 0.751           & 0.801           & 0.849           \\
CONCH       & 0.515           & 0.762           & 0.828           & 0.875           & 0.931           \\
PLIP        & 0.474           & 0.837           & 0.892           & 0.922           & 0.953           \\
UNI         & 0.542           & 0.872           & 0.909           & 0.947           & 0.979           \\
RenalPath   & 0.514           & 0.785           & 0.857           & 0.904           & 0.950           \\
GloPath     & \textbf{0.576}  & \textbf{0.889}  & \textbf{0.931}  & \textbf{0.954}  & \textbf{0.984}  \\ \hline
\end{tabularx}
\end{table}

\begin{table}[]
\caption{Comparison of methods on AIDPATH-G using PTL-based few-shot learning. The bold value indicates the best model.}\label{aid_tpl}
\centering
\begin{tabularx}{0.55\textwidth}{cccccc}
\hline
Method      & k=1             & k=5             & k=10            & k=25            & k=100           \\ \hline
RandomInit  & 0.578           & 0.661           & 0.696           & 0.712           & 0.710           \\
ImageNetPre & 0.573           & 0.661           & 0.692           & 0.708           & 0.707           \\
CONCH       & 0.606           & 0.783           & 0.827           & 0.838           & 0.856           \\
PLIP        & 0.674           & 0.854           & 0.886           & 0.903           & 0.910           \\
UNI         & 0.718           & 0.884           & 0.901           & 0.919           & 0.933           \\
RenalPath   & 0.619           & 0.796           & 0.823           & 0.840           & 0.835           \\
GloPath     & \textbf{0.795}  & \textbf{0.898}  & \textbf{0.898}  & \textbf{0.912}  & \textbf{0.914}  \\ \hline
\end{tabularx}
\end{table}

\begin{table}[]
\caption{Comparison of methods on lesion grading (ROC-AUC). The bold value indicates the best model.}\label{grading_10task}
\begin{tabularx}{\textwidth}{cXXXXXXXXXXX}
\hline
Method      & PAS-MH-Grading  & MT-MH-Grading   & PASM-MH-Grading & PAS-Cre-Grading & MT-Cre-Grading  & PASM-Cre-Grading & PAS-MP-Grading  & PASM-MP-Grading & PAS-EP-Grading  & PASM-EP-Grading \\ \hline
RandomInit  & 0.584           & 0.704           & 0.716           & 0.824           & 0.827           & 0.796            & 0.854           & 0.815           & 0.766           & 0.849           \\
ImageNetPre & 0.584           & 0.548           & 0.736           & 0.857           & 0.856           & 0.809            & 0.854           & 0.864           & 0.851           & 0.900           \\
CONCH       & 0.700           & 0.866           & 0.849           & 0.927           & 0.954           & 0.931            & 0.910           & 0.849           & 0.932           & \textbf{0.969}  \\
PLIP        & 0.673           & 0.864           & 0.804           & 0.911           & 0.822           & 0.926            & 0.894           & 0.870           & 0.909           & 0.902           \\
UNI         & \textbf{0.736}  & 0.884           & 0.833           & 0.943           & 0.945           & 0.913            & \textbf{0.920}  & 0.915           & \textbf{0.954}  & 0.961           \\
RenalPath   & 0.729           & 0.886           & 0.846           & 0.933           & 0.943           & 0.916            & 0.896           & 0.906           & 0.933           & 0.924           \\
GloPath     & 0.726           & \textbf{0.897}  & \textbf{0.863}  & \textbf{0.960}  & \textbf{0.959}  & \textbf{0.956}   & 0.902           & \textbf{0.918}  & 0.950           & 0.962           \\ \hline
\end{tabularx}
\end{table}

\begin{table}[]
\caption{Comparison of methods on semantic segmentation of the glomerular bow and tuft. The bold value indicates the best model.}\label{seg}
\centering
\begin{tabularx}{0.3\textwidth}{ccc}
\hline
Method      & Bow                  & Tuft                 \\ \hline
RandomInit  & 0.754           & 0.736           \\
ImageNetPre & 0.754           & 0.738           \\
CONCH       & 0.754           & 0.768           \\
PLIP        & 0.754           & 0.826           \\
UNI         & 0.835           & 0.887           \\
RenalPath   & 0.824           & 0.876           \\
GloPath     & \textbf{0.836}  & \textbf{0.888}  \\ \hline
\end{tabularx}
\end{table}

\begin{table}[]
\caption{Comparison of significance level of pathomics-data mining of XJ-Light-1 for GloPath, UNI, and RenalPath.}\label{sig-level-private}
\centering
\begin{tabularx}{0.3\textwidth}{cccc}
\hline
Significance Level & *** & ** & * \\ \hline
GloPath            & 18  & 12 & 7 \\
UNI                & 21  & 8  & 8 \\
RenalPath          & 14  & 9  & 6 \\ \hline
\end{tabularx}
\end{table}

\newpage

\begin{table}[]
\caption{Comparison of significance level of pathomics-data mining of KPMP-G for GloPath, UNI, and RenalPath.}\label{sig-level-public}
\centering
\begin{tabularx}{0.3\textwidth}{cccc}
\hline
Significance Level & *** & ** & *  \\ \hline
GloPath            & 24  & 12 & 7  \\
UNI                & 16  & 10 & 10 \\
RenalPath          & 16  & 13 & 6  \\ \hline
\end{tabularx}
\end{table}

\begin{table}[]
\caption{Significance and effect size of clinicopathological correlation analysis for Gender on XJ-Light-1 (p-value (D))}\label{gender-private}
\centering
\begin{tabularx}{0.65\textwidth}{cccc}
\hline
\multicolumn{4}{c}{Clinic Parameter: Gender}                                                                \\ \hline
Pathology Phenotype & \multicolumn{1}{c}{GloPath} & \multicolumn{1}{c}{UNI} & \multicolumn{1}{c}{RenalPath} \\ \hline
AreaBow\_mean       & 0.0066 (0.1738)  & 0.0125 (0.1636)             & 0.0125 (0.1636)                  \\
AreaBow\_med        & 0.0184 (0.1572)  & 0.0365 (0.1463)             & 0.0342 (0.1452)                  \\
AreaTuft\_mean      & 0.0027 (0.1870)  & 0.0006 (0.1847)             & 0.0032 (0.2065)                   \\
AreaTuft\_med       & 0.0456 (0.1410)  & 0.0371 (0.1661)             & 0.0107 (0.1449)                   \\
Ratio\_mean         & 0.1220 (0.1210)  & 0.1731 (0.0812)             & 0.5433 (0.1131)                   \\
Ratio\_med          & 0.3850 (0.0921)  & 0.2302 (0.1177)             & 0.1416 (0.1061)                   \\
CirBow\_mean        & 0.4913 (0.0846)  & 0.4109 (0.0706)             & 0.7139 (0.0902)                   \\
CirBow\_med         & 0.4620 (0.0866)  & 0.0884 (0.0570)             & 0.9043 (0.1279)                   \\
CirTuft\_mean       & 0.5324 (0.0819)  & 0.0945 (0.1300)             & 0.0800 (0.1265)                    \\
CirTuft\_med        & 0.5933 (0.0781)  & 0.0173 (0.1339)             & 0.0659 (0.1583)                   \\
EccBow\_mean        & 0.9879 (0.0446)  & 0.9481 (0.0424)             & 0.9931 (0.0522)                   \\
EccBow\_med         & 0.8543 (0.0611)  & 0.7352 (0.0545)             & 0.9294 (0.0692)                  \\
EccTuft\_mean       & 0.7980 (0.0651)  & 0.8452 (0.0599)             & 0.8692 (0.0618)                   \\
EccTuft\_med        & 0.9351 (0.0538)  & 0.3753 (0.0525)             & 0.9458 (0.0929)                   \\ \hline
\end{tabularx}
\end{table}

\begin{table}[]
\caption{Significance and effect size of clinicopathological correlation analysis for Age on XJ-Light-1 (p-value (D))}\label{age-private}
\centering
\begin{tabularx}{0.65\textwidth}{cccc}
\hline
\multicolumn{4}{c}{Clinic Parameter: Age}                                                                \\ \hline
Pathology Phenotype & \multicolumn{1}{c}{GloPath} & \multicolumn{1}{c}{UNI} & \multicolumn{1}{c}{RenalPath} \\ \hline
AreaBow\_mean       & 0.6849 (0.0791) & 0.5131 (0.0661) & 0.7777 (0.0827)                        \\
AreaBow\_med        & 0.9145 (0.0556) & 0.9608 (0.0669) & 0.7636 (0.0500)                        \\
AreaTuft\_mean      & 0.7726 (0.0664) & 0.9135 (0.0669) & 0.7631 (0.0556)                        \\
AreaTuft\_med       & 0.2771 (0.1003) & 0.5695 (0.0889) & 0.4179 (0.0787)                        \\
Ratio\_mean         & 0.0127 (0.1621) & 0.1341 (0.1625) & 0.0123 (0.1180)                        \\
Ratio\_med          & 0.0996 (0.1245) & 0.1303 (0.1122) & 0.1725 (0.1187)                        \\
CirBow\_mean        & 0.0393 (0.1427) & 0.0375 (0.1112) & 0.1795 (0.1437)                        \\
CirBow\_med         & 0.0382 (0.1433) & 0.0193 (0.0786) & 0.5718 (0.1555)                        \\
CirTuft\_mean       & 0.0086 (0.1682) & 0.0691 (0.1450) & 0.0347 (0.1319)                        \\
CirTuft\_med        & 0.0030 (0.1840) & 0.0244 (0.1976) & 0.0012 (0.1512)                        \\
EccBow\_mean        & 0.5043 (0.0831) & 0.4434 (0.0776) & 0.5909 (0.0874)                        \\
EccBow\_med         & 0.0994 (0.1245) & 0.4390 (0.0840) & 0.4882 (0.0877)                        \\
EccTuft\_mean       & 0.0304 (0.1473) & 0.0639 (0.1584) & 0.0160 (0.1335)                        \\
EccTuft\_med        & 0.0420 (0.1414) & 0.0122 (0.1783) & 0.0044 (0.1627)                        \\ \hline
\end{tabularx}
\end{table}

\begin{table}[]
\caption{Significance and effect size of clinicopathological correlation analysis for Creatinine on XJ-Light-1 (p-value (D))}\label{creatinine-private}
\centering
\begin{tabularx}{0.65\textwidth}{cccc}
\hline
\multicolumn{4}{c}{Clinic Parameter: Creatinine}                                                                \\ \hline
Pathology Phenotype & \multicolumn{1}{c}{GloPath} & \multicolumn{1}{c}{UNI} & \multicolumn{1}{c}{RenalPath} \\ \hline
AreaBow\_mean       & 0.1154 (0.1472)  & 0.3181 (0.1462)  & 0.1200 (0.1175)                         \\
AreaBow\_med        & 0.0513 (0.1672)  & 0.0661 (0.1984)  & 0.0117 (0.1612)                         \\
AreaTuft\_mean      & 0.1389 (0.1422)  & 0.1460 (0.1586)  & 0.0734 (0.1409)                         \\
AreaTuft\_med       & 0.0548 (0.1656)  & 0.0524 (0.1969)  & 0.0127 (0.1667)                         \\
Ratio\_mean         & 0.0419 (0.1718)  & 0.2449 (0.1469)  & 0.1169 (0.1259)                         \\
Ratio\_med          & 0.1460 (0.1409)  & 0.1060 (0.1691)  & 0.0472 (0.1494)                         \\
CirBow\_mean        & 0.9753 (0.0574)  & 0.0782 (0.0786)  & 0.7952 (0.1571)                         \\
CirBow\_med         & 0.7473 (0.0823)  & 0.0218 (0.1296)  & 0.2164 (0.1859)                         \\
CirTuft\_mean       & 0.0602 (0.1634)  & 0.2043 (0.1600)  & 0.0694 (0.1313)                         \\
CirTuft\_med        & 0.1132 (0.1477)  & 0.0419 (0.1218)  & 0.2791 (0.1718)                         \\
EccBow\_mean        & 0.9967 (0.0480)  & 0.9908 (0.0797)  & 0.7801 (0.0521)                         \\
EccBow\_med         & 0.5091 (0.1004)  & 0.9252 (0.0994)  & 0.5221 (0.0659)                         \\
EccTuft\_mean       & 0.4917 (0.1018)  & 0.6458 (0.0946)  & 0.5843 (0.0900)                         \\
EccTuft\_med        & 0.8975 (0.0692)  & 0.4898 (0.0902)  & 0.6433 (0.1019)                         \\ \hline
\end{tabularx}
\end{table}

\begin{table}[]
\caption{Significance and effect size of clinicopathological correlation analysis for IgA (Binary) on XJ-Light-1 (p-value (D))}\label{IgA2-private}
\centering
\begin{tabularx}{0.65\textwidth}{cccc}
\hline
\multicolumn{4}{c}{Clinic Parameter: IgA (Binary)}                                                                \\ \hline
Pathology Phenotype & \multicolumn{1}{c}{GloPath} & \multicolumn{1}{c}{UNI} & \multicolumn{1}{c}{RenalPath} \\ \hline
AreaBow\_mean       & 0.0007 (0.2240) & 0.0002 (0.2156) & 0.0012 (0.2438)                        \\
AreaBow\_med        & 0.0091 (0.1838) & 0.0098 (0.2059) & 0.0023 (0.1824)                        \\
AreaTuft\_mean      & 0.0041 (0.1969) & 0.0027 (0.2062) & 0.0023 (0.2033)                        \\
AreaTuft\_med       & 0.0829 (0.1407) & 0.0333 (0.1484) & 0.0583 (0.1598)                        \\
Ratio\_mean         & 0.5024 (0.0913) & 0.3824 (0.0978) & 0.4158 (0.1006)                        \\
Ratio\_med          & 0.6412 (0.0817) & 0.3629 (0.0777) & 0.7009 (0.1022)                        \\
CirBow\_mean        & 0.2376 (0.1146) & 0.8695 (0.1125) & 0.2565 (0.0652)                        \\
CirBow\_med         & 0.3270 (0.1054) & 0.7324 (0.0898) & 0.5235 (0.0756)                        \\
CirTuft\_mean       & 0.0158 (0.1739) & 0.2263 (0.1511) & 0.0513 (0.1160)                        \\
CirTuft\_med        & 0.0243 (0.1659) & 0.2270 (0.1677) & 0.0221 (0.1159)                        \\
EccBow\_mean        & 0.4487 (0.0953) & 0.5257 (0.0913) & 0.5031 (0.0897)                        \\
EccBow\_med         & 0.5266 (0.0896) & 0.8384 (0.1165) & 0.2218 (0.0678)                        \\
EccTuft\_mean       & 0.2027 (0.1189) & 0.7522 (0.0949) & 0.4534 (0.1742)                        \\
EccTuft\_med        & 0.4899 (0.0922) & 0.6155 (0.0617) & 0.9073 (0.0834)                        \\ \hline
\end{tabularx}
\end{table}

\begin{table}[]
\caption{Significance and effect size of clinicopathological correlation analysis for IgA Lee Score on XJ-Light-1 (p-value ($\epsilon^2$))}\label{IgALee-private}
\centering
\begin{tabularx}{0.65\textwidth}{cccc}
\hline
\multicolumn{4}{c}{Clinic Parameter: IgA Lee Score}                                                                \\ \hline
Pathology Phenotype & \multicolumn{1}{c}{GloPath} & \multicolumn{1}{c}{UNI} & \multicolumn{1}{c}{RenalPath} \\ \hline
AreaBow\_mean       & 0.3151 (0.0008) & 0.3200 (0.0033) & 0.1986 (0.0008)                        \\
AreaBow\_med        & 0.0028 (0.0266) & 0.0034 (0.0320) & 0.0010 (0.0254)                        \\
AreaTuft\_mean      & 0.1219 (0.0060) & 0.1274 (0.0083) & 0.0805 (0.0058)                        \\
AreaTuft\_med       & 0.0001 (0.0463) & 0.0001 (0.0546) & 0.0000 (0.0446)                        \\
Ratio\_mean         & 0.0122 (0.0185) & 0.0131 (0.0133) & 0.0316 (0.0181)                        \\
Ratio\_med          & 0.0009 (0.0327) & 0.0010 (0.0278) & 0.0022 (0.0321)                        \\
CirBow\_mean        & 0.1823 (0.0038) & 0.0022 (0.0000) & 0.9849 (0.0279)                        \\
CirBow\_med         & 0.0897 (0.0077) & 0.0004 (0.0000) & 0.8824 (0.0372)                        \\
CirTuft\_mean       & 0.0054 (0.0230) & 0.0033 (0.0052) & 0.1421 (0.0257)                        \\
CirTuft\_med        & 0.0014 (0.0303) & 0.0040 (0.0058) & 0.1271 (0.0243)                        \\
EccBow\_mean        & 0.6610 (0.0000) & 0.3627 (0.0000) & 0.4726 (0.0000)                        \\
EccBow\_med         & 0.6400 (0.0000) & 0.3028 (0.0007) & 0.3245 (0.0011)                        \\
EccTuft\_mean       & 0.0413 (0.0119) & 0.0727 (0.0064) & 0.1140 (0.0088)                        \\
EccTuft\_med        & 0.1139 (0.0064) & 0.0617 (0.0074) & 0.0950 (0.0097)                        \\ \hline
\end{tabularx}
\end{table}

\begin{table}[]
\caption{Significance and effect size of clinicopathological correlation analysis for Disease on XJ-Light-1 (p-value ($\epsilon^2$))}\label{disease-private}
\centering
\begin{tabularx}{0.65\textwidth}{cccc}
\hline
\multicolumn{4}{c}{Clinic Parameter: Disease}                                                                \\ \hline
Pathology Phenotype & \multicolumn{1}{c}{GloPath} & \multicolumn{1}{c}{UNI} & \multicolumn{1}{c}{RenalPath} \\ \hline
AreaBow\_mean       & 0.0000 (0.1013) & 0.0000 (0.1051) & 0.0000 (0.1038)                        \\
AreaBow\_med        & 0.0000 (0.0652) & 0.0000 (0.0688) & 0.0000 (0.0172)                        \\
AreaTuft\_mean      & 0.0000 (0.0972) & 0.0000 (0.1099) & 0.0000 (0.0980)                        \\
AreaTuft\_med       & 0.0003 (0.0476) & 0.0002 (0.0552) & 0.0001 (0.0501)                        \\
Ratio\_mean         & 0.0080 (0.0267) & 0.0258 (0.0317) & 0.0036 (0.0193)                        \\
Ratio\_med          & 0.0010 (0.0395) & 0.0011 (0.0528) & 0.0001 (0.0388)                        \\
CirBow\_mean        & 0.1014 (0.0102) & 0.0208 (0.0000) & 0.6053 (0.0207)                        \\
CirBow\_med         & 0.0765 (0.0122) & 0.0170 (0.0000) & 0.9493 (0.0220)                        \\
CirTuft\_mean       & 0.0144 (0.0230) & 0.2787 (0.0054) & 0.2017 (0.0030)                        \\
CirTuft\_med        & 0.0240 (0.0198) & 0.0112 (0.0098) & 0.1080 (0.0246)                        \\
EccBow\_mean        & 0.0697 (0.0128) & 0.1483 (0.0107) & 0.0947 (0.0076)                        \\
EccBow\_med         & 0.0420 (0.0161) & 0.2737 (0.0051) & 0.2096 (0.0031)                        \\
EccTuft\_mean       & 0.0413 (0.0162) & 0.2053 (0.0086) & 0.1275 (0.0052)                        \\
EccTuft\_med        & 0.0687 (0.0129) & 0.0340 (0.0089) & 0.1229 (0.0175)                        \\ \hline
\end{tabularx}
\end{table}

\begin{table}[]
\caption{Significance and effect size of clinicopathological correlation analysis for Lesion on XJ-Light-1 (p-value ($\epsilon^2$))}\label{lesion-private}
\centering
\begin{tabularx}{0.65\textwidth}{cccc}
\hline
\multicolumn{4}{c}{Clinic Parameter: Lesion}                                                                \\ \hline
Pathology Phenotype & \multicolumn{1}{c}{GloPath} & \multicolumn{1}{c}{UNI} & \multicolumn{1}{c}{RenalPath} \\ \hline
AreaBow\_mean       & 0.0963 (0.0073) & 0.0819 (0.0062) & 0.1184 (0.0082)                        \\
AreaBow\_med        & 0.8887 (0.0000) & 0.8090 (0.0000) & 0.9395 (0.0000)                        \\
AreaTuft\_mean      & 0.4036 (0.0000) & 0.3238 (0.0002) & 0.3569 (0.0007)                        \\
AreaTuft\_med       & 0.8736 (0.0000) & 0.9567 (0.0000) & 0.8917 (0.0000)                        \\
Ratio\_mean         & 0.0273 (0.0141) & 0.0359 (0.0152) & 0.0226 (0.0126)                        \\
Ratio\_med          & 0.0674 (0.0092) & 0.1806 (0.0114) & 0.0453 (0.0039)                        \\
CirBow\_mean        & 0.0603 (0.0098) & 0.0419 (0.0000) & 0.7402 (0.0119)                        \\
CirBow\_med         & 0.0795 (0.0083) & 0.0813 (0.0000) & 0.8173 (0.0082)                        \\
CirTuft\_mean       & 0.0078 (0.0210) & 0.0204 (0.0069) & 0.1035 (0.0157)                        \\
CirTuft\_med        & 0.0126 (0.0183) & 0.0158 (0.0129) & 0.0341 (0.0171)                        \\
EccBow\_mean        & 0.8825 (0.0000) & 0.8717 (0.0000) & 0.8085 (0.0000)                        \\
EccBow\_med         & 0.6739 (0.0000) & 0.9555 (0.0000) & 0.9018 (0.0000)                        \\
EccTuft\_mean       & 0.2192 (0.0028) & 0.3932 (0.0016) & 0.2723 (0.0000)                        \\
EccTuft\_med        & 0.4888 (0.0000) & 0.3548 (0.0000) & 0.4740 (0.0002)                        \\ \hline
\end{tabularx}
\end{table}

\begin{table}[]
\caption{Multivariate Regression Analysis of Pathology Phenotypes Associated with Creatinine. Other dependent variables: Gender, Age, Disease Type (IgAN, LN, MN, DN, and MCD). G, A, D, and P indicate gender, age, disease type, and pathology phenotype respectively.}\label{multi_vari_creatine}
\begin{tabular}{llllllllll}
\hline
Pathology Phenotype      & $R^2$  & coef (G) & p (G)  & coef (A) & p (A)  & coef (D) & p (D)  & coef (P)   & p (P)  \\ \hline
Proposed\_AreaBow\_mean  & 0.1071 & 23.1147  & 0.0135 & 1.5866   & 0.0000 & -45.8914 & 0.0001 & -0.0004    & 0.0476 \\
Proposed\_AreaBow\_med   & 0.1188 & 23.4835  & 0.0113 & 1.5234   & 0.0000 & -44.6479 & 0.0001 & -0.0005    & 0.0031 \\
Proposed\_AreaTuft\_mean & 0.1202 & 25.1494  & 0.0070 & 1.5210   & 0.0000 & -42.4377 & 0.0002 & -0.0008    & 0.0023 \\
Proposed\_AreaTuft\_med  & 0.1204 & 23.5983  & 0.0108 & 1.4747   & 0.0000 & -44.5297 & 0.0001 & -0.0006    & 0.0022 \\
Proposed\_Ratio\_mean    & 0.1167 & 22.6231  & 0.0145 & 1.4406   & 0.0000 & -46.7163 & 0.0000 & -201.0923  & 0.0050 \\
Proposed\_Ratio\_med     & 0.1262 & 22.6289  & 0.0139 & 1.4404   & 0.0000 & -44.5344 & 0.0001 & -219.8215  & 0.0006 \\
Proposed\_CirBow\_mean   & 0.0978 & 21.0700  & 0.0242 & 1.6248   & 0.0000 & -50.9139 & 0.0000 & 833.2359   & 0.6992 \\
Proposed\_CirBow\_med    & 0.0980 & 21.0893  & 0.0240 & 1.6308   & 0.0000 & -51.2372 & 0.0000 & 1027.1659  & 0.6443 \\
Proposed\_CirTuft\_mean  & 0.1088 & 19.9190  & 0.0317 & 1.4999   & 0.0000 & -51.0965 & 0.0000 & -2921.3320 & 0.0320 \\
Proposed\_CirTuft\_med   & 0.1140 & 19.4368  & 0.0356 & 1.4759   & 0.0000 & -50.1649 & 0.0000 & -3435.9837 & 0.0095 \\
Proposed\_EccBow\_mean   & 0.0981 & 20.8299  & 0.0255 & 1.6174   & 0.0000 & -50.7702 & 0.0000 & -68.7994   & 0.6224 \\
Proposed\_EccBow\_med    & 0.0980 & 20.9027  & 0.0250 & 1.6191   & 0.0000 & -50.8867 & 0.0000 & -56.3432   & 0.6425 \\
Proposed\_EccTuft\_mean  & 0.0977 & 20.7669  & 0.0259 & 1.5903   & 0.0000 & -50.7936 & 0.0000 & 39.3679    & 0.7597 \\
Proposed\_EccTuft\_med   & 0.0994 & 20.6860  & 0.0264 & 1.5696   & 0.0000 & -50.6826 & 0.0000 & 94.7958    & 0.3782 \\ \hline
\end{tabular}
\end{table}

\begin{table}[]
\caption{Multivariate Regression Analysis of Pathology Phenotypes Associated with Disease Type. Other dependent variables: Gender, Age, Disease Type (IgAN, LN, MN, DN, and MCD).}\label{multi_vari_disease}
\begin{tabular}{llllllllll}
\hline
Pathology Phenotype      & $R^2$  & coef (G) & p (G)  & coef (A) & p (A)  & coef (D) & p (D)  & coef (P)  & p (P)  \\ \hline
Proposed\_AreaBow\_mean  & 0.1193 & 0.1964   & 0.5911 & -0.0126  & 0.3174 & -0.9831  & 0.1347 & 0.0000    & 0.0001 \\
Proposed\_AreaBow\_med   & 0.1202 & 0.1415   & 0.6986 & -0.0149  & 0.2406 & -1.1577  & 0.0750 & 0.0000    & 0.0001 \\
Proposed\_AreaTuft\_mean & 0.1291 & 0.2466   & 0.5036 & -0.0152  & 0.2349 & -0.9330  & 0.1560 & 0.0000    & 0.0000 \\
Proposed\_AreaTuft\_med  & 0.1125 & 0.1601   & 0.6600 & -0.0158  & 0.2156 & -1.2221  & 0.0587 & 0.0000    & 0.0001 \\
Proposed\_Ratio\_mean    & 0.0660 & 0.1283   & 0.7201 & -0.0134  & 0.2865 & -1.3799  & 0.0299 & -5.5755   & 0.0170 \\
Proposed\_Ratio\_med     & 0.0763 & 0.1272   & 0.7241 & -0.0133  & 0.2876 & -1.3252  & 0.0375 & -5.6022   & 0.0042 \\
Proposed\_CirBow\_mean   & 0.0546 & 0.0252   & 0.9438 & -0.0112  & 0.3667 & -1.4670  & 0.0206 & -117.2984 & 0.0994 \\
Proposed\_CirBow\_med    & 0.0506 & 0.0383   & 0.9145 & -0.0106  & 0.3898 & -1.4466  & 0.0226 & -95.1447  & 0.1872 \\
Proposed\_CirTuft\_mean  & 0.0811 & 0.0453   & 0.8999 & -0.0143  & 0.2596 & -1.5632  & 0.0147 & -136.2075 & 0.0019 \\
Proposed\_CirTuft\_med   & 0.0905 & 0.0168   & 0.9632 & -0.0150  & 0.2397 & -1.5344  & 0.0181 & -131.6665 & 0.0006 \\
Proposed\_EccBow\_mean   & 0.0455 & 0.0724   & 0.8382 & -0.0087  & 0.4733 & -1.4987  & 0.0177 & 3.3006    & 0.5222 \\
Proposed\_EccBow\_med    & 0.0514 & 0.0470   & 0.8950 & -0.0098  & 0.4262 & -1.4918  & 0.0184 & 6.0578    & 0.1703 \\
Proposed\_EccTuft\_mean  & 0.0645 & 0.0591   & 0.8684 & -0.0113  & 0.3613 & -1.5036  & 0.0180 & 9.9814    & 0.0224 \\
Proposed\_EccTuft\_med   & 0.0688 & 0.0552   & 0.8775 & -0.0110  & 0.3735 & -1.5086  & 0.0179 & 9.3923    & 0.0126 \\ \hline
\end{tabular}
\end{table}

\begin{table}[]
\caption{Significance and effect size of clinicopathological correlation analysis for Enrollment Category on KPMP-G (p-value ($\epsilon^2$))}\label{enrollment-public}
\centering
\begin{tabularx}{0.65\textwidth}{cccc}
\hline
\multicolumn{4}{c}{Clinic Parameter: Enrollment Category} \\ \hline
Pathology Phenotype   & GloPath   & UNI      & RenalPath  \\ \hline
AreaBow\_mean         & 0.0041 (0.0000)    & 0.0024 (0.0000)   & 0.0059 (0.0000)     \\
AreaBow\_med          & 0.6383 (0.0000)    & 0.5592 (0.0000)   & 0.6403 (0.0000)     \\
AreaTuft\_mean        & 0.1044 (0.0000)    & 0.2850 (0.0000)   & 0.0335 (0.0000)     \\
AreaTuft\_med         & 0.3037 (0.0000)    & 0.7116 (0.0000)   & 0.8076 (0.0000)     \\
Ratio\_mean           & 0.5496 (0.0000)    & 0.0072 (0.0000)   & 0.0457 (0.0000)     \\
Ratio\_med            & 0.3247 (0.0000)    & 0.0057 (0.0000)   & 0.2520 (0.0000)     \\
CirBow\_mean          & 0.0099 (0.0000)    & 0.2775 (0.0000)   & 0.3117 (0.0000)     \\
CirBow\_med           & 0.0120 (0.0000)    & 0.1452 (0.0000)   & 0.7453 (0.0000)     \\
CirTuft\_mean         & 0.0636 (0.0000)    & 0.0024 (0.0000)   & 0.0102 (0.0000)     \\
CirTuft\_med          & 0.0070 (0.0000)    & 0.0020 (0.0000)   & 0.1613 (0.0000)     \\
EccBow\_mean          & 0.6204 (0.0000)    & 0.8672 (0.0000)   & 0.5014 (0.0000)     \\
EccBow\_med           & 0.8631 (0.0000)    & 0.4035 (0.0000)   & 0.1044 (0.0000)     \\
EccTuft\_mean         & 0.6127 (0.0000)    & 0.7135 (0.0000)   & 0.7825 (0.0000)     \\
EccTuft\_med          & 0.3627 (0.0000)    & 0.7948 (0.0000)   & 0.3665 (0.0000)     \\ \hline
\end{tabularx}
\end{table}

\begin{table}[]
\caption{Significance and effect size of clinicopathological correlation analysis for Gender on KPMP-G (p-value (D))}\label{gender-public}
\centering
\begin{tabularx}{0.65\textwidth}{cccc}
\hline
\multicolumn{4}{c}{Clinic Parameter: Gender}       \\ \hline
Pathology Phenotype & GloPath & UNI    & RenalPath \\ \hline
AreaBow\_mean       & 0.5660 (0.0869)  & 0.9777 (0.0884) & 0.5440 (0.0515)    \\
AreaBow\_med        & 0.9727 (0.0526)  & 0.7460 (0.0484) & 0.9880 (0.0746)    \\
AreaTuft\_mean      & 0.9923 (0.0465)  & 0.9671 (0.0563) & 0.9516 (0.0537)    \\
AreaTuft\_med       & 0.7014 (0.0777)  & 0.0800 (0.0909) & 0.5088 (0.0142)    \\
Ratio\_mean         & 0.1904 (0.1207)  & 0.4768 (0.1238) & 0.1689 (0.0932)    \\
Ratio\_med          & 0.3891 (0.1000)  & 0.2627 (0.1453) & 0.0677 (0.0112)    \\
CirBow\_mean        & 0.0328 (0.1603)  & 0.1356 (0.1022) & 0.3633 (0.1293)    \\
CirBow\_med         & 0.0111 (0.1805)  & 0.2411 (0.0935) & 0.4736 (0.1143)    \\
CirTuft\_mean       & 0.3041 (0.1077)  & 0.0367 (0.0973) & 0.4227 (0.1581)    \\
CirTuft\_med        & 0.1535 (0.1262)  & 0.0587 (0.1427) & 0.0764 (0.1484)    \\
EccBow\_mean        & 0.7400 (0.0750)  & 0.5683 (0.0520) & 0.9752 (0.0867)    \\
EccBow\_med         & 0.6525 (0.0810)  & 0.3592 (0.0727) & 0.7735 (0.1026)    \\
EccTuft\_mean       & 0.5883 (0.0853)  & 0.9848 (0.0687) & 0.8276 (0.0495)    \\
EccTuft\_med        & 0.9075 (0.0616)  & 0.7889 (0.0408) & 0.9986 (0.0716)    \\ \hline
\end{tabularx}
\end{table}

\begin{table}[]
\caption{Significance and effect size of clinicopathological correlation analysis for Age on KPMP-G (p-value ($\epsilon^2$))}\label{age-public}
\centering
\begin{tabularx}{0.65\textwidth}{cccc}
\hline
\multicolumn{4}{c}{Clinic Parameter: Age}       \\ \hline
Pathology Phenotype & GloPath & UNI    & RenalPath \\ \hline
AreaBow\_mean       & 0.0591 (0.0185)  & 0.0607 (0.0165) & 0.0747 (0.0182)    \\
AreaBow\_med        & 0.0125 (0.0313)  & 0.0189 (0.0268) & 0.0217 (0.0280)    \\
AreaTuft\_mean      & 0.0256 (0.0255)  & 0.0320 (0.0274) & 0.0204 (0.0236)    \\
AreaTuft\_med       & 0.0351 (0.0229)  & 0.1036 (0.0201) & 0.0488 (0.0136)    \\
Ratio\_mean         & 0.3937 (0.0006)  & 0.0583 (0.0000) & 0.4656 (0.0186)    \\
Ratio\_med          & 0.5305 (0.0000)  & 0.0892 (0.0000) & 0.6226 (0.0149)    \\
CirBow\_mean        & 0.1248 (0.0119)  & 0.5318 (0.0000) & 0.7286 (0.0000)    \\
CirBow\_med         & 0.4274 (0.0000)  & 0.3330 (0.0000) & 0.8157 (0.0024)    \\
CirTuft\_mean       & 0.7107 (0.0000)  & 0.5707 (0.0000) & 0.8937 (0.0000)    \\
CirTuft\_med        & 0.5385 (0.0000)  & 0.5693 (0.0000) & 0.9221 (0.0000)    \\
EccBow\_mean        & 0.0634 (0.0179)  & 0.2042 (0.0161) & 0.0774 (0.0073)    \\
EccBow\_med         & 0.0623 (0.0180)  & 0.3442 (0.0069) & 0.2127 (0.0021)    \\
EccTuft\_mean       & 0.0623 (0.0180)  & 0.0316 (0.0359) & 0.0070 (0.0237)    \\
EccTuft\_med        & 0.1209 (0.0122)  & 0.1140 (0.0211) & 0.0433 (0.0127)    \\ \hline
\end{tabularx}
\end{table}

\begin{table}[]
\caption{Significance and effect size of clinicopathological correlation analysis for Proteinuria on KPMP-G (p-value ($\epsilon^2$))}\label{proteinuria-public}
\centering
\begin{tabularx}{0.65\textwidth}{cccc}
\hline
\multicolumn{4}{c}{Clinic Parameter: Proteinuria}       \\ \hline
Pathology Phenotype & GloPath & UNI    & RenalPath \\ \hline
AreaBow\_mean       & 0.0127 (0.0223)  & 0.0189 (0.0157) & 0.0319 (0.0195)    \\
AreaBow\_med        & 0.2012 (0.0021)  & 0.1729 (0.0003) & 0.2519 (0.0032)    \\
AreaTuft\_mean      & 0.0140 (0.0216)  & 0.0367 (0.0081) & 0.0904 (0.0147)    \\
AreaTuft\_med       & 0.2247 (0.0012)  & 0.2766 (0.0000) & 0.4925 (0.0000)    \\
Ratio\_mean         & 0.8597 (0.0000)  & 0.4209 (0.0000) & 0.4607 (0.0000)    \\
Ratio\_med          & 0.7821 (0.0000)  & 0.7489 (0.0043) & 0.1499 (0.0000)    \\
CirBow\_mean        & 0.4734 (0.0000)  & 0.6128 (0.0011) & 0.2275 (0.0000)    \\
CirBow\_med         & 0.3488 (0.0000)  & 0.5162 (0.0000) & 0.4318 (0.0000)    \\
CirTuft\_mean       & 0.7259 (0.0000)  & 0.3088 (0.0108) & 0.0628 (0.0000)    \\
CirTuft\_med        & 0.7205 (0.0000)  & 0.2006 (0.0167) & 0.0280 (0.0021)    \\
EccBow\_mean        & 0.8221 (0.0000)  & 0.8177 (0.0000) & 0.6549 (0.0000)    \\
EccBow\_med         & 0.7867 (0.0000)  & 0.8618 (0.0000) & 0.7874 (0.0000)    \\
EccTuft\_mean       & 0.3629 (0.0000)  & 0.4663 (0.0000) & 0.5441 (0.0000)    \\
EccTuft\_med        & 0.5128 (0.0000)  & 0.3355 (0.0000) & 0.7193 (0.0000)    \\ \hline
\end{tabularx}
\end{table}

\begin{table}[]
\caption{Significance and effect size of clinicopathological correlation analysis for A1c on KPMP-G (p-value ($\epsilon^2$))}\label{a1c-public}
\centering
\begin{tabularx}{0.65\textwidth}{cccc}
\hline
\multicolumn{4}{c}{Clinic Parameter: A1c}       \\ \hline
Pathology Phenotype & GloPath & UNI    & RenalPath \\ \hline
AreaBow\_mean       & 0.0001 (0.0574)  & 0.0003 (0.0506) & 0.0003 (0.0523)    \\
AreaBow\_med        & 0.0011 (0.0429)  & 0.0008 (0.0405) & 0.0015 (0.0450)    \\
AreaTuft\_mean      & 0.0007 (0.0453)  & 0.0007 (0.0440) & 0.0009 (0.0458)    \\
AreaTuft\_med       & 0.0051 (0.0320)  & 0.0132 (0.0344) & 0.0035 (0.0252)    \\
Ratio\_mean         & 0.5624 (0.0000)  & 0.0639 (0.0000) & 0.6657 (0.0139)    \\
Ratio\_med          & 0.3730 (0.0004)  & 0.0300 (0.0000) & 0.6526 (0.0194)    \\
CirBow\_mean        & 0.9727 (0.0000)  & 0.4376 (0.0000) & 0.7993 (0.0000)    \\
CirBow\_med         & 0.7370 (0.0000)  & 0.7429 (0.0000) & 0.7916 (0.0000)    \\
CirTuft\_mean       & 0.0923 (0.0112)  & 0.1097 (0.0000) & 0.7729 (0.0099)    \\
CirTuft\_med        & 0.1790 (0.0062)  & 0.0804 (0.0000) & 0.7616 (0.0122)    \\
EccBow\_mean        & 0.0406 (0.0172)  & 0.0485 (0.0230) & 0.0181 (0.0159)    \\
EccBow\_med         & 0.0548 (0.0150)  & 0.3759 (0.0274) & 0.0097 (0.0003)    \\
EccTuft\_mean       & 0.0995 (0.0106)  & 0.0289 (0.0078) & 0.1456 (0.0196)    \\
EccTuft\_med        & 0.1737 (0.0064)  & 0.0197 (0.0173) & 0.0398 (0.0224)    \\ \hline
\end{tabularx}
\end{table}

\begin{table}[]
\caption{Significance and effect size of clinicopathological correlation analysis for Albuminuria on KPMP-G (p-value (D))}\label{albuminuria-public}
\centering
\begin{tabularx}{0.65\textwidth}{cccc}
\hline
\multicolumn{4}{c}{Clinic Parameter: Albuminuria}       \\ \hline
Pathology Phenotype & GloPath & UNI    & RenalPath \\ \hline
AreaBow\_mean       & 0.0002 (0.2975)  & 0.0004 (0.2859) & 0.0005 (0.2885)    \\
AreaBow\_med        & 0.0467 (0.1907)  & 0.0263 (0.1842) & 0.0596 (0.2049)    \\
AreaTuft\_mean      & 0.0003 (0.2949)  & 0.0010 (0.2476) & 0.0037 (0.2717)    \\
AreaTuft\_med       & 0.1949 (0.1494)  & 0.1971 (0.1343) & 0.3000 (0.1490)    \\
Ratio\_mean         & 0.8728 (0.0804)  & 0.8504 (0.1389) & 0.2643 (0.0828)    \\
Ratio\_med          & 0.7901 (0.0886)  & 0.3085 (0.1528) & 0.1756 (0.1333)    \\
CirBow\_mean        & 0.0329 (0.1995)  & 0.7690 (0.1977) & 0.0354 (0.0906)    \\
CirBow\_med         & 0.0745 (0.1782)  & 0.4520 (0.1684) & 0.1051 (0.1180)    \\
CirTuft\_mean       & 0.0242 (0.2069)  & 0.0824 (0.2516) & 0.0030 (0.1754)    \\
CirTuft\_med        & 0.1637 (0.1550)  & 0.0696 (0.2704) & 0.0011 (0.1800)    \\
EccBow\_mean        & 0.4869 (0.1149)  & 0.3201 (0.1560) & 0.1586 (0.1319)    \\
EccBow\_med         & 0.1777 (0.1524)  & 0.7709 (0.1474) & 0.2068 (0.0904)    \\
EccTuft\_mean       & 0.7293 (0.0940)  & 0.6466 (0.1105) & 0.5366 (0.1010)    \\
EccTuft\_med        & 0.8199 (0.0858)  & 0.8728 (0.1157) & 0.4784 (0.0804)    \\ \hline
\end{tabularx}
\end{table}

\begin{table}[]
\caption{Significance and effect size of clinicopathological correlation analysis for Diabetes History on KPMP-G (p-value (D))}\label{diabetes-public}
\centering
\begin{tabularx}{0.65\textwidth}{cccc}
\hline
\multicolumn{4}{c}{Clinic Parameter: Diabetes History}       \\ \hline
Pathology Phenotype & GloPath & UNI    & RenalPath \\ \hline
AreaBow\_mean       & 0.0004 (0.2446)  & 0.0006 (0.2297) & 0.0012 (0.2394)    \\
AreaBow\_med        & 0.0085 (0.1964)  & 0.0158 (0.1466) & 0.0939 (0.1851)    \\
AreaTuft\_mean      & 0.0239 (0.1767)  & 0.1065 (0.1485) & 0.0869 (0.1435)    \\
AreaTuft\_med       & 0.1842 (0.1292)  & 0.3436 (0.1231) & 0.2290 (0.1103)    \\
Ratio\_mean         & 0.5747 (0.0914)  & 0.9438 (0.1464) & 0.0946 (0.0609)    \\
Ratio\_med          & 0.4944 (0.0976)  & 0.8057 (0.0794) & 0.7468 (0.0748)    \\
CirBow\_mean        & 0.0481 (0.1620)  & 0.7128 (0.0755) & 0.7959 (0.0815)    \\
CirBow\_med         & 0.0204 (0.1801)  & 0.9922 (0.1145) & 0.3033 (0.0495)    \\
CirTuft\_mean       & 0.4785 (0.0991)  & 0.4438 (0.1106) & 0.3399 (0.1014)    \\
CirTuft\_med        & 0.6729 (0.0847)  & 0.0681 (0.0796) & 0.7440 (0.1542)    \\
EccBow\_mean        & 0.0030 (0.2146)  & 0.0076 (0.2212) & 0.0020 (0.1988)    \\
EccBow\_med         & 0.0121 (0.1902)  & 0.0553 (0.2097) & 0.0041 (0.1590)    \\
EccTuft\_mean       & 0.0083 (0.1967)  & 0.0066 (0.1586) & 0.0563 (0.2008)    \\
EccTuft\_med        & 0.0110 (0.1917)  & 0.1413 (0.1621) & 0.0479 (0.1362)    \\ \hline
\end{tabularx}
\end{table}

\begin{table}[]
\caption{Significance and effect size of clinicopathological correlation analysis for Hypertension History on KPMP-G (p-value (D))}\label{hypertension-public}
\centering
\begin{tabularx}{0.65\textwidth}{cccc}
\hline
\multicolumn{4}{c}{Clinic Parameter: Hypertension History}       \\ \hline
Pathology Phenotype & GloPath & UNI    & RenalPath \\ \hline
AreaBow\_mean       & 0.3550 (0.1368)  & 0.4629 (0.1098) & 0.6268 (0.1252)    \\
AreaBow\_med        & 0.3560 (0.1367)  & 0.4251 (0.1368) & 0.3550 (0.1290)    \\
AreaTuft\_mean      & 0.9289 (0.0768)  & 0.9085 (0.1069) & 0.6563 (0.0817)    \\
AreaTuft\_med       & 0.1961 (0.1596)  & 0.1222 (0.1559) & 0.2163 (0.1752)    \\
Ratio\_mean         & 0.0075 (0.2485)  & 0.0041 (0.3143) & 0.0003 (0.2609)    \\
Ratio\_med          & 0.0268 (0.2179)  & 0.0007 (0.2186) & 0.0256 (0.2954)    \\
CirBow\_mean        & 0.0173 (0.2292)  & 0.2707 (0.1095) & 0.6311 (0.1478)    \\
CirBow\_med         & 0.0023 (0.2724)  & 0.4071 (0.1867) & 0.0848 (0.1306)    \\
CirTuft\_mean       & 0.0143 (0.2334)  & 0.0322 (0.2876) & 0.0011 (0.2137)    \\
CirTuft\_med        & 0.0138 (0.2338)  & 0.0167 (0.2261) & 0.0190 (0.2296)    \\
EccBow\_mean        & 0.6984 (0.1031)  & 0.8321 (0.1140) & 0.5798 (0.0906)    \\
EccBow\_med         & 0.9888 (0.0637)  & 0.6152 (0.1143) & 0.5755 (0.1106)    \\
EccTuft\_mean       & 0.1422 (0.1708)  & 0.1106 (0.1712) & 0.1389 (0.1787)    \\
EccTuft\_med        & 0.7903 (0.0947)  & 0.1666 (0.1877) & 0.0795 (0.1645)    \\ \hline
\end{tabularx}
\end{table}

\begin{table}[]
\caption{Significance and effect size of clinicopathological correlation analysis for eGFR on KPMP-G (p-value ($\epsilon^2$))}\label{egfr-public}
\centering
\begin{tabularx}{0.65\textwidth}{cccc}
\hline
\multicolumn{4}{c}{Clinic Parameter: eGFR}                                                                \\ \hline
Pathology Phenotype & \multicolumn{1}{c}{GloPath} & \multicolumn{1}{c}{UNI} & \multicolumn{1}{c}{RenalPath} \\ \hline
AreaBow\_mean       & 0.9834 (0.0000) & 0.9294 (0.0000) & 0.8818 (0.0000)                        \\
AreaBow\_med        & 0.0972 (0.0086) & 0.1308 (0.0100) & 0.0784 (0.0067)                        \\
AreaTuft\_mean      & 0.2920 (0.0015) & 0.0550 (0.0042) & 0.1933 (0.0123)                        \\
AreaTuft\_med       & 0.0240 (0.0177) & 0.0041 (0.0239) & 0.0093 (0.0291)                        \\
Ratio\_mean         & 0.0006 (0.0412) & 0.0000 (0.0642) & 0.0000 (0.0844)                        \\
Ratio\_med          & 0.0002 (0.0481) & 0.0000 (0.0636) & 0.0000 (0.0784)                        \\
CirBow\_mean        & 0.0025 (0.0324) & 0.2314 (0.0027) & 0.2435 (0.0030)                        \\
CirBow\_med         & 0.0046 (0.0285) & 0.1615 (0.0053) & 0.1629 (0.0053)                        \\
CirTuft\_mean       & 0.0100 (0.0234) & 0.0003 (0.0173) & 0.0258 (0.0463)                        \\
CirTuft\_med        & 0.0181 (0.0195) & 0.0004 (0.0115) & 0.0628 (0.0440)                        \\
EccBow\_mean        & 0.6074 (0.0000) & 0.6184 (0.0005) & 0.3383 (0.0000)                        \\
EccBow\_med         & 0.3967 (0.0000) & 0.2768 (0.0000) & 0.3666 (0.0018)                        \\
EccTuft\_mean       & 0.1259 (0.0070) & 0.0494 (0.0093) & 0.0879 (0.0130)                        \\
EccTuft\_med        & 0.1903 (0.0043) & 0.0588 (0.0050) & 0.1711 (0.0119)                        \\ \hline
\end{tabularx}
\end{table}

\begin{table}[]
\caption{Comparison of methods on XJ-IF for cross-modality diagnosis based on full supervision.}\label{xj_if}
\centering
\begin{tabularx}{0.8\textwidth}{cccccccc}
\hline
Method  & RandomInit & ImageNetPre & CONCH  & PLIP   & UNI    & RenalPath & GloPath \\ \hline
Region  & 0.548      & 0.508       & 0.850  & 0.836  & 0.917  & 0.914     & \textbf{0.942}  \\
Pattern & 0.473      & 0.479       & 0.801  & 0.849  & 0.928  & 0.912     & \textbf{0.944}  \\ \hline
\end{tabularx}
\end{table}

\begin{table}[]
\caption{Comparison of methods on deposition region classification on XJ-IF using LR-based few-shot learning. The bold value indicates the best model and the hyphen mean the metrics value is lower than 0.5 and not applicable.}\label{region_lr}
\centering
\begin{tabularx}{0.5\textwidth}{cccccc}
\hline
Method      & k=1        & k=5        & k=10            & k=25            & k=100           \\ \hline
RandomInit  & -          & -          & -               & 0.501           & -               \\
ImageNetPre & -          & -          & -               & -               & -               \\
CONCH       & -          & -          & -               & -               & 0.823           \\
PLIP        & -          & -          & -               & 0.561           & 0.848           \\
UNI         & -          & -          & 0.704           & 0.859           & 0.929           \\
RenalPath   & -          & -          & 0.657           & 0.685           & 0.862           \\
GloPath     & \textbf{-} & \textbf{-} & \textbf{0.784}  & \textbf{0.911}  & \textbf{0.945}  \\ \hline
\end{tabularx}
\end{table}

\begin{table}[]
\caption{Comparison of methods on deposition pattern classification on XJ-IF using LR-based few-shot learning. The bold value indicates the best model and the hyphen mean the metrics value is lower than 0.5 and not applicable.}\label{pattern_lr}
\centering
\begin{tabularx}{0.5\textwidth}{cccccc}
\hline
Method      & k=1        & k=5             & k=10            & k=25            & k=100           \\ \hline
RandomInit  & -          & \textbf{0.500}  & 0.503           & -               & 0.505           \\
ImageNetPre & -          & -               & 0.503           & 0.500           & 0.505           \\
CONCH       & -          & -               & -               & -               & 0.804           \\
PLIP        & -          & -               & -               & 0.520           & 0.825           \\
UNI         & -          & -               & 0.679           & 0.854           & 0.922           \\
RenalPath   & -          & -               & 0.674           & 0.694           & 0.859           \\
GloPath     & \textbf{-} & \textbf{-}      & \textbf{0.852}  & \textbf{0.904}  & \textbf{0.941}  \\ \hline
\end{tabularx}
\end{table}

\begin{table}[]
\caption{Comparison of methods on deposition region classification on XJ-IF using MLP-based few-shot learning. The bold value indicates the best model and the hyphen mean the metrics value is lower than 0.5 and not applicable.}\label{region_mlp}
\centering
\begin{tabularx}{0.5\textwidth}{cccccc}
\hline
Method      & k=1             & k=5             & k=10            & k=25            & k=100           \\ \hline
RandomInit  & 0.511           & 0.514           & 0.521           & 0.525           & 0.550           \\
ImageNetPre & 0.512           & 0.512           & 0.521           & 0.519           & 0.544           \\
CONCH       & 0.560           & 0.635           & 0.719           & 0.794           & 0.868           \\
PLIP        & 0.610           & 0.733           & 0.797           & 0.840           & 0.880           \\
UNI         & 0.625           & 0.717           & 0.830           & 0.885           & \textbf{0.939}  \\
RenalPath   & 0.660           & 0.739           & 0.792           & 0.852           & 0.901           \\
GloPath     & \textbf{0.666}  & \textbf{0.764}  & \textbf{0.851}  & \textbf{0.903}  & 0.937           \\ \hline
\end{tabularx}
\end{table}

\begin{table}[]
\caption{Comparison of methods on deposition pattern classification on XJ-IF using MLP-based few-shot learning. The bold value indicates the best model and the hyphen mean the metrics value is lower than 0.5 and not applicable.}\label{pattern_mlp}
\centering
\begin{tabularx}{0.5\textwidth}{cccccc}
\hline
Method      & k=1             & k=5             & k=10            & k=25            & k=100           \\ \hline
RandomInit  & 0.500           & 0.507           & 0.510           & 0.516           & 0.538           \\
ImageNetPre & 0.500           & 0.507           & 0.507           & 0.512           & 0.532           \\
CONCH       & 0.536           & 0.663           & 0.695           & 0.753           & 0.852           \\
PLIP        & 0.663           & 0.758           & 0.762           & 0.779           & 0.864           \\
UNI         & 0.618           & 0.787           & 0.822           & 0.868           & 0.924           \\
RenalPath   & 0.662           & 0.761           & 0.790           & 0.817           & 0.893           \\
GloPath     & \textbf{0.665}  & \textbf{0.835}  & \textbf{0.843}  & \textbf{0.887}  & \textbf{0.926}  \\ \hline
\end{tabularx}
\end{table}

\begin{table}[]
\caption{Comparison of methods on deposition region classification on XJ-IF using RF-based few-shot learning. The bold value indicates the best model and the hyphen mean the metrics value is lower than 0.5 and not applicable.}\label{region_rf}
\centering
\begin{tabularx}{0.5\textwidth}{cccccc}
\hline
Method      & k=1             & k=5             & k=10            & k=25            & k=100           \\ \hline
RandomInit  & 0.511           & 0.516           & 0.525           & 0.532           & 0.553           \\
ImageNetPre & 0.512           & 0.515           & 0.518           & 0.513           & 0.507           \\
CONCH       & 0.556           & 0.624           & 0.705           & 0.788           & 0.869           \\
PLIP        & 0.608           & 0.730           & 0.796           & 0.830           & 0.876           \\
UNI         & 0.626           & 0.706           & 0.815           & 0.879           & 0.939           \\
RenalPath   & 0.664           & 0.749           & 0.786           & 0.845           & 0.898           \\
GloPath     & \textbf{0.668}  & \textbf{0.757}  & \textbf{0.843}  & \textbf{0.905}  & \textbf{0.945}  \\ \hline
\end{tabularx}
\end{table}

\begin{table}[]
\caption{Comparison of methods on deposition pattern classification on XJ-IF using RF-based few-shot learning. The bold value indicates the best model and the hyphen mean the metrics value is lower than 0.5 and not applicable.}\label{pattern_rf}
\centering
\begin{tabularx}{0.5\textwidth}{cccccc}
\hline
Method      & k=1             & k=5             & k=10            & k=25            & k=100           \\ \hline
RandomInit  & -               & 0.509           & 0.513           & 0.512           & 0.514           \\
ImageNetPre & -               & 0.501           & 0.508           & 0.507           & 0.510           \\
CONCH       & 0.537           & 0.659           & 0.687           & 0.741           & 0.852           \\
PLIP        & 0.658           & 0.759           & 0.752           & 0.774           & 0.860           \\
UNI         & 0.612           & 0.783           & 0.813           & 0.864           & 0.927           \\
RenalPath   & 0.667           & 0.769           & 0.785           & 0.812           & 0.889           \\
GloPath     & \textbf{0.645}  & \textbf{0.834}  & \textbf{0.843}  & \textbf{0.896}  & \textbf{0.939}  \\ \hline
\end{tabularx}
\end{table}

\begin{table}[]
\caption{Comparison of methods on deposition region classification on XJ-IF using PTL-based few-shot learning. The bold value indicates the best model and the hyphen mean the metrics value is lower than 0.5 and not applicable.}\label{region_tpl}
\centering
\begin{tabularx}{0.5\textwidth}{cccccc}
\hline
Method      & k=1             & k=5             & k=10            & k=25            & k=100           \\ \hline
RandomInit  & 0.506           & 0.514           & 0.529           & 0.541           & 0.556           \\
ImageNetPre & 0.499           & 0.515           & 0.531           & 0.536           & 0.558           \\
CONCH       & 0.552           & 0.610           & 0.687           & 0.751           & 0.841           \\
PLIP        & 0.592           & 0.722           & 0.797           & 0.827           & 0.860           \\
UNI         & 0.593           & 0.682           & 0.791           & 0.850           & 0.908           \\
RenalPath   & 0.592           & 0.721           & 0.791           & 0.833           & 0.863           \\
GloPath     & \textbf{0.614}  & \textbf{0.725}  & \textbf{0.820}  & \textbf{0.886}  & \textbf{0.928}  \\ \hline
\end{tabularx}
\end{table}

\begin{table}[]
\caption{Comparison of methods on deposition pattern classification on XJ-IF using PTL-based few-shot learning. The bold value indicates the best model and the hyphen mean the metrics value is lower than 0.5 and not applicable.}\label{pattern_tpl}
\centering
\begin{tabularx}{0.5\textwidth}{cccccc}
\hline
Method      & k=1             & k=5             & k=10            & k=25            & k=100           \\ \hline
RandomInit  & 0.508           & 0.508           & 0.517           & 0.526           & 0.559           \\
ImageNetPre & 0.503           & 0.508           & 0.512           & 0.520           & 0.555           \\
CONCH       & 0.537           & 0.640           & 0.660           & 0.708           & 0.818           \\
PLIP        & \textbf{0.653}  & 0.757           & 0.764           & 0.782           & 0.842           \\
UNI         & 0.634           & 0.759           & 0.794           & 0.841           & 0.907           \\
RenalPath   & 0.620           & 0.754           & 0.771           & 0.805           & 0.862           \\
GloPath     & 0.558           & \textbf{0.781}  & \textbf{0.793}  & \textbf{0.876}  & \textbf{0.929}  \\ \hline
\end{tabularx}
\end{table}

\begin{table}[]
\caption{Performance of GloPath in large-scale real-world study.}\label{openset}
\begin{tabular}{cccccccccccc}
\hline
Lesion  & MH    & PAS(+) & PAS(-) & Crum  & ADH   & SS    & MN    & MP    & EP    & Cre   & GS    \\ \hline
ROC-AUC & 0.897 & 0.857  & 0.983  & 0.747 & 0.790 & 0.920 & 0.919 & 0.982 & 0.999 & 0.954 & 0.997 \\ \hline
\end{tabular}
\end{table}

\begin{table}[]
\caption{Details of the glomerular lesion annotation on XJ-Light-1.}\label{27-lesion-anno}
\begin{tabularx}{\textwidth}{cXXXXXXXXXXXXXXXX}
\hline
Staining & MH  & PAS(+) & PAS(-) & Crum & ADH & SS & SFN & MN  & MP  & EP & Cre & GS  \\ \hline
PAS      & 905 & 479    & 75     & 31   & 222 & 76 & -   & 101 & 189 & 69 & 110 & 263 \\
MT       & 282 & -      & -      & -    & 39  & 30 & 67  & -   & -   & -  & 38  & 128 \\
PASM     & 298 & -      & 73     & -    & 44  & 34 & 38  & 216 & 132 & 64 & 114 & 164 \\ \hline
\end{tabularx}
\end{table}

\begin{table}[]
\caption{Details of the IF markers in XJ-IF.}\label{markers}
%
\end{table}

\begin{table}[]
\caption{Details of the annotated images on XJ-IF.}\label{xjif-anno}
\centering
\begin{tabularx}{0.5\textwidth}{ccccc}
\hline
\multirow{2}{*}{Class} & \multicolumn{2}{c}{Region} & \multicolumn{2}{c}{Pattern} \\
                       & Capillary    & Membrane    & Scatted      & Diffused     \\ \hline
Number                 & 1053         & 658         & 747          & 962          \\ \hline
\end{tabularx}
\end{table}

\begin{table}[]
\caption{Details of the clinical variables on XJ-Light-1.}\label{clinic_private}
\centering
%
\end{table}

\begin{table}[]
\caption{Details of the clinical variables on KPMP-G.}\label{clinic_public}
\centering
%
\end{table}

\begin{table}[]
\caption{Details of the morphological variables. $S$, $P$, $a$, and $b$ refer to the area enclosed by the contour, the perimeter, the major axis, and the minor axis of the ellipse fitted to the contour of the bow or tuft segmented by the model, respectively.}\label{morphology}
\centering
%
\end{table}

\end{document}